\documentclass[a4paper,fleqn]{cas-dc}

\usepackage{tabularx}
\usepackage{dblfloatfix,caption,afterpage}
\usepackage{multirow, makecell}
\usepackage{multicol}
\usepackage{array}
\usepackage{mathtools}
\usepackage{xurl}
\usepackage{nicefrac, xfrac}
\usepackage{enumitem}
\usepackage{subcaption}
\usepackage{placeins}
\usepackage{float}
\usepackage{wrapfig}
\usepackage{longtable}
\usepackage{booktabs}

\usepackage[authoryear]{natbib}

\def\tsc#1{\csdef{#1}{\textsc{\lowercase{#1}}\xspace}}
\tsc{WGM}
\tsc{QE}

\begin{document}
\let\WriteBookmarks\relax
\def\floatpagepagefraction{1}
\def\textpagefraction{.001}

\shorttitle{Beyond the Hamming Cliff: Compact Spatial Encodings Preserving Semantic Distance in Spiking Neural Networks}    

\shortauthors{Davies, S. et al.}  

\title [mode = title]{Beyond the Hamming Cliff: Compact Spatial Encodings Preserving Semantic Distance in Spiking Neural Networks}

\tnotemark[1] 

\tnotetext[1]{
By pre-processing input signals with Gray, Thermometer, Johnson and hybrid encodings before passing them to spiking neural networks running on SpiNNaker hardware, Davies et al. address the semantic-spatial disparity that causes the ``Hamming cliff'' in standard binary representations. Continuous semantic patterns are presented as a set of simultaneous spikes at the input layer, preserving ultra-low latency processing. Training the network on multiple continuous values under each scheme, and measuring accuracy, required homeostatic scaling and false-positive rates, shows that removing the cliff is necessary but not sufficient: Gray encoding removes it yet retains little global distance structure, whereas Thermometer and Johnson preserve semantic distance at a code width that grows with the value range. The proposed hybrid encodings contain all three as limiting cases of a single parameter, recovering most of that structure at a fraction of the width. The homeostatic factor is shown to set a radius of tolerance around each trained pattern, so that the choice of encoding alone decides which values the network treats as similar.
} 

%

\author[1]{Sergio Davies}[type=author, prefix=Dr, orcid=0000-0001-5330-5527]

\cormark[1]

\fnmark[1]

\ead{sergio.davies@mmu.ac.uk}

\ead[url]{https://www.mmu.ac.uk/staff/profile/dr-sergio-davies}

\credit{Conceptualisation of this study, Design of the methodology, Software development, Interpretation of the results, Preparation of the article}

\cortext[1]{Corresponding author}

\affiliation[1]{organization={Department of Computing and Mathematics, Manchester Metropolitan University},
            addressline={John Dalton Building, Oxford Road}, 
            city={Manchester},
            postcode={M15 6BH}, 
            country={United Kingdom}}



\author[2]{Muhammad Aitsam}[type=author, prefix=Dr]




\credit{Performing experiments on SpiNNaker2, Review of the article}

\author[1]{Morgan-Tyler Stone}[type=collab, prefix=Mr]




\credit{Review of the article}

\author[2,3]{Alessandro Di\ Nuovo}[type=collab, prefix=Prof.]




\credit{Review of the article}

\affiliation[2]{organization={Centre of Excellence in Artificial Intelligence and Robotics, Sheffield Hallam University},
            addressline={Howard Street}, 
            city={Sheffield},
            postcode={S1 1WB}, 
            country={United Kingdom}}

\affiliation[3]{organization={Universit\`{a} degli Studi di Catania},
            addressline={Viale Andrea Doria 6}, 
            city={Catania},
            postcode={95100}, 
            country={Italy}}

\author[1]{Liangxiu Han}[type=collab, prefix=Prof.]




\credit{Review of the article}

\begin{abstract}
Spatial encoding lets a spiking neural network classify a value without a temporal integration window, by presenting it as one pattern of simultaneous spikes, but makes classification depend on the geometry of the binary code.
In standard binary, adjacent values such as $127$ and $128$ differ in every bit, a ``Hamming cliff'' that forces supervised STDP networks with homeostasis into excessive scaling and false positives.
We present a controlled characterisation of this dependence rather than an application benchmark: every experiment runs on SpiNNaker hardware, in a dual-rail network with a single output neuron, and presents every value of a synthetic integer range.
Gray, Thermometer and Johnson encodings all remove the cliff, but only Thermometer and Johnson preserve global semantic distance, at a width that grows linearly with the range.
We introduce a family of hybrid encodings, combining a coarse magnitude code with a Gray code for fine position, that contains all three as exact limiting cases of a single parameter; a sevenfold reduction in width costs one percent of the correlation between Hamming and semantic distance.
The homeostatic factor sets a Hamming radius of tolerance that the network follows exactly in $93\%$ of conditions, and elsewhere to within one bit, so the encoding alone determines what that tolerance means: a contiguous interval for Thermometer and Johnson, wrapping at the range ends for the cyclic codes, and scattered values for Binary and Gray.
Measuring the limit on simultaneous input on SpiNNaker and SpiNNaker2, we show that on SpiNNaker it is a silent software default, and that the hybrids remain within it at $1024$ values where their parent codes do not. Evaluation on a downstream task, such as network-traffic classification, is left to future work.
\end{abstract}



\begin{keywords}
Spiking Neural Networks \sep SNN \sep Spatial Pattern \sep STDP \sep Spike Timing Dependent Plasticity \sep Supervised Learning \sep Spike Encoding \sep Hamming Distance
\end{keywords}

\maketitle

\section{Introduction}
Artificial Neural Networks (ANNs) owe much of their success to gradient-based training \citep{Linnainmaa1976TaylorError, Schmidhuber2015DeepOverview}.
In these second-generation networks \citep{Maass1997NetworksModels}, inputs are continuous values, so numerically adjacent inputs map to adjacent activation levels and semantic proximity is preserved by construction; biologically, this corresponds to a rate code, in which firing frequency carries the magnitude of the stimulus \citep{Gerstner2014NeuronalDynamics}.
Third-generation Spiking Neural Networks (SNNs) \citep{Maass1997NetworksModels} add time to the model, and a rate code then costs latency: to decode a value, the network must ``wait and count'' spikes over an integration window.
It also costs energy. Event-driven neuromorphic hardware spends its energy per event rather than per clock cycle, routing every spike as a packet and processing it at each synapse it reaches \citep{Furber2014TheProject, Davidson2021ComparisonHardware}, so a code that spends many spikes on each value undermines both the low latency and the low power that such hardware promises \citep{Davies2018Loihi:Learning}.

Spatial encoding avoids this bottleneck by representing a value as a pattern of simultaneous spikes \citep{Davies2025SupervisedSpiNNaker}.
Because neuromorphic hardware delivers information through discrete, addressable synapses, the value is first converted to a binary code word, and each bit is signalled by the presence or absence of a spike from one source \citep{Auge2021ANetworks}.
The input activity per decision is then fixed by the code width, and the output is delayed only by the depth of the network \citep{Goltz2021FastTimes}.
The conversion to binary, however, introduces a structural vulnerability, the ``Hamming cliff'' \citep{Caruana1988RepresentationAlgorithms}: numerically adjacent values can have very different code words.
The values $127$ and $128$, for instance, are $01111111$ and $10000000$ in 8-bit binary, so two adjacent values differ in every bit, and semantic proximity and Hamming distance \citep{Hamming1950ErrorCodes} disagree as strongly as they can.
The 8-bit range is used for most of our experiments, as the widest at which every encoding compared here fits within the default input limit of first-generation SpiNNaker (Section~\ref{sec:packetlimit}).

In earlier work \citep{Davies2025SupervisedSpiNNaker}, we showed that an SNN trained with supervised STDP classifies a single spatial pattern accurately, but that its performance degrades when it is trained on several.
Large Hamming distances between the trained patterns create ``don't care'' synapses, which receive conflicting evidence from different patterns and so contribute nothing to the output; the network must then rely on stronger homeostatic scaling, which sharply increases false positives.
Standard binary encoding therefore prevents the network from recognising contiguous ranges of values.
This matters wherever the inputs are continuous or ordinal.
In network-traffic classification, for instance, packet sizes, flow durations and inter-arrival times carry meaning in their proximity: two flows of similar duration are similar events, and an encoding that maps them to distant spatial patterns discards that information before the network sees it.
Other features are cyclic: time of day is the clearest case, where 23:59 and 00:00 are adjacent, and such features need an encoding whose topology reflects that.

This paper examines spatial encodings that restructure the signal before it reaches the network, while keeping one spike per bit.
The established alternatives are not equivalent: Gray encoding removes the cliff but preserves little global distance structure, whereas Thermometer and Johnson encodings preserve it at the cost of a code width that grows linearly with the value range.
We introduce a family of hybrid encodings, which combine a coarse magnitude code with a Gray code for fine position; the family contains all three as limiting cases of a single parameter and trades code width against preservation of semantic distance at a favourable rate.
On hardware, the homeostatic factor sets a radius of tolerance in code space around each trained code word, so the encoding alone decides which values the network treats as similar.
The same analysis explains the degradation observed in our earlier work \citep{Davies2025SupervisedSpiNNaker}: the summed weights of trained code words cancel wherever they disagree, and the homeostatic gain that compensates is what admits the false positives.
Because a spatial code delivers all its bits at once, its width is also bounded by the number of simultaneous input events a neuron can receive: about $255$ under the default configuration of first-generation SpiNNaker, and $512$ on SpiNNaker2 \citep{Mayr2019SpiNNakerLearning,Gonzalez2024SpiNNaker2:Learning}.
The hybrid construction reduces code width by an order of magnitude.
To our knowledge, this is the first systematic characterisation of how the choice of spatial code governs the homeostatic scaling, and hence the tolerance and false-positive behaviour, of supervised STDP learning on neuromorphic hardware.

The main contributions of this work are:
\begin{itemize}
    \item an account of how the geometry of a spatial code sets the homeostatic tolerance of a supervised STDP network (Sections~\ref{sec:coherence} and~\ref{sec:exp5});
    \item a hybrid coarse--fine encoding family with Thermometer, Johnson and Gray as limiting cases (Section~\ref{sec:encodings});
    \item evidence that removing the Hamming cliff is necessary but not sufficient for semantic clustering (Sections~\ref{sec:exp1} and~\ref{sec:exp3});
    \item a characterisation of the per-neuron input limits on SpiNNaker and SpiNNaker2 (Section~\ref{sec:packetlimit});
    \item an exhaustive hardware evaluation of eight encodings, with code and data released (Sections~\ref{sec:exp2}--\ref{sec:exp6}).
\end{itemize}

The scope of this work is deliberately narrow: we characterise the encodings rather than evaluate them on a task.
Every experiment presents every value in the range to the trained network, so the relations reported are exact over the whole input space and a relation such as the tolerance law of Section~\ref{sec:exp5} can be checked condition by condition. A natural dataset allows neither: its values are unevenly distributed and its features correlated, so the effect of code geometry on learning would be confounded with the data.
The network (a dual-rail input layer feeding a single output neuron), the learning rule and the measurement protocol are held constant, so the encoding is the only variable, and the comparison throughout is against standard binary encoding, the representation used in our prior work.
How the ranking carries over to a downstream task, where feature distributions are uneven and the encodings interact with later processing, is the subject of separate work.

\section{Theoretical Background}
\label{sec:background}

\subsection{Spatial Encoding and the Hamming Vulnerability}

In spatial encoding, all bits of a code word are presented to the network
in the same timestep, so the network has no access to the value itself,
only to which of its input synapses are active. Its response to two
values therefore depends on how many bits their code words share, that
is, on the Hamming distance between them. In the supervised STDP network
of \citet{Davies2025SupervisedSpiNNaker}, a larger Hamming distance
between trained code words required a larger homeostatic scaling factor
for all of them to be recognised, and the larger factor admitted more
false positives. The following subsections formalise this dependence and
the properties of the encodings that determine it.

\subsection{Formal Definition of the Hamming Cliff}
\label{sec:cliff}

\subsubsection{Notation}

Let $\mathcal{S} = \{0, 1, \dots, N-1\}$ be a set of $N$ discrete semantic values.
An encoding $f$ maps each value to a \emph{code word}: a string of $W$ symbols, each either $0$ or $1$. We write $c_i = f(i)$ for the code word of value $i$, $W$ for the \emph{code width}, that is the number of bits in every code word of a given encoding, and $H(c_i, c_j)$ for the Hamming distance between two code words, the number of positions at which they differ \citep{Hamming1950ErrorCodes}.

A width-$W$ code admits $2^{W}$ distinct strings, of which an encoding uses $N$, one per semantic value; hence $W \geq \log_2 N$.
Standard binary attains this minimum: taking $W = \log_2 N$, every available string is used, so $N = 2^{W}$, and each value is represented by its ordinary binary expansion.
The encodings introduced below are wider and use only a small fraction of the strings available to them: Thermometer, for example, has $W = N-1$ and uses $N$ of the $2^{N-1}$ possible strings.
The width is what the hardware must carry for each classification, and Section~\ref{sec:packetlimit} shows that it is bounded.

\subsubsection{Preservation of semantic proximity}
\label{sec:proximity}

Semantic proximity depends on the topology of the quantity being
encoded. We therefore write $d_S(i,j)$ for the semantic distance between
two values, and consider two cases: a bounded quantity, such as a
magnitude, with linear distance
\begin{equation}
d_{\text{lin}}(i,j) = |i - j|,
\end{equation}
and a periodic quantity, such as time of day or an angle, with circular
distance
\begin{equation}
d_{\text{cyc}}(i,j) = \min\bigl(|i - j|,\; N - |i - j|\bigr),
\end{equation}
under which the extremes $0$ and $N-1$ are adjacent. Two values that are
close under $d_S$ should have code words that are close in Hamming
distance. We therefore say that $f$ preserves semantic proximity under
$d_S$ when the Hamming distance between code words grows with the
semantic distance between the values they represent:
\begin{multline}
d_S(i,j) \leq d_S(i,k) \;\Longrightarrow\; H(c_i, c_j) \leq H(c_i, c_k), \\
\forall\, i,j,k \in \mathcal{S}.
\label{eq:monotone}
\end{multline}
An encoding is judged against the metric of the quantity it is meant to
represent: a linear code against $d_{\text{lin}}$, a cyclic code against
$d_{\text{cyc}}$.

\subsubsection{The cliff}

A Hamming cliff is a local violation of Eq.~\eqref{eq:monotone}: two
values that are adjacent under $d_S$ but whose code words are
disproportionately far apart. Taking the smallest distance between any
two distinct code words as the natural unit of comparison, $f$ suffers
from a Hamming cliff between two values $i$ and $j$ with $d_S(i,j) = 1$
when
\begin{equation}
H(c_i, c_j) \;\gg\; \min_{\substack{k,l \in \mathcal{S} \\ k \neq l}}
H(c_k, c_l).
\label{eq:cliff}
\end{equation}
The relation $\gg$ is left informal here; Section~\ref{sec:exp1} makes it
quantitative by reporting the ratio of the two sides. Under
$d_{\text{cyc}}$ the values $0$ and $N-1$ are adjacent, so a code used for
a periodic quantity can also have a cliff across the wraparound.

In standard binary encoding the cliff occurs at power-of-two transitions,
most severely between $2^{W-1}-1$ and $2^{W-1}$, where the transition
costs up to $W$ bits although the two values are adjacent and the
expected distance is $H = 1$. For $W = 8$ this is the pair $127$ and
$128$ introduced above, at a cost of $8$ bits against an expected $1$.

\subsubsection{Consequence for the network}

In an SNN trained by supervised STDP, homeostasis rescales every synaptic weight by a single common factor, chosen so that the network responds to each of the patterns it was trained on.
Write $S_i$ for the smallest factor at which pattern $c_i$ causes the output neuron to fire; the factor adopted by the network is the largest of these over the trained set.
$S_i$ grows with the number of synapses that disagree with the target pattern $T$, because those synapses contribute nothing to the output and the remainder must compensate for their absence.

A Hamming cliff between two adjacent values therefore produces an abrupt change in the scaling they demand,
\begin{equation}
\Delta S = |S_{i+1} - S_i| \;\propto\;
\bigl| H(f(i+1), T) - H(f(i), T) \bigr|
\label{eq:scalingjump}
\end{equation}
and it is this compensatory scaling that produces the instability and the false-positive activations reported by \citet{Davies2025SupervisedSpiNNaker}.

\subsection{Encoding Topologies}
\label{sec:enc-top}

To address the Hamming cliff, we consider alternative mapping techniques that constrain the relationship between semantic and spatial distance:

\begin{itemize}
    \item \textbf{Gray Encoding:} A reflected binary code where adjacent integers are represented by codewords that differ by only a single bit, ensuring minimal spatial transitions between semantically adjacent states \citep{Gray1953PulseCommunication}.
    \item \textbf{Thermometer Encoding:} Maps a magnitude by progressively filling a bit-field, creating a \emph{unary} representation: the value is given by the \emph{number} of bits set, not by which particular bits are set \citep{Buckman2018ThermometerExamples}. Representing $N$ values therefore requires $N-1$ bits, and successive values differ in a single bit.
    \item \textbf{Johnson (Twisted Ring) Encoding:} A cyclic code in which bits are set one at a time from one end of the field and then cleared again from that same end, so that $W$ bits generate $2W$ distinct states before returning to the start \citep{Mano2012DigitalDesign}. Successive values differ in a single bit, and because the sequence closes on itself the first and last values are adjacent.
\end{itemize}

These codes are long established in digital logic. The criterion that
bit-flip distance should track semantic distance is also established in
the weightless neural network literature, where binary encodings for
WiSARD are evaluated on exactly this basis \citep{Kappaun2016EvaluatingWiSARD}, and unary
thermometer codes are the standard solution \citep{Bacellar2022DistributiveNetworks}. Codes
that combine a coarse unary field with a fine binary one are likewise
familiar from segmented data converters. What has not been examined is
their use as spatial codes for spiking neural networks, where each bit
is a discrete synapse, all bits are presented simultaneously, and
homeostatic scaling determines which code words the network accepts.
The contribution of this paper lies there: a coarse and fine
construction whose Gray fine field is reflected in alternate blocks, so
that single-bit adjacency survives every block boundary, and a
quantitative account of how the topology of such a code determines both
the network's tolerance and its demands on the hardware.

\subsection{Topological Analysis}
\label{sec:topology}

The structural utility of these schemes is best understood through their
distinct topological properties, as defined in digital logic theory
\citep{Mano2012DigitalDesign}. Thermometer encoding enforces linear
boundary conditions on the semantic space, wherein the minimum and
maximum values are maintained at maximal Hamming distance. Conversely,
Johnson encoding imposes a periodic topology; by forming a `twisted
ring', the encoding identifies the semantic extremes $0$ and $N-1$ as
adjacent, regardless of the code width $W$. This cyclic property makes
Johnson encoding a natural candidate for quantities with circular
continuity, such as orientation, phase or time of day, whereas
Thermometer encoding suits bounded magnitudes.

These schemes must, however, be distinguished by \emph{which} property
they preserve. All three guarantee \emph{local} smoothness: values
adjacent under the metric each is designed for differ in a single bit,
eliminating the cliff of Section~\ref{sec:cliff}. Only Thermometer and
Johnson additionally preserve \emph{global} structure, each exactly
under its own metric: $H(c_i, c_j) = d_{\text{lin}}(i,j)$ for
Thermometer and $H(c_i, c_j) = d_{\text{cyc}}(i,j)$ for Johnson. Gray
encoding offers no such guarantee. Its reflected construction places
every value $i$ a single bit from $2^W - 1 - i$, however far apart the
two are, so that the extremes $0$ and $2^W - 1$ differ in one bit;
Hamming distance therefore carries little information about semantic
separation beyond immediate neighbours.

Removing the Hamming cliff is therefore necessary but not sufficient for
the clustering behaviour sought here, and Section~\ref{sec:exp1}
quantifies the gap between the two properties. These three codes are
not, however, exclusive alternatives: the reflected construction of
Section~\ref{sec:encodings} interpolates between them and contains all
three as limiting cases.

\subsection{Relation to Other Low-Latency Representations}
\label{sec:representations}

\begin{table*}[t]
\centering
\caption{Low-latency input representations compared on the three
properties that distinguish a spatial code. Event counts are for a
scalar of $N$ levels; for the spatial codes every bit fires on one rail
of its pair, so a code of width $W$ delivers $W$ events whatever its
content. $T$ is the length of the encoding window, $d$ the semantic
distance $d_S$, and $k$ the width of a binarised receptive field.}
\label{tab:representations}
\renewcommand{\arraystretch}{1.35}
\begin{tabular}{@{}>{\raggedright\arraybackslash}p{0.17\textwidth}
                >{\raggedright\arraybackslash}p{0.19\textwidth}
                >{\raggedright\arraybackslash}p{0.17\textwidth}
                >{\raggedright\arraybackslash}p{0.38\textwidth}@{}}
\toprule
\textbf{Representation} & \textbf{Decision available} &
\textbf{Input events per value} & \textbf{How code distance relates to $d$} \\
\midrule
Rate & after an integration window & many, growing with precision & graded, in spike counts \\
Latency (time to first spike) & up to $T$, depending on the value & one per input neuron & graded, in spike times \\
Rank order & once enough of the order has arrived & one per neuron of the population & relative order across neurons, not the magnitude of one value \\
Population, Gaussian receptive fields & usually up to $T$, as fields are read out in time & several, one per active field & preserved locally; binarised to a window of $k$ active neurons, $H = 2\min(d, k)$, saturating beyond the field width \\
One-hot (place) & one timestep & $N$ & none: every distinct pair at $H = 2$ \\
\midrule
Binary & one timestep & $\lceil \log_2 N \rceil$ & none, with cliffs at power-of-two boundaries \\
Gray & one timestep & $\lceil \log_2 N \rceil$ & local only: $H = 1$ for $d = 1$ \\
Thermometer, Johnson & one timestep & $N - 1$, $N/2$ & exact: $H = d$, linear or circular \\
Hybrid (this work) & one timestep & $\approx N/B + \log_2 B$ & global, slope $1/B$, ripple at most $\log_2 B$ \\
\bottomrule
\end{tabular}
\end{table*}

Spatial binary codes are one of several ways to give a spiking network a
fast decision about a scalar \citep{Auge2021ANetworks}. They are
distinguished from the others by three properties: when the decision
becomes available, how many input events it costs, and how the distance
between code words relates to semantic separation
(Table~\ref{tab:representations}).

Latency coding, or time to first spike, represents a value by when a
single spike occurs \citep{Mostafa2017SupervisedNetworks, Goltz2021FastTimes}. It is
frugal in events, one per input neuron, and preserves proximity
naturally, since close values fire at close times; but the decision
cannot be taken until the latest spike that could carry information has
had time to arrive, so its latency is bounded by the encoding window
rather than by the depth of the network. Rank-order coding
\citep{Thorpe1998RankCoding} reads information from the order in which a
population fires. It is equally frugal, but it encodes the relative
ordering of several inputs rather than the magnitude of one, and it too
must wait for that order to be established. Both trade time for events.
Spatial codes make the opposite trade: every bit is delivered in the
same timestep, so the decision follows within a fixed number of
timesteps of the input, and the cost is paid in width instead.

Population coding with Gaussian receptive fields \citep{Bohte2002Error-backpropagationNeurons}
preserves semantic proximity through overlapping graded fields, and is
usually read out in time, the strength of each field becoming the
latency of its spike. Binarised and presented spatially, it becomes a
window of $k$ active neurons that moves with the value, and its Hamming
distance at semantic separation $d$ is $2\min(d, k)$: proximity is
preserved within the width of a field and lost beyond it, every pair of
values further apart than $k$ being equidistant. One-hot, or place,
coding is the limiting case $k = 1$, in which every pair of distinct
values lies at distance $2$ and code distance carries no information
about separation at all. A place-coded population is nevertheless the
natural input layer from which a spatial code can be computed on-chip,
and in a single-rail arrangement it costs only one event per value; the
dual-rail wiring used here, in which every bit must be presented,
removes that advantage.

The spatial codes studied in this paper differ from all of these in that
distance in code space, rather than in time, carries semantic
separation, and they differ from one another in how much of it survives.
Thermometer and Johnson preserve it exactly and globally, where a
population window preserves it only within its own width, and the
hybrid construction introduced here preserves it globally at a fraction
of the width. It is this global structure, rather than local smoothness
alone, that homeostatic tolerance turns into a semantic interval.

\section{Methods}
\label{sec:methods}

\subsection{Network Architecture}
\label{sec:architecture}

The network follows the supervised STDP arrangement of \citet{Davies2025SupervisedSpiNNaker}.
A code word is presented as a dual-rail spatial pattern: for each bit, one neuron fires if the bit is $1$ and a second fires if it is $0$, so exactly one of the two rails is active per bit and the whole code word is presented within a single simulation timestep.
All neurons are leaky integrate-and-fire units with delta synapses, which deliver their current within one timestep, with the parameters listed in Table~\ref{tab:neuron}.

\begin{table}[t]
\caption{Neuron, synapse and simulation parameters. Neuron values are the sPyNNaker defaults for \texttt{IF\_curr\_delta}, as logged on the hardware.}
\label{tab:neuron}
\centering
\resizebox{\columnwidth}{!}{
\begin{tabular}{llr}
\toprule
Parameter & Symbol & Value \\
\midrule
Membrane time constant & $\tau_m$ & $20$\,ms \\
Membrane capacitance & $C_m$ & $1.0$\,nF \\
Resting potential & $V_\text{rest}$ & $-65$\,mV \\
Reset potential & $V_\text{reset}$ & $-65$\,mV \\
Firing threshold & $V_\text{th}$ & $-50$\,mV$^{a}$ \\
Refractory period & $\tau_\text{refrac}$ & $0.1$\,ms \\
Offset current & $I_\text{offset}$ & $0$\,nA \\
STDP time constants & $\tau_+,\tau_-$ & $5$\,ms \\
STDP amplitudes & $A_+,A_-$ & $1,\,-1$ \\
Maximum weight & $w_\text{max}$ & $2$ \\
Timestep & & $1$\,ms \\
Time scale factor & & $100$ \\
Incoming spike buffer & & $255$ spikes$^{b}$ \\
\bottomrule
\multicolumn{3}{p{0.9\columnwidth}}{\raggedright\footnotesize $^{a}$\,Scaled with code width for the output neuron (Section~\ref{sec:deviations}).}\\
\multicolumn{3}{p{0.9\columnwidth}}{\raggedright\footnotesize $^{b}$\,Allocated with $256$ slots (sPyNNaker default \texttt{incoming\_spike\_buffer\_size}), one of which is always kept empty. During training, 256 input events were occasionally accepted when a spike was consumed during the burst (Section~\ref{sec:packetlimit}).}
\end{tabular}
}
\end{table}

\begin{figure}
\centering
\includegraphics[width=\linewidth]{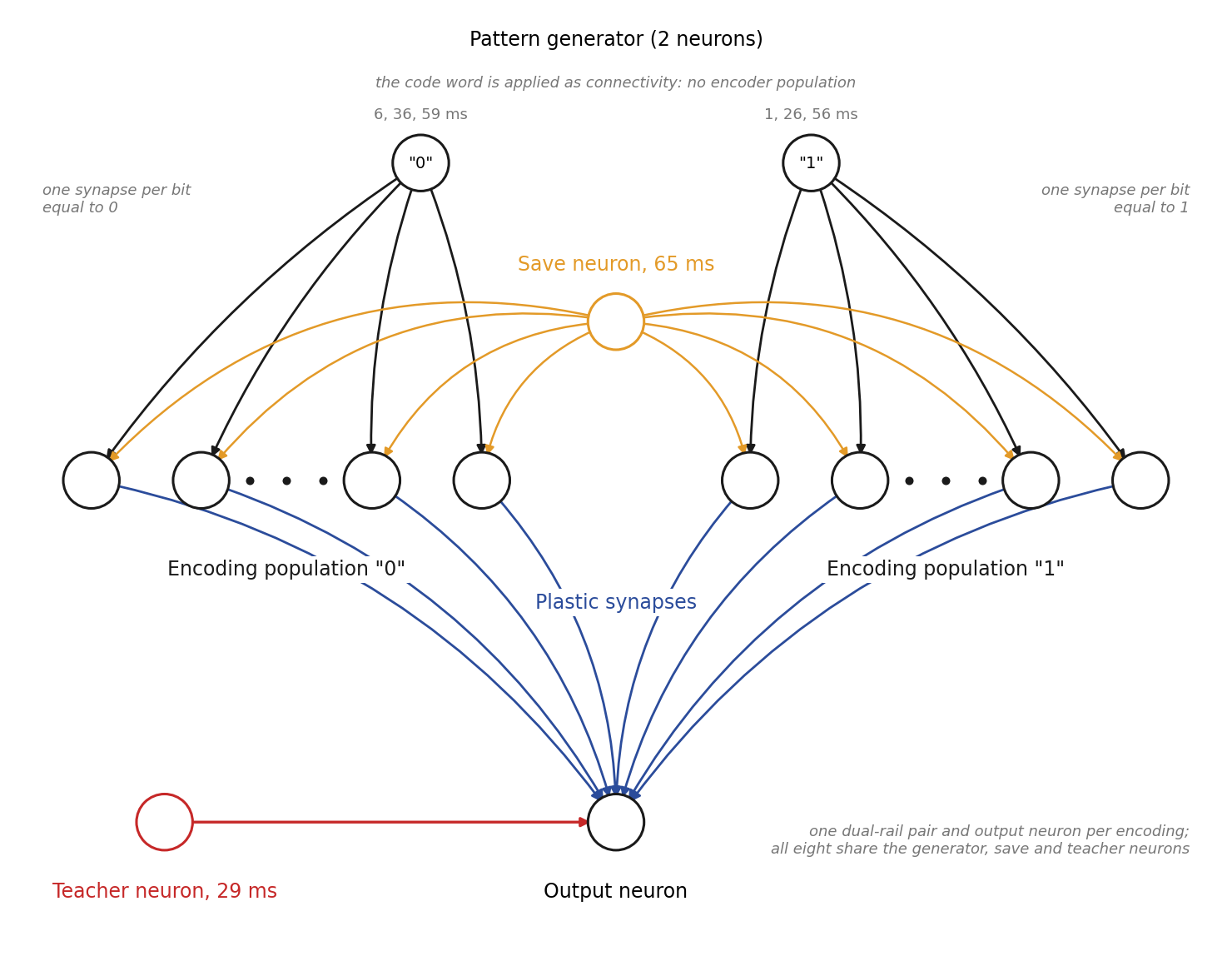}
\caption{The network used for training. The pattern generator's two neurons fire at the times encoding a ``1'' and a ``0'' respectively, and the code word of the presented value is realised as the connectivity between them and the dual-rail encoding populations (Section~\ref{sec:encodingstep}). In blue the STDP-enabled synapses, whose weights begin at zero; in orange the save neuron, which triggers the deferred weight updates; in red the teacher neuron, the only source of drive to the output neuron during training. Each encoding has its own pair of encoding populations and its own output neuron, and all eight share the generator, save and teacher neurons.}
\label{fig:train}
\end{figure}

\subsubsection{Training}

The two rails (Fig.~\ref{fig:train}) project to a single output neuron through plastic synapses governed by a nearest-pair STDP rule, in which each pre-synaptic spike is paired only with the nearest post-synaptic spike rather than with all of them ($\tau_+ = \tau_- = 5$~ms, $A_+ = 1$, $A_- = -1$).
Spikes encoding a ``1'' are presented at $1$, $26$ and $56$~ms and those encoding a ``0'' at $6$, $36$ and $59$~ms, while a teacher neuron fires at $29$~ms and drives the output neuron to fire at $32$~ms.
The two rails are offset from the teacher spike by different intervals:
given the time constants above, the nearer interval produces a large
weight change and the further one a negligible change, so the STDP rule
potentiates the synapses of bits matching the presented value and
depresses the others.
The simulator computes STDP weight changes only when a pre-synaptic spike
arrives, so a save neuron fires at $65$~ms; this final spike serves no
purpose other than to trigger that computation and make the learned
weights available to be read back \citep{Jin2010ImplementingHardware}.

All synapses in the training network are excitatory, and the plastic
weights begin at zero. The output neuron therefore fires solely under the
teacher's drive, and the plastic synapses serve only to record the spike
timings from which STDP computes the weights.

\begin{figure}
\centering
\includegraphics[width=\linewidth]{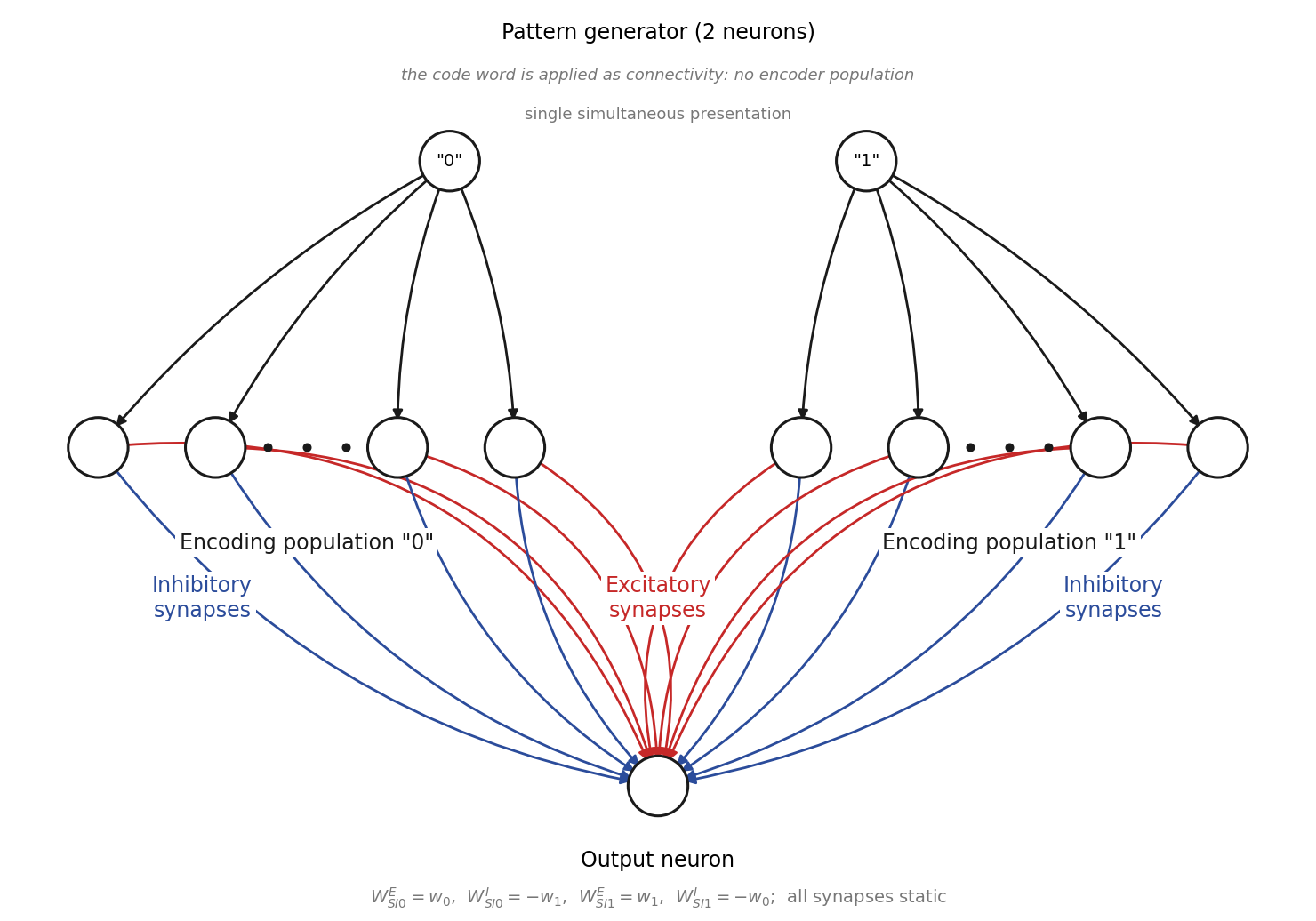}
\caption{The network used for testing. The teacher and save neurons are removed and the learned weights are applied as static synapses, wired reciprocally so that a bit on which two trained code words disagree contributes equal excitation and inhibition and cancels exactly (Eq.~\ref{eq:reciprocal}). In red the excitatory synapses, in blue the inhibitory ones.}
\label{fig:test}
\end{figure}

\subsubsection{Testing}
\label{sec:testing}

The teacher and save neurons are removed, the code word is presented once, and the learned weights are applied as static synapses wired reciprocally (Fig.~\ref{fig:test}): rail~1 projects excitatory weight $w_1$ and inhibitory weight $w_0$ to the output neuron, and rail~0 the converse,

\begin{equation}
W^{I}_{SI0} = -W^{E}_{SI1}, \qquad W^{I}_{SI1} = -W^{E}_{SI0}.
\label{eq:reciprocal}
\end{equation}

Inhibition appears only in this network.
A bit on which two trained code words disagree acquires equal weight on both rails and therefore contributes equal excitation and inhibition, cancelling exactly: the ``don't care'' synapse of \citet{Davies2025SupervisedSpiNNaker}.
Classification is the firing of the output neuron within one timestep of the input, so the decision latency is fixed by the depth of the network rather than by an integration window.

Each of the eight encodings has its own dual-rail pair and its own output neuron, and all eight are trained in a single simulation; the pattern generator, save neuron and teacher neuron are shared between them.

\subsubsection{Deviations from the previous work}
\label{sec:deviations}

Two parameters differ from \citet{Davies2025SupervisedSpiNNaker}, both required by the wider code words used here.
First, the output neuron's firing threshold is scaled with the code width.
A width-$W$ code delivers up to $W$ potentiated synapses' worth of current in a single timestep, which for wide codes exceeds the default threshold and causes the output neuron to fire without the teacher, corrupting the learned pattern; the threshold is therefore set to twice the accumulated drive, subject to a floor at the default value.

Second, the STDP weight ceiling $w_{max}$ is reduced from $20$ to $2$.
The simulator holds synaptic weights in a fixed-point accumulator whose scaling is fixed at run time by the largest total current that could arrive at a neuron, namely $w_{max}$ times the fan-in \citep{vanAlbada2018PerformanceModel}; a larger product consumes more integer bits and leaves fewer fractional ones, coarsening the resolution.
At the widest code considered the fan-in reaches $510$, two rails of $255$ bits, and at the default ceiling the resulting step size is comparable to the learned weight of $\approx 0.37$ itself, which therefore cannot be represented.
Lowering the ceiling restores the fractional precision, and the learned weight is then recorded as $0.375$.

Neither affects what is learned: homeostasis rescales the weights afterwards, and \citet{Davies2025SupervisedSpiNNaker} note that the STDP parameter values are of little consequence for exactly this reason.
A corollary is that the absolute homeostatic factors of encodings with different widths are not directly comparable, since their thresholds differ; comparisons in Section~\ref{sec:results} are therefore made within an encoding, or after normalisation.

\subsection{Application of the Encoding}
\label{sec:encodingstep}

The encoding is applied as a pre-processing step rather than computed by the network. For a given value, the code word determines the connectivity: the generator neuron that fires at the ``1'' times (Section~\ref{sec:architecture}) connects to every bit of the code word equal to one, and the neuron that fires at the ``0'' times to every bit equal to zero.
The network is therefore constructed anew for each value presented, and no population computes the mapping from value to code word. This matches the framing of the work as a pre-processing of the input signal, and it also removes the source-side spike load: each generator neuron emits one packet per presentation time, which the router fans out, rather than one packet per bit.

The same mapping could be computed on-chip, by a place-coded input population with one neuron per value projecting to the bits of its own code word through fixed connectivity.
Such a network would present identical activity to the encoding populations, so the results reported here do not depend on which of the two arrangements is used.

\begin{table}[t]
\centering
\caption{The five classification metrics used in \citet{Davies2025SupervisedSpiNNaker}, evaluated for every experiment and tabulated in full in Appendix~\ref{app:metrics}.}
\label{tab:metrics}
\renewcommand{\arraystretch}{1.5}
\begin{tabular}{@{}l c >{\raggedright\arraybackslash}m{0.40\linewidth}@{}}
\toprule
\textbf{Metric} & \textbf{Formula} & \textbf{What it measures} \\
\midrule
Accuracy & $\dfrac{t_p+t_n}{t_p+t_n+f_p+f_n}$
  & the share of all values classified correctly; dominated by $t_n$
    when few values are trained \\
    \midrule
Precision & $\dfrac{t_p}{t_p+f_p}$
  & the share of accepted values that were trained \\
  \midrule
\makecell[l]{Negative \\ prediction} & $\dfrac{t_n}{t_n+f_n}$
  & the share of rejected values that were untrained \\
  \midrule
Sensitivity & $\dfrac{t_p}{t_p+f_n}$
  & the share of trained values that the network detects \\
  \midrule
Specificity & $\dfrac{t_n}{t_n+f_p}$
  & the share of untrained values that the network rejects \\
\bottomrule
\end{tabular}
\end{table}

\subsection{Training and Testing Procedure}
\label{sec:procedure}

For each experiment a set of values is trained, one value per simulation
and each on a network whose plastic weights begin at zero, and the
resulting weight vectors are summed synapse by synapse, following
\citet{Davies2025SupervisedSpiNNaker}. This procedure makes the learned
weights a linear function of the trained set; its relation to sequential
learning on a single network is discussed in
Section~\ref{sec:limitations}. Where a single value serves as a
reference, we use $240 = 11110000_2$, the 8-bit analogue of the code word
$992 = 1111100000_2$ used throughout \citet{Davies2025SupervisedSpiNNaker},
so that results here can be compared with that work.

Homeostasis is then applied as a single multiplier on every synaptic
weight. The multiplier is determined per encoding by binary search: the
smallest factor at which every trained value causes the output neuron to
fire. Each trained value is searched independently and the maximum taken,
then rounded up to the next multiple of $0.01$ so that the network does
not operate exactly at the firing boundary. The search resolves the
factor to $0.01$, which bounds the precision of the tolerance
measurements in Section~\ref{sec:exp5}.

Classification is then measured exhaustively: every value in the range is
presented and the output recorded. A value is \emph{accepted} when the
output neuron fires for it. Accepted values that were trained are true
positives ($t_p$) and those that were not are false positives ($f_p$);
rejected values are true negatives ($t_n$) and false negatives ($f_n$)
respectively. Table~\ref{tab:metrics} gives the five classification
metrics derived from these counts. The main text reports accuracy and
precision, which are the two that discriminate between encodings here;
sensitivity is unity throughout, since the homeostatic search guarantees
that every trained value fires, and serves as a check that training
succeeded rather than as a measure of performance. The complete set is
tabulated for each experiment in Appendix~\ref{app:metrics}.

The false-positive count follows \citet{Davies2025SupervisedSpiNNaker},
but it does not distinguish two quite different outcomes. An accepted
value adjacent to, or lying between, trained values is generalisation
over a neighbourhood, and for an encoding that preserves semantic
distance it is the behaviour sought; an accepted value far from every
trained value is an error in the ordinary sense. We refer to the first as
a \emph{neighbouring activation}. Sections~\ref{sec:exp2}
and~\ref{sec:exp3} report where accepted values lie as well as how many
there are, so that the two can be told apart.

Several independent replicas of the testing network share each
simulation, each with its own generator, rails and output neuron and
injecting at a staggered time. Across the four experiments reported here,
$216$ trained networks were measured against every value in their range,
giving $131{,}072$ individual classifications on hardware. No result
below is sampled or extrapolated.

All experiments were run on first-generation SpiNNaker hardware using the
sPyNNaker implementation of PyNN \citep{Rhodes2018SPyNNaker:SpiNNaker,
Davison2008PyNN:Simulators}, with the default incoming spike buffer, which holds $255$ packets, and a time scale factor of $100$, so that
each $1$~ms timestep spans more than a millisecond of wall-clock time;
latencies are therefore reported in timesteps. The input-capacity
measurements of Section~\ref{sec:packetlimit} were additionally made
on SpiNNaker2; every other result is from first-generation SpiNNaker.
Code and results are available on GitHub \citep{Davies2026EncodedDataset}
at the URL: \url{https://github.com/sergiodavies/EncodedSpatialLearningCodeAndDataset}.

\subsection{Input Capacity and Spatial Code Width}
\label{sec:packetlimit}

Because all bits of a spatial code word are presented simultaneously and
each bit fires on exactly one rail, a code of width $W$ delivers $W$ spike
packets to its output neuron within a single timestep. Any limit on the
number of packets a neuron can receive at once is therefore a limit on
code width. We measured this limit on both platforms.

\subsubsection{SpiNNaker}

sPyNNaker places incoming packets in a circular buffer on each core, allocated with $256$ slots by default, of which one is always kept empty, so that at most $255$ packets can be held; packets that arrive when the buffer is full are discarded without error.

Training an output neuron on code words of increasing width under the default configuration, the number of synapses potentiated tracked the width exactly up to $255$ and then stopped, at $255$ or occasionally $256$ when a packet was consumed during the burst, for widths of $300$, $480$ and $992$. Raising the buffer to $1024$ slots removed the limit: every synapse was potentiated at every width tested. The bound is therefore a default of the software rather than a property of the hardware. It is, however, silent: the network learns a truncated pattern while reporting no fault, and the downstream metrics remain plausible.

To see whether the limit is a fixed capacity, the measurement was repeated under the default buffer with static synapses, as used in the testing network, counting the inputs delivered from the deflection of a non-leaking membrane. Here the limit is neither fixed at $255$ nor abrupt: every input arrived up to $385$, and beyond that about two in five additional inputs still did. Spreading the same inputs across two and four source populations raised the point at which loss began to $426$ and $435$. Both observations indicate a buffer that drains while it fills: spreading the inputs spreads their arrival in time, and static synapses are cheap enough to process that the buffer empties appreciably during a burst. Plastic synapses cost more per spike, so during training it barely drains and its depth alone sets the limit. The limit was the same at every synaptic weight tested, so it is a count of events rather than of current.

\subsubsection{SpiNNaker2}

The same deflection measurement on SpiNNaker2 shows a different kind of
limit. Every input event reaches the neuron up to $512$ simultaneous
inputs, and delivery stops abruptly there: $513$ inputs produce the
deflection of exactly $512$. The limit is the same at every synaptic
weight tested and does not change when the inputs are split across
several source populations, so it is a count of input events per neuron
rather than a bound on current or on the traffic from a single source.
The measurements were made with the default configuration of
py-spinnaker2; whether the limit can be configured, and whether it
bounds events per timestep or connections per neuron, was not
established. For spatial codes the two coincide, since every bit of a
code word is delivered in the same timestep.

\subsubsection{Consequences for the encodings}

Since a code of width $W$ delivers $W$ input events, each limit bounds
the range of values an encoding can represent. Under SpiNNaker's default
buffer of $255$ packets, Thermometer, which needs $N - 1$ bits, is
limited to $256$ values, and Johnson, which needs $N/2$, to $510$; under
SpiNNaker2's limit of $512$ input events per neuron the corresponding
figures are $513$ and $1024$. Binary and Gray need
$\lceil \log_2 N \rceil$ bits and are unconstrained in practice, and the
hybrids lie between the two.

Neither limit is fixed: SpiNNaker's buffer can be enlarged, and
SpiNNaker2's limit appears to be a property of its current software.
What is fixed is the cost: a spatial code requires one packet, one buffer
entry and one synaptic event per bit for every decision, so enlarging
the buffer accommodates a wide code but does not make it cheaper. The
experiments here use the default configuration, under which $N = 256$ is
the widest range at which every encoding fits. This range also contains
the canonical cliff, $127_{10} = 01111111_{2}$ against
$128_{10} = 10000000_{2}$.

\subsection{Encodings Evaluated}
\label{sec:encodings}

Eight encodings are compared, all over the same $N = 256$ values: Binary, Gray, Thermometer, Johnson, and four hybrids.
Their conceptual properties are discussed in Sections~\ref{sec:enc-top} and~\ref{sec:topology}; this section gives the constructions used in the experiments, writing $v$ for the semantic value being encoded and $W$ for the resulting code width.
Table~\ref{tab:cliffpair} shows the code words of the canonical cliff pair, $127$ and $128$, under each.

\subsubsection{Binary}

The standard base-2 representation, $W = \log_2 N = 8$ bits.
Adjacent values may differ in up to $8$ bits; $127$ and $128$ differ in all of them.

\subsubsection{Gray}

The reflected binary code
\begin{equation}
g(v) = v \oplus \lfloor v/2 \rfloor ,
\end{equation}
where $\oplus$ is bitwise exclusive-or, again $8$ bits wide.
Each increment of $v$ flips exactly one bit of $g(v)$, so adjacent values always differ in a single bit; $127$ and $128$ map to $01000000$ and $11000000$.
The guarantee is purely local: values far apart in $v$ may also be one bit apart.

\subsubsection{Thermometer}

A unary code of $N - 1 = 255$ bits in which the first $v$ bits are set.
Writing $t(v)_k$ for the bit in position $k$ of the code word, counted
from the left,
\begin{equation}
t(v)_k = \begin{cases} 1 & k < v \\ 0 & k \geq v \end{cases},
\qquad k = 0, \dots, N-2
\end{equation}
so that $t(v) = 1^{v}0^{\,N-1-v}$, a run of $v$ ones followed by a run of
zeros. The Hamming distance between the code words of two values is exactly
their separation, at the cost of a width linear in $N$.

\subsubsection{Johnson}

A twisted-ring code of $N/2 = 128$ bits with $N$ distinct states. For
$v \leq N/2$ the first $v$ bits are set; beyond that, the set bits are
cleared from the same end at which they were filled, so that the code
returns to all zeros after $N$ steps. Writing $j(v)_k$ for the bit in
position $k$ of the code word, counted from the left,
\begin{equation}
j(v)_k = \begin{cases}
\begin{array}{lll}
1 & k < v, & v \leq N/2, \\
1 & v - N/2 \leq k < N/2, & v > N/2, \\
0 & \text{otherwise}
\end{array}
\end{cases}
\end{equation}
so that $j(v) = 1^{v}0^{\,N/2-v}$ while the field is filling and
$j(v) = 0^{\,v-N/2}1^{\,N-v}$ while it is clearing. With four bits the
sequence is $0000$, $1000$, $1100$, $1110$, $1111$, $0111$, $0011$,
$0001$, and back to $0000$. Filling and clearing must proceed from the
same end: filling from one end and clearing from the other retraces
earlier states and yields only $N/2 + 1$ distinct code words. The Hamming
distance between the code words of two values is the shorter of the two
distances around the ring, so the extremes $0$ and $N-1$ are adjacent.

\begin{table}[t]
\centering
\caption{Code words of the canonical Hamming cliff pair $127$ and $128$. Long
codes are written in run-length form, $1^{a}0^{b}$ denoting $a$ ones
followed by $b$ zeros.}
\label{tab:cliffpair}
\begin{tabular}{lrllr}
\toprule
\textbf{Encoding} & \textbf{Width} & $127$ & $128$ & $H$ \\
\midrule
Binary            & 8   & \texttt{01111111} & \texttt{10000000} & 8 \\
Gray              & 8   & \texttt{01000000} & \texttt{11000000} & 1 \\
Thermometer       & 255 & $1^{127}0^{128}$  & $1^{128}0^{127}$  & 1 \\
Hybrid lin. $B{=}4$ & 65 & $1^{31}0^{34}$    & $1^{32}0^{33}$    & 1 \\
Hybrid lin. $B{=}8$ & 34 & $1^{15}0^{19}$    & $1^{16}0^{18}$    & 1 \\
Johnson           & 128 & $1^{127}0^{1}$    & $1^{128}$         & 1 \\
Hybrid cyc. $B{=}4$ & 34 & $1^{31}0^{3}$     & $1^{32}0^{2}$     & 1 \\
Hybrid cyc. $B{=}8$ & 19 & $1^{15}0^{4}$     & $1^{16}0^{3}$     & 1 \\
\bottomrule
\end{tabular}
\end{table}

\begin{table}[t]
\centering
\caption{The linear hybrid construction at $N = 16$ and $B = 4$, giving a $5$-bit code: three coarse bits for the four blocks and two fine bits for the offset. For each value $v$, $b = \lfloor v/B \rfloor$ is the block it falls in and $r = v \bmod B$ its offset within that block; ``refl.'' is the offset after reflection, equal to $r$ in even blocks and to $B - 1 - r$ in odd ones. ``Coarse'' is the Thermometer code of $b$ and ``fine'' the Gray code of the reflected offset; the two concatenated form the code word. Because of the reflection the fine field runs $00,01,11,10$ through block $0$ and back again through block $1$, so that $\Delta$, the number of bits that change from the previous value, is one everywhere, including at the block boundaries $v = 4$, $8$ and $12$.}
\label{tab:hybridexample}
\begin{tabular}{@{}rrr|r|ll|r@{}}
\toprule
$v$ & $b$ & $r$ & \textbf{refl.} & \textbf{coarse} & \textbf{fine} & $\Delta$ \\
\midrule
 0 & 0 & 0 & 0 & 000 & 00 & --- \\
 1 & 0 & 1 & 1 & 000 & 01 & 1 \\
 2 & 0 & 2 & 2 & 000 & 11 & 1 \\
 3 & 0 & 3 & 3 & 000 & 10 & 1 \\
\addlinespace
 4 & 1 & 0 & 3 & 100 & 10 & 1 \\
 5 & 1 & 1 & 2 & 100 & 11 & 1 \\
 6 & 1 & 2 & 1 & 100 & 01 & 1 \\
 7 & 1 & 3 & 0 & 100 & 00 & 1 \\
\addlinespace
 8 & 2 & 0 & 0 & 110 & 00 & 1 \\
 9 & 2 & 1 & 1 & 110 & 01 & 1 \\
10 & 2 & 2 & 2 & 110 & 11 & 1 \\
11 & 2 & 3 & 3 & 110 & 10 & 1 \\
\addlinespace
12 & 3 & 0 & 3 & 111 & 10 & 1 \\
13 & 3 & 1 & 2 & 111 & 11 & 1 \\
14 & 3 & 2 & 1 & 111 & 01 & 1 \\
15 & 3 & 3 & 0 & 111 & 00 & 1 \\
\bottomrule
\end{tabular}
\end{table}

\subsubsection{Hybrid}

The hybrid encodings split a value into a coarse part and a fine part, and encode each with a different code: a unary-style code for the coarse part (Thermometer for the linear variant, Johnson for the cyclic one), which preserves distance but is wide, and a Gray code for the fine part, which is compact but preserves only local structure.

A value $v$ is divided by the \emph{block size} $B$, giving a \emph{block index} $b = \lfloor v/B \rfloor$ and an \emph{offset} $r = v \bmod B$ within that block. The code word is the concatenation of two disjoint fields:

\begin{itemize}
    \item the \emph{coarse field} encodes which of the $N/B$ blocks the value falls in, using Thermometer for the linear variant or a Johnson ring for the cyclic one;
    \item the \emph{fine field} encodes the offset within the block, as a $\log_2 B$-bit Gray code.
\end{itemize}

Taken naively this reintroduces a cliff at every block boundary, because
the offset would jump from $B-1$ back to $0$ while the coarse field also
changes, costing up to $\log_2 B + 1$ bits between two adjacent values.
The construction therefore \emph{reflects} the offset in odd-numbered
blocks, $r \mapsto B - 1 - r$, so that the offset runs forwards through
one block and backwards through the next. At a boundary the two values
then share the same fine field and differ only in one bit of the coarse
field.

Table~\ref{tab:hybridexample} works this through for a small case.
Reading down the final column, every pair of adjacent values differs in
exactly one bit, at every position in the range including the block
boundaries at $v = 4$, $8$ and $12$.

The cliff pair of Table~\ref{tab:cliffpair} behaves the same way at the
scale used in the experiments: at $B = 8$, the value $127$ sits at the
reflected end of odd block $15$ and $128$ at the start of even block
$16$, so both have fine field $000$ and differ in a single coarse bit.

The construction requires $B$ to be a power of two that divides $N$, so
that the fine field occupies a whole number of bits and the range splits
into $N/B$ whole blocks; with $N$ itself a power of two, both hold for
every such $B$. The code widths are then
\begin{subequations}
\label{eq:widths}
\begin{align}
W_{\text{lin}}(B) &= \frac{N}{B} - 1 + \log_2 B, \label{eq:widthlin}\\[3ex]
W_{\text{cyc}}(B) &=
\begin{cases}
\dfrac{N}{2B} + \log_2 B, & N/B \geq 2,\\[3ex]
\log_2 N, & B = N.
\end{cases} \label{eq:widthcyc}
\end{align}
\end{subequations}
For the cyclic variant the coarse field is a Johnson ring of $N/(2B)$
bits, which has exactly $N/B$ states, one per block. This requires $N/B$
to be even, which it is whenever $N/B \geq 2$, since both are powers of
two. At $B = N$ there is a single block, the coarse field vanishes, and
both variants reduce exactly to Gray. Equation~\eqref{eq:widthlin} gives
that width directly, but a ring cannot be built over a single block, so
for the cyclic variant the endpoint is stated separately. We evaluate
$B = 4$ and $B = 8$ for both variants.

\begin{figure*}[t]
\centering
\begin{subfigure}{0.24\textwidth}\includegraphics[width=\linewidth]{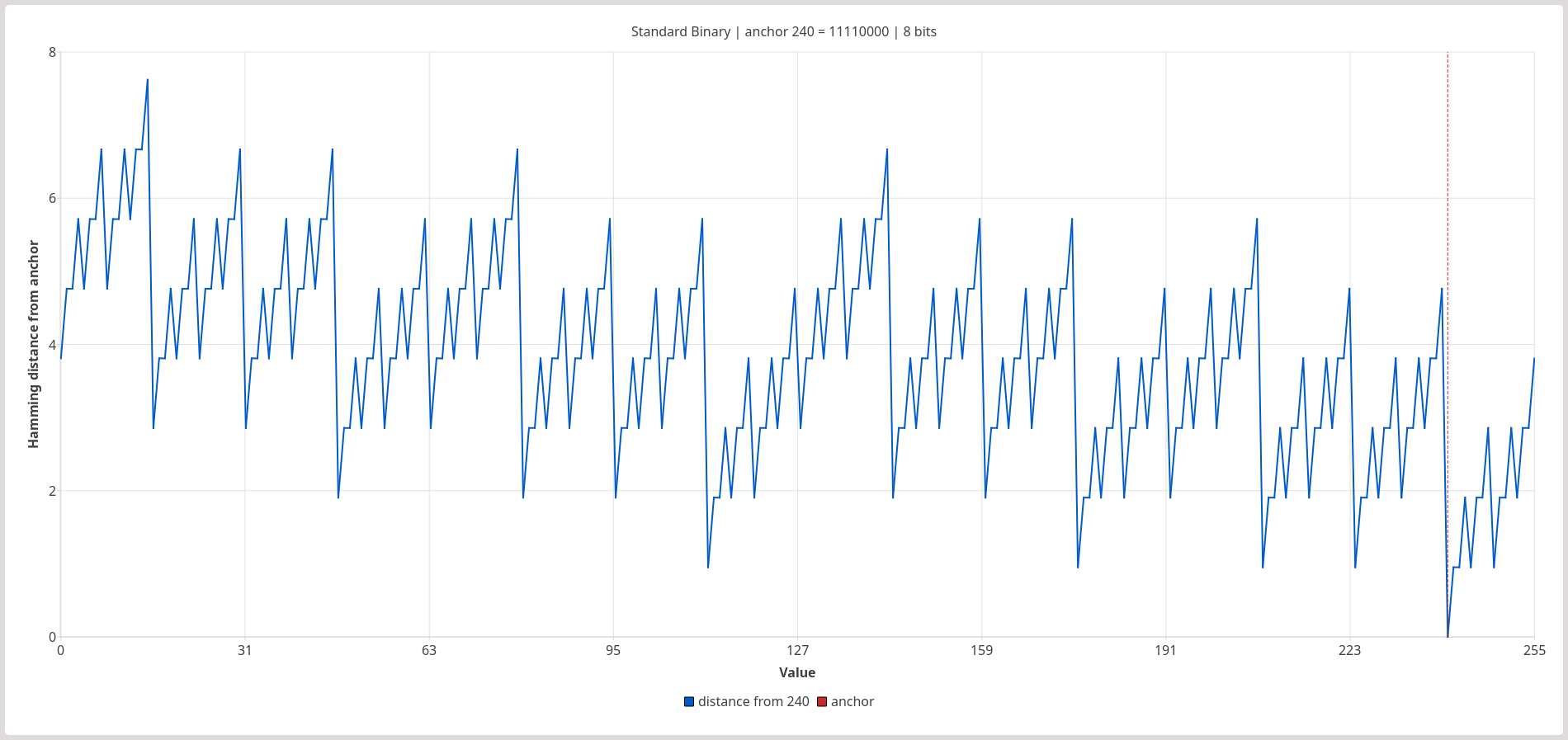}\caption{Binary}\end{subfigure}\hfill
\begin{subfigure}{0.24\textwidth}\includegraphics[width=\linewidth]{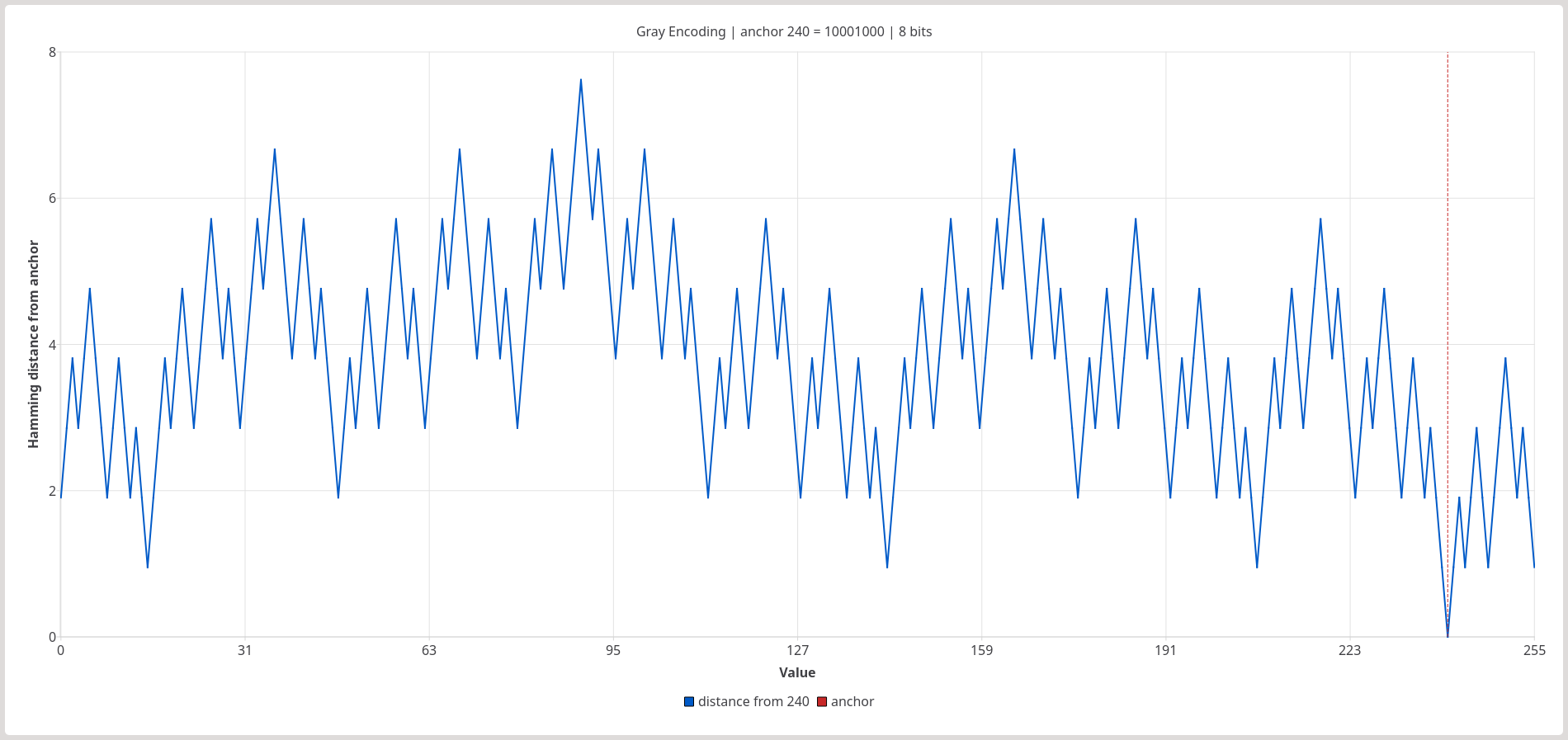}\caption{Gray}\end{subfigure}\hfill
\begin{subfigure}{0.24\textwidth}\includegraphics[width=\linewidth]{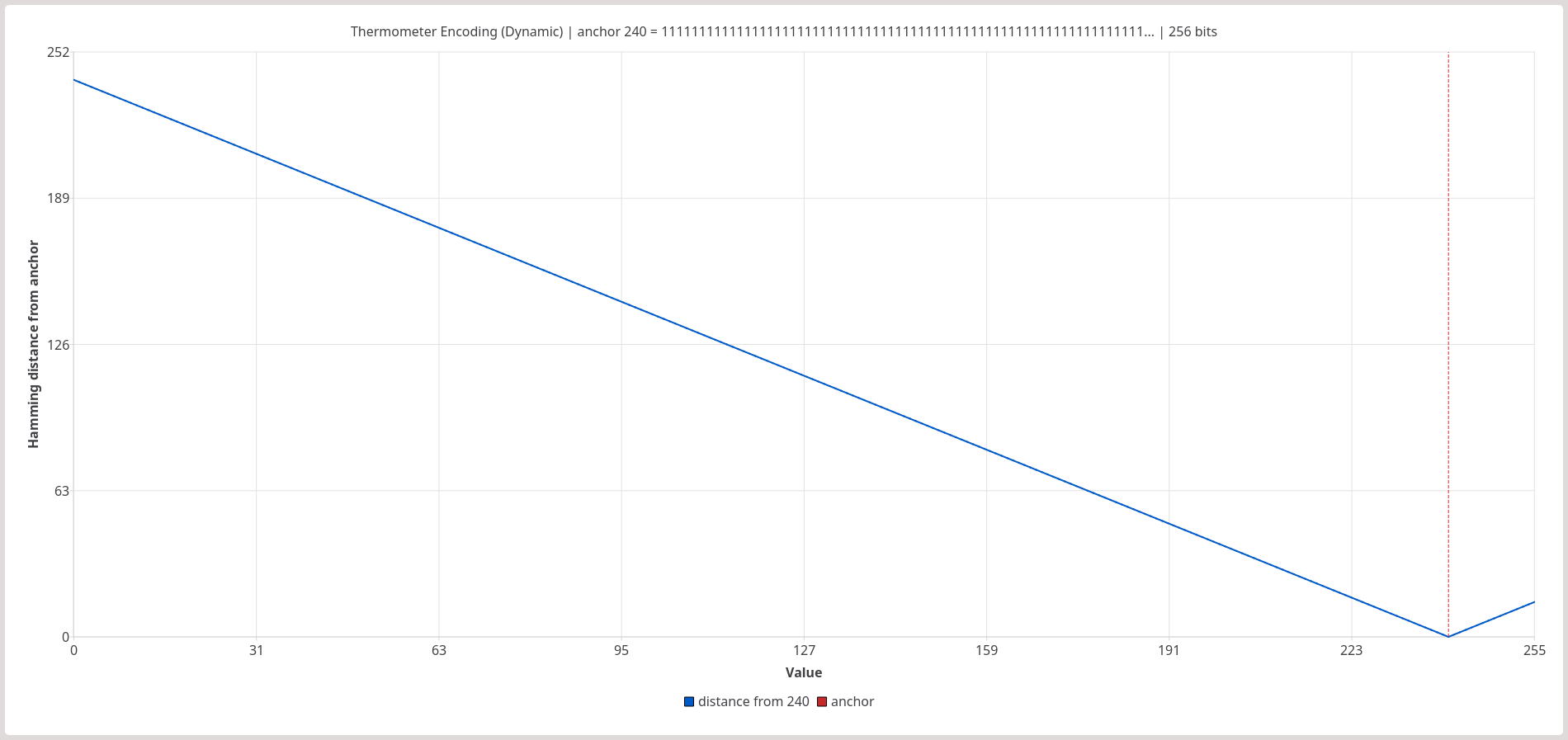}\caption{Thermometer}\end{subfigure}\hfill
\begin{subfigure}{0.24\textwidth}\includegraphics[width=\linewidth]{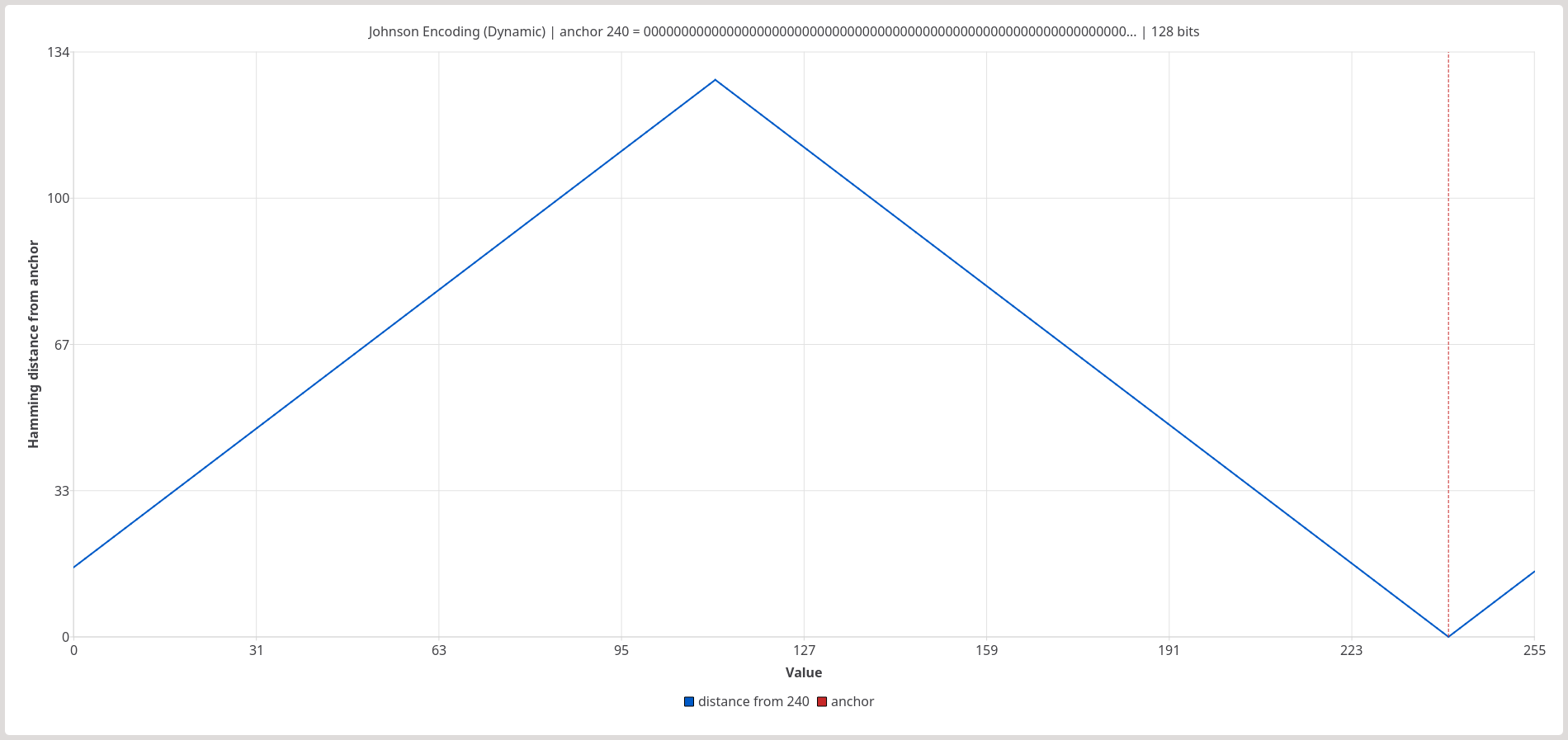}\caption{Johnson}\end{subfigure}

\medskip
\begin{subfigure}{0.24\textwidth}\includegraphics[width=\linewidth]{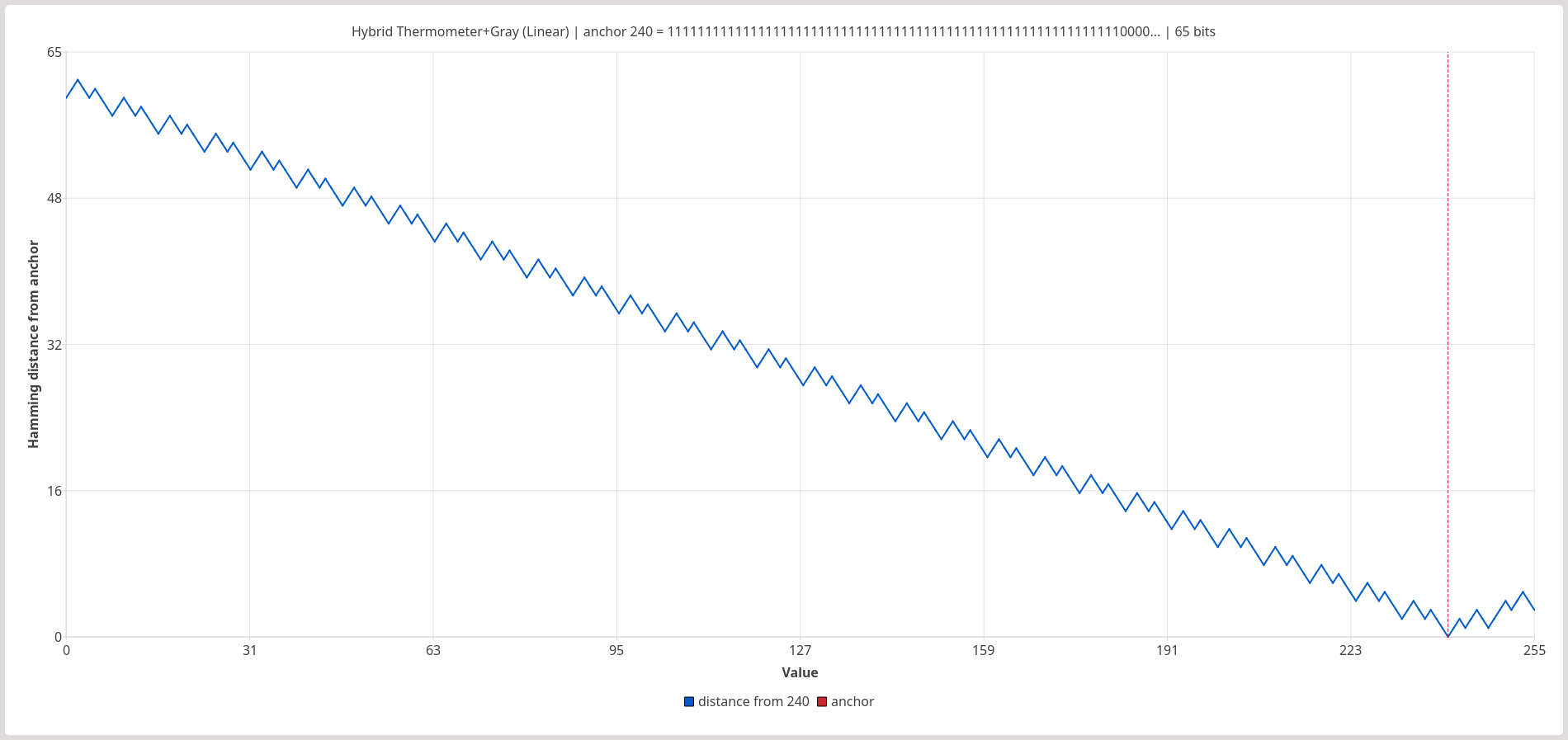}\caption{Hybrid lin.\ $B{=}4$}\end{subfigure}\hfill
\begin{subfigure}{0.24\textwidth}\includegraphics[width=\linewidth]{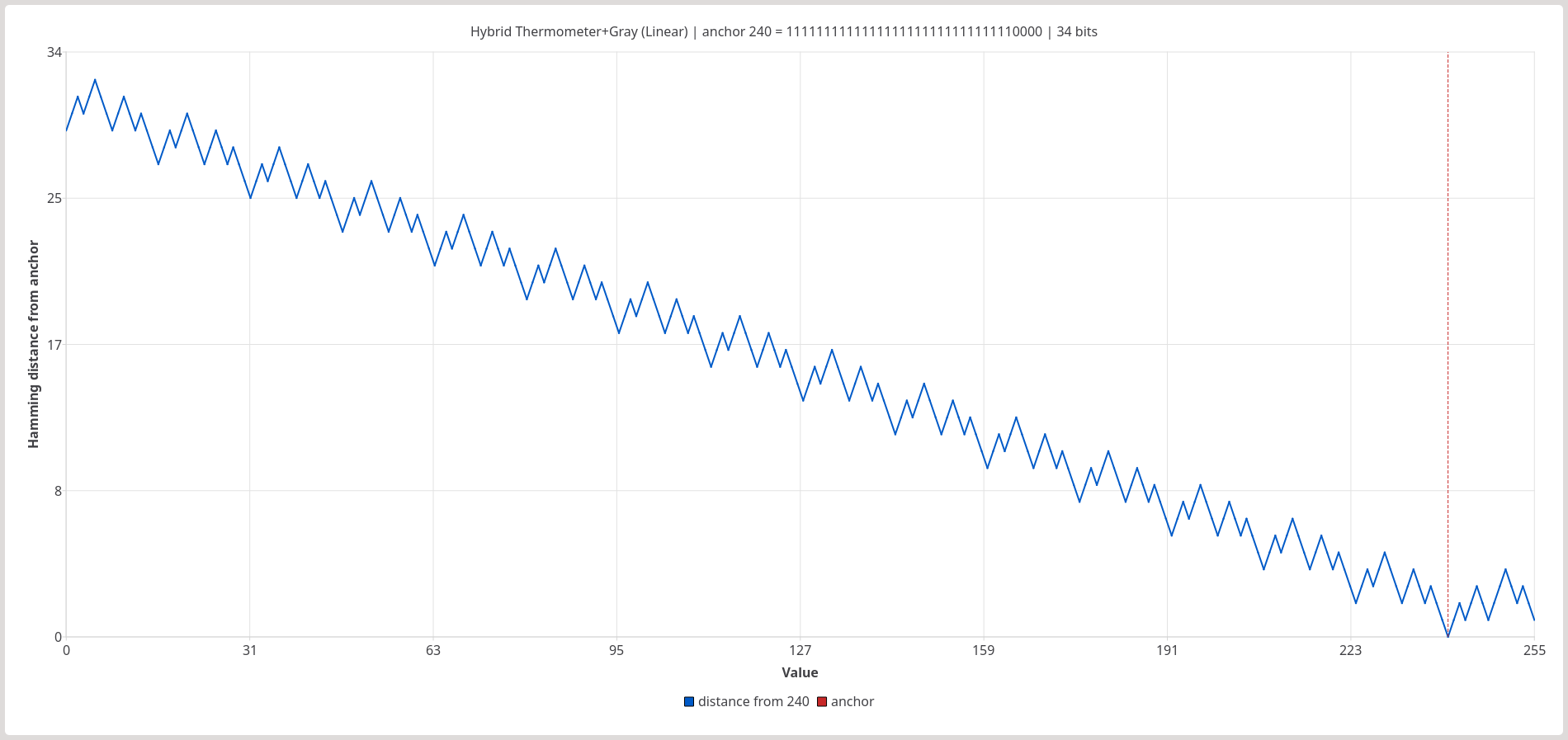}\caption{Hybrid lin.\ $B{=}8$}\end{subfigure}\hfill
\begin{subfigure}{0.24\textwidth}\includegraphics[width=\linewidth]{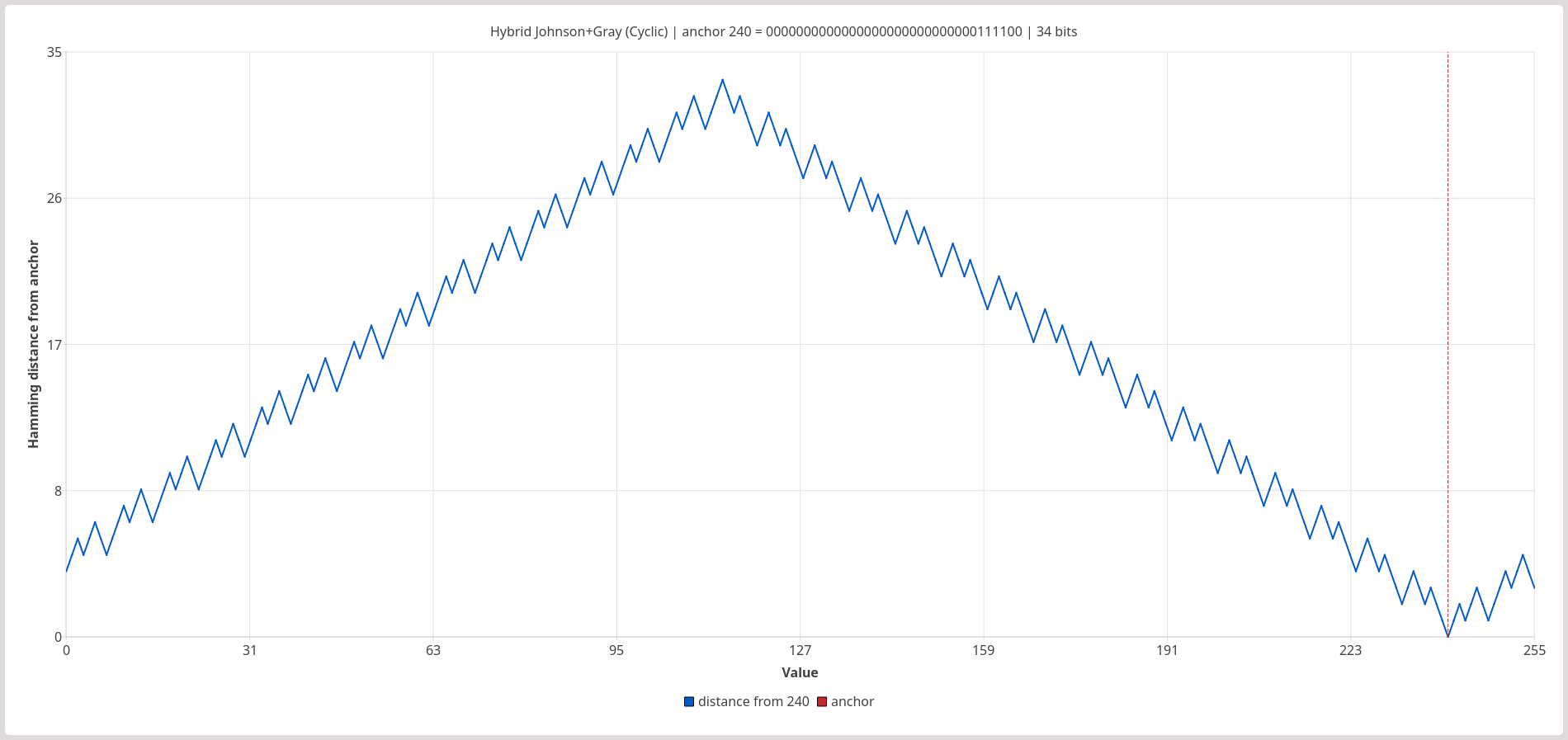}\caption{Hybrid cyc.\ $B{=}4$}\end{subfigure}\hfill
\begin{subfigure}{0.24\textwidth}\includegraphics[width=\linewidth]{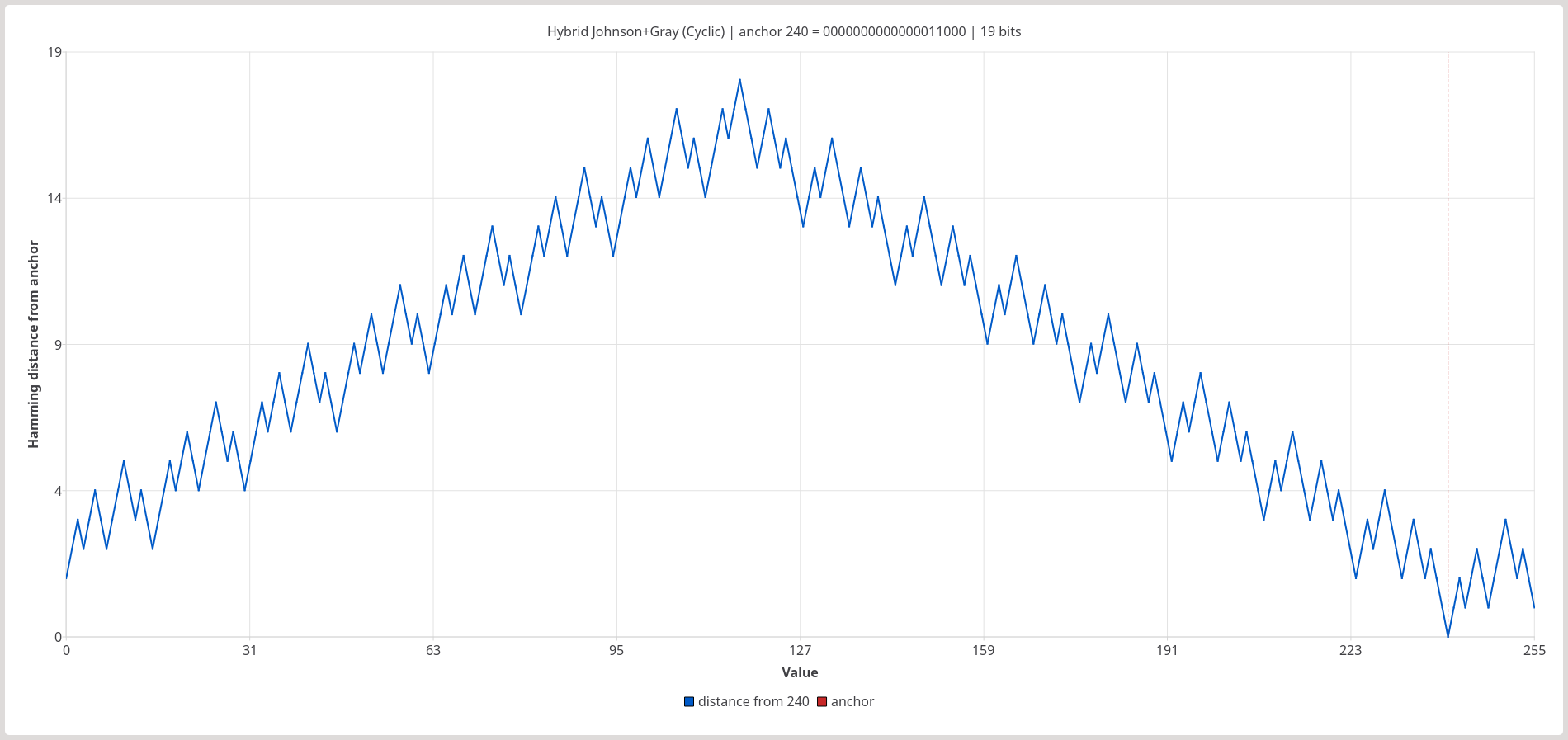}\caption{Hybrid cyc.\ $B{=}8$}\end{subfigure}
\caption{Hamming distance from the anchor value $240$ to every value in
$0$--$255$, under each of the eight encodings. Top row, the established
codes; bottom row, the hybrids derived from them. Vertical scales differ
between panels, since the code widths range from $8$ to $255$ bits: the
comparison is of shape, not of magnitude.}
\label{fig:slices}
\end{figure*}

\begin{figure*}[t]
\centering
\begin{subfigure}{0.24\textwidth}\includegraphics[width=\linewidth]{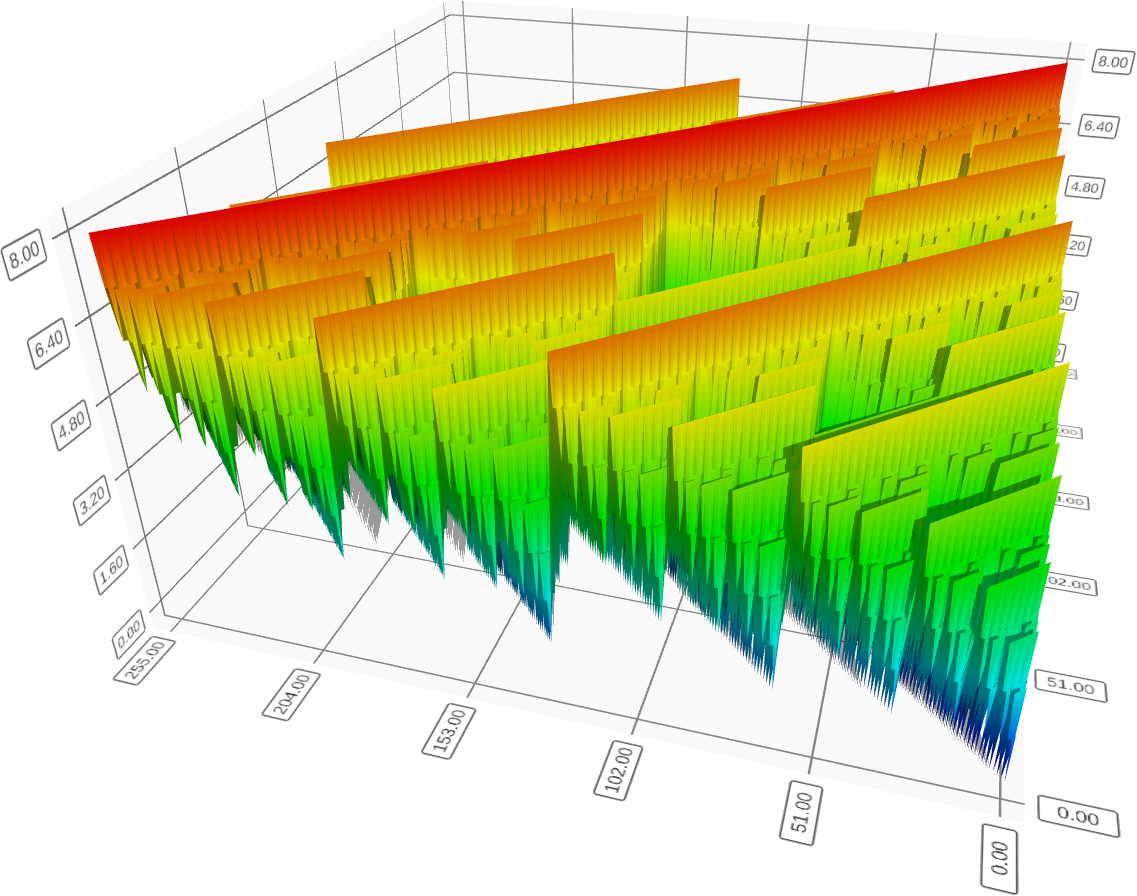}\caption{Binary}\end{subfigure}\hfill
\begin{subfigure}{0.24\textwidth}\includegraphics[width=\linewidth]{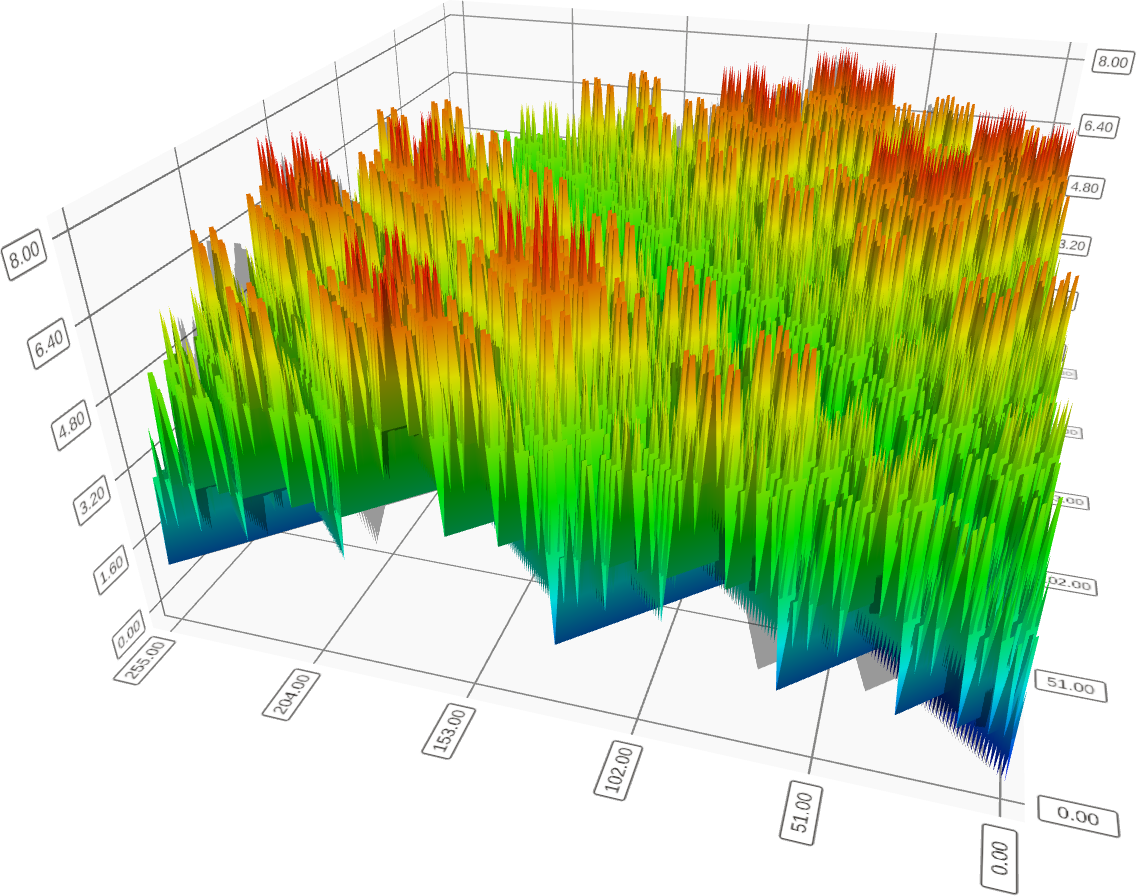}\caption{Gray}\end{subfigure}\hfill
\begin{subfigure}{0.24\textwidth}\includegraphics[width=\linewidth]{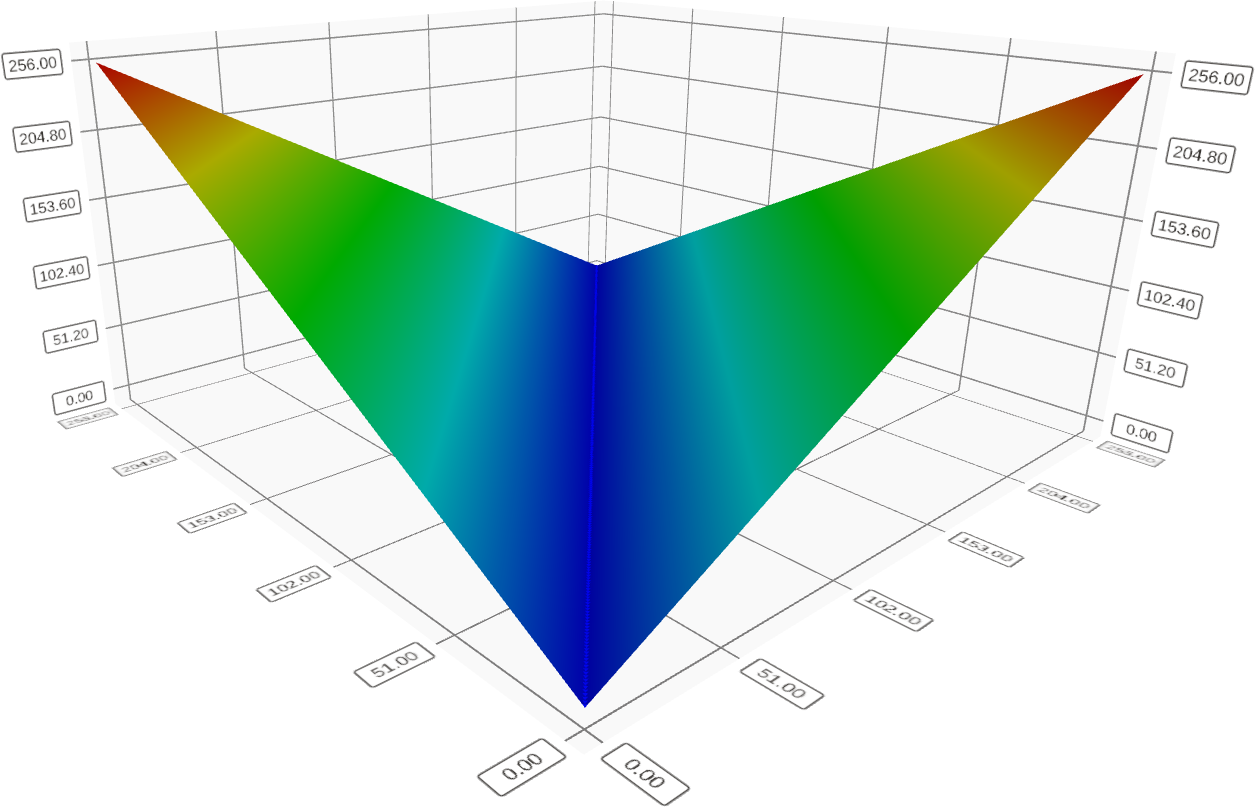}\caption{Thermometer}\end{subfigure}\hfill
\begin{subfigure}{0.24\textwidth}\includegraphics[width=\linewidth]{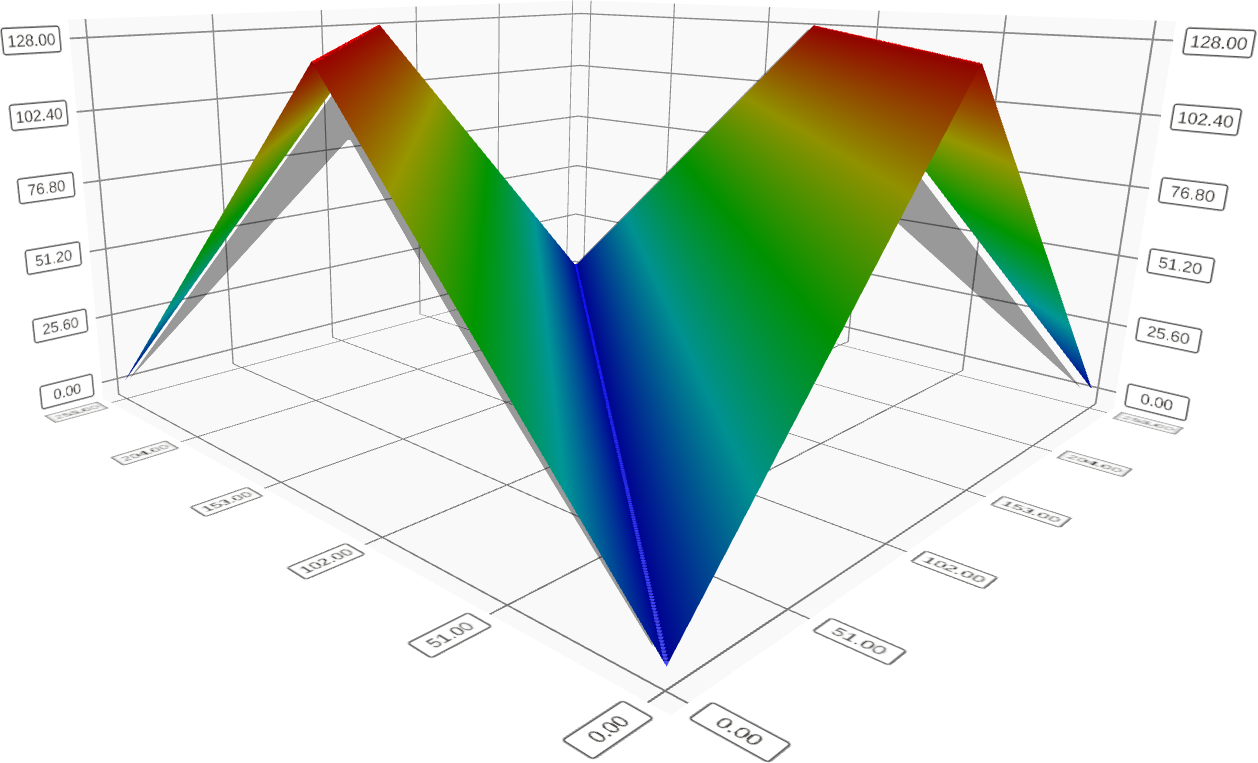}\caption{Johnson}\end{subfigure}

\medskip
\begin{subfigure}{0.24\textwidth}\includegraphics[width=\linewidth]{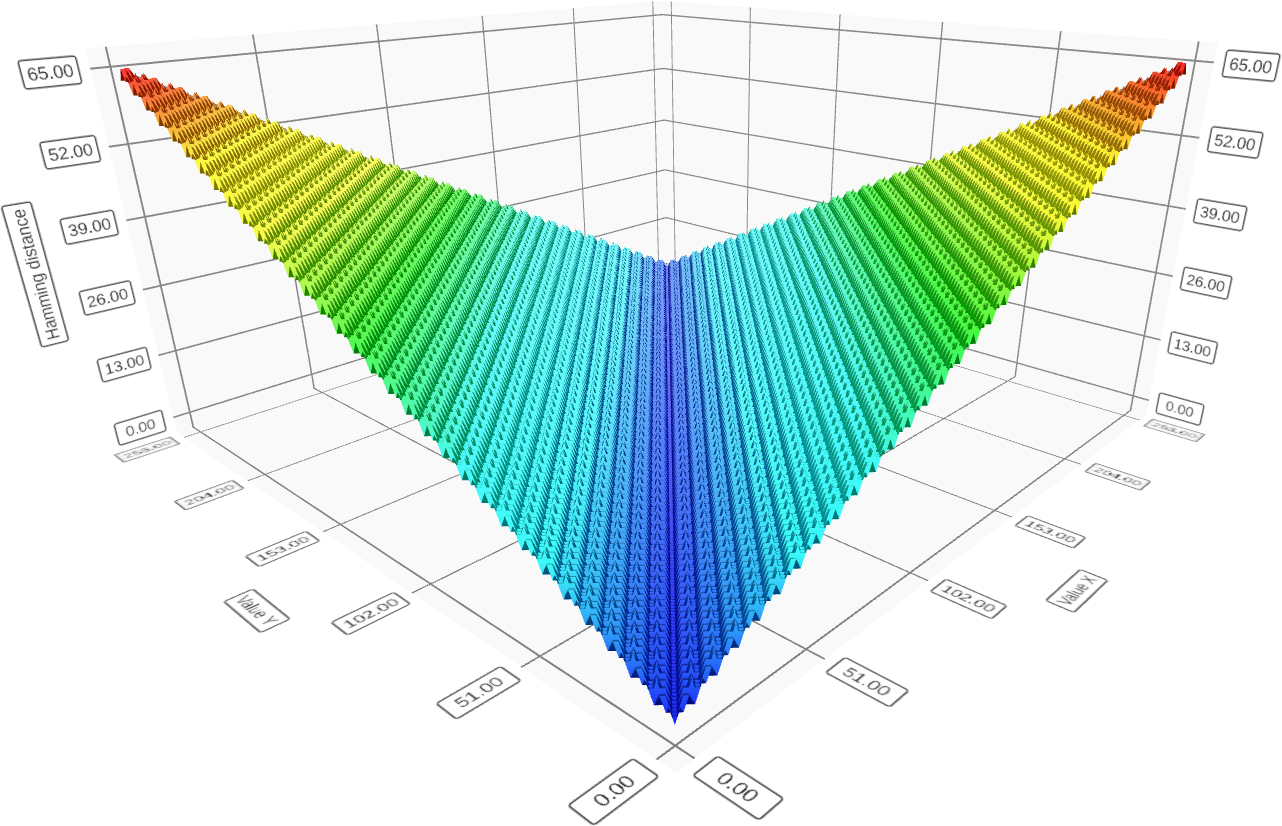}\caption{Hybrid lin.\ $B{=}4$}\end{subfigure}\hfill
\begin{subfigure}{0.24\textwidth}\includegraphics[width=\linewidth]{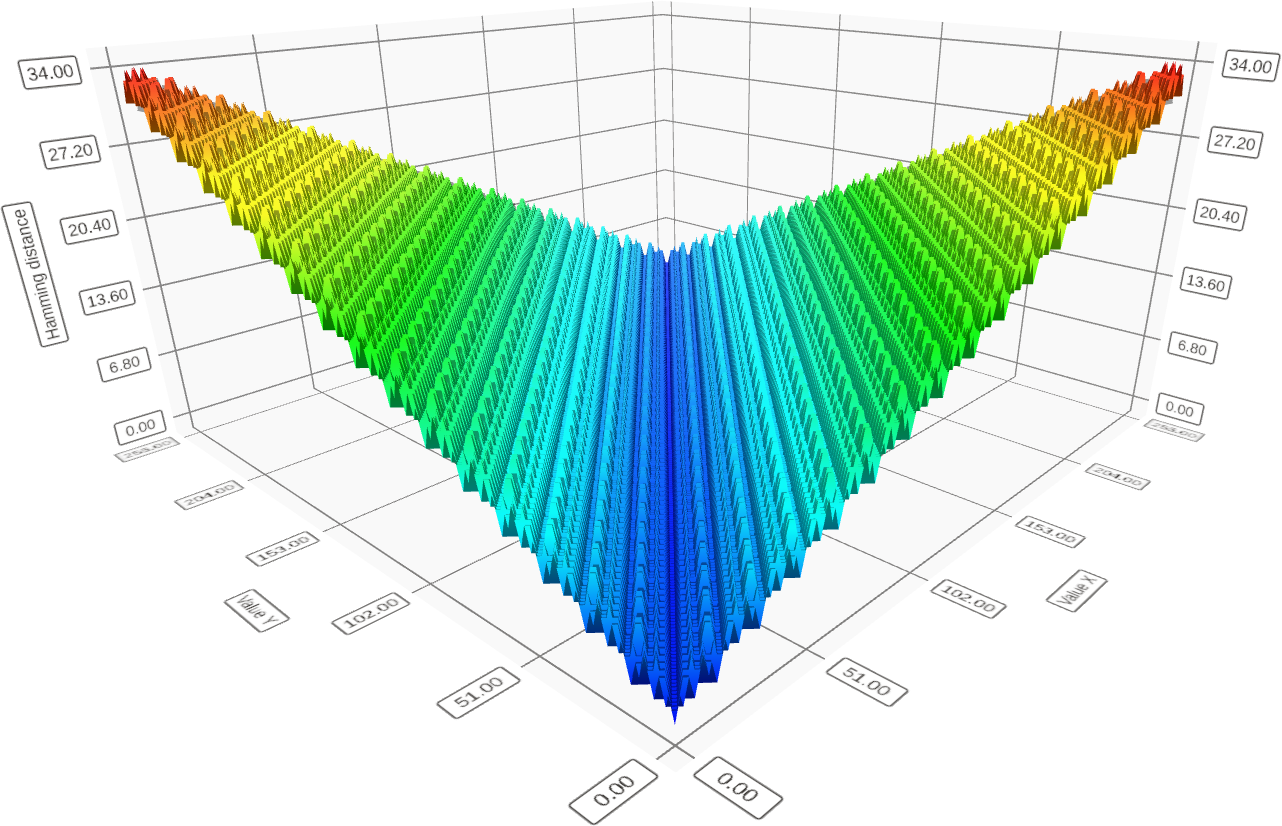}\caption{Hybrid lin.\ $B{=}8$}\end{subfigure}\hfill
\begin{subfigure}{0.24\textwidth}\includegraphics[width=\linewidth]{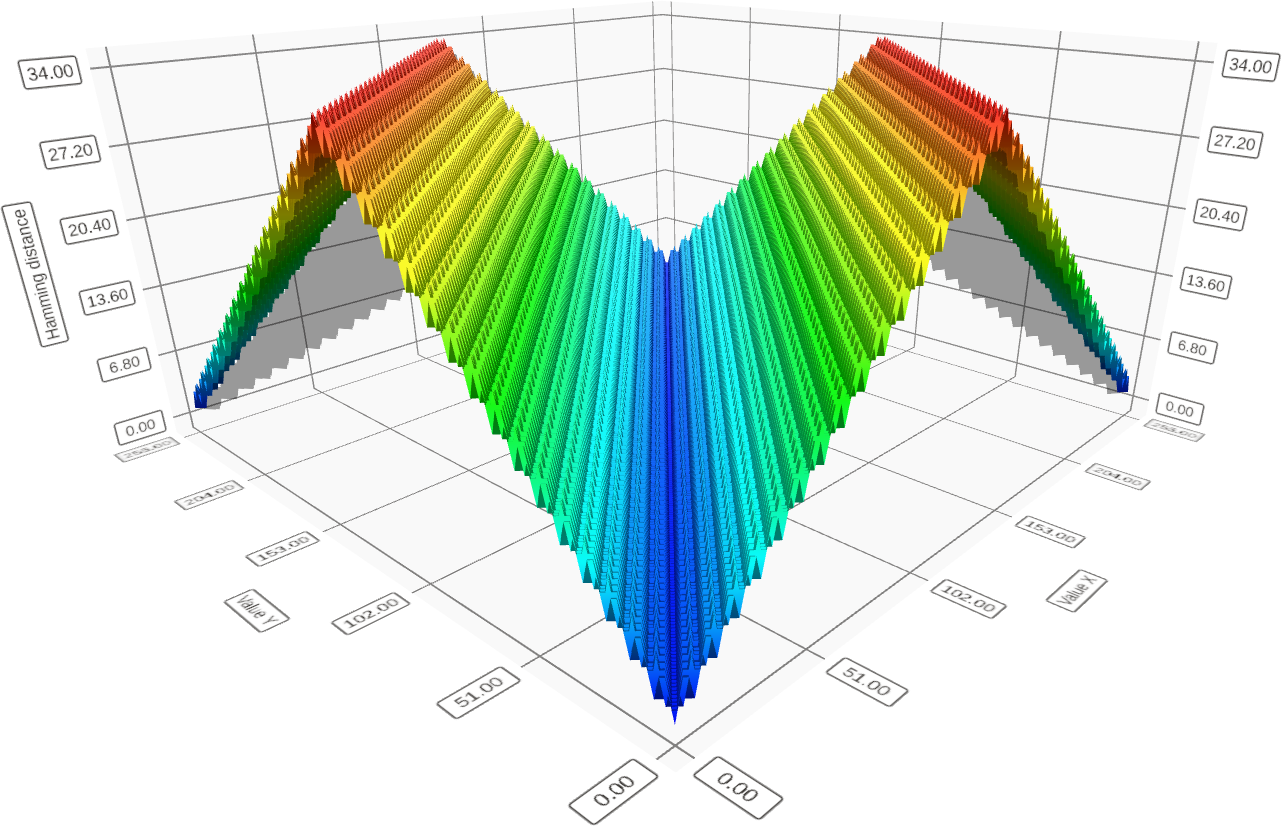}\caption{Hybrid cyc.\ $B{=}4$}\end{subfigure}\hfill
\begin{subfigure}{0.24\textwidth}\includegraphics[width=\linewidth]{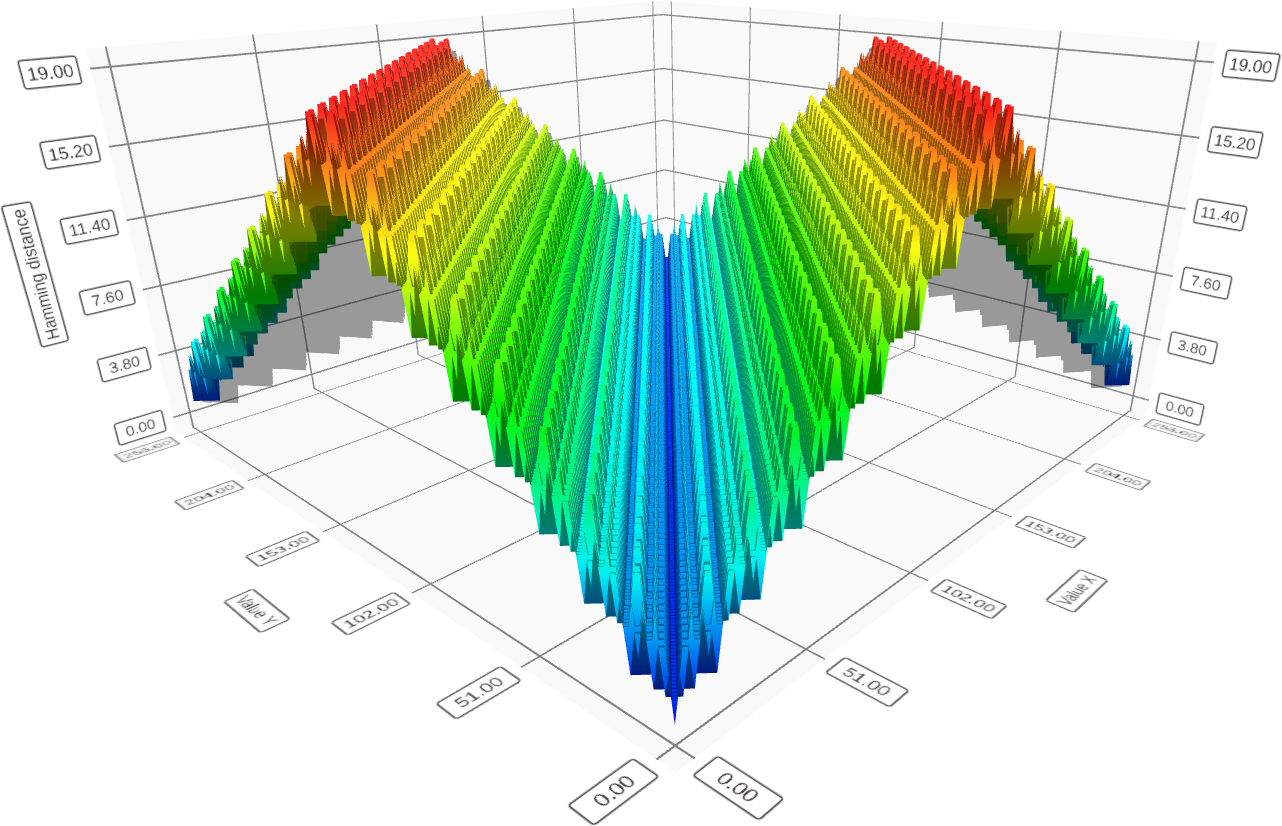}\caption{Hybrid cyc.\ $B{=}8$}\end{subfigure}
\caption{Hamming distance $H(f(i), f(j))$ between every pair of values
in $0$--$255$, under each of the eight encodings. Thermometer is a plane
rising away from the diagonal, since $H = |i-j|$ exactly; Johnson folds
that plane into a ridge, the distance falling again beyond $N/2$ as the
values meet around the ring. The hybrids reproduce those shapes in
stepped form, the step width being the block size. Binary and Gray are
flat and textured: distance is roughly constant away from the diagonal
and carries little information about separation. Vertical scales differ
between panels, since the code widths range from $8$ to $255$ bits.}
\label{fig:surfaces}
\end{figure*}

\subsubsection{Limiting cases}

The block size interpolates between the established encodings.
At $B = 1$ the fine field vanishes and the linear and cyclic hybrids reduce exactly to Thermometer and Johnson encoding.
At $B = N$ the coarse field vanishes and \emph{both} reduce to Gray encoding.
For the cyclic variant this happens earlier, since a Johnson ring of one or two bits is itself a Gray code: from $B = N/4$ onwards the cyclic hybrid has Gray's Hamming distances exactly.
We verified all four identities bit for bit over the $256$ values.
Thermometer, Johnson and Gray are therefore not separate proposals but points on a single continuum, with Gray the common limit of the linear and cyclic families.

Figure~\ref{fig:slices} shows the Hamming distance from the anchor value
$240$ to every other value.
The linear family is V-shaped, reaching its maximum at the far end of the range; Thermometer is strictly monotone on each side of the anchor, while the linear hybrids follow the same ramp at slope $1/B$ with the fine-field ripple superimposed.
The cyclic family rises to a maximum near $v = 112$, the point diametrically opposite the anchor on the ring, then falls back as the values approach $0$, which lies only $16$ steps from $240$ around the ring.
Gray, the limit of both families, shows no systematic trend.

\subsubsection{Width and input activity}

Because every bit costs one spike packet per classification, width is the number of input events per decision.
On event-driven hardware the dynamic energy of a computation grows with the number of events it involves \citep{Furber2012SpiNNaker:Design, Davidson2021ComparisonHardware}, so width bounds the input contribution to that energy; energy itself was not measured here.
Width is also the number of buffer entries each decision occupies on the receiving core (Section~\ref{sec:packetlimit}).

\section{Results}
\label{sec:results}

\subsection{Hamming Structure of the Encodings}
\label{sec:exp1}

Before considering network behaviour we characterise the encodings
themselves, over all $\binom{256}{2}$ value pairs, shown in
Figure~\ref{fig:surfaces}. Hamming distance is the appropriate measure
here because each differing bit corresponds to exactly one synapse whose contribution to the output neuron reverses, from an excitatory $+w$ to an inhibitory $-w$ under the reciprocal wiring of Eq.~\eqref{eq:reciprocal}, so the distance between two code words is the loss of drive they produce, in units of $2w$. Three measures are reported, because no single
statistic captures the property at issue.

\subsubsection{Cliff severity}

Cliff severity is the largest cost of a single semantic step, expressed in units of the smallest distance the code ever puts between two distinct values:
\begin{equation}
\max_{d_S(i,j) = 1} H(c_i, c_j) \Big/
\min_{\substack{k,l \in \mathcal{S}\\ k \neq l}} H(c_k, c_l),
\label{eq:severity}
\end{equation}
following Section~\ref{sec:cliff}, with adjacency taken under each encoding's own metric, so that for the ring codes the pair $N-1$ and $0$ is included. We report it together with the proportion of adjacent pairs that cost more than one bit. Every other encoding considered has a severity of exactly $1$, and none of its adjacent pairs costs more than one bit: no pair of values adjacent under its metric, including the wraparound pair of the ring codes, differs in more than a single bit, by construction.

Binary has a severity of $8$, with $49.8\%$ of adjacent pairs costing
more than one bit. The cliff is therefore not an isolated defect at the
power-of-two boundary but the tail of a geometric distribution, in which
half of all adjacent pairs cost one bit, a quarter cost two, an eighth
three, and so on. Every other encoding considered has a severity of
exactly $1$: no pair of values adjacent under its metric, including the
wraparound pair of the ring codes, differs in more than a single bit, by
construction.

\subsubsection{Global monotonicity}

Global monotonicity is measured by Spearman's rank correlation $\rho_s$
between semantic distance and Hamming distance, taken over all
$\binom{256}{2}$ pairs of values: each pair contributes one observation,
its semantic distance $d_S(i,j)$ and the distance $H(c_i, c_j)$ between
its code words. Spearman's coefficient is the Pearson correlation of the
ranks of the two quantities \citep{Spearman1904TheThings}; it equals $1$
when Hamming distance increases monotonically with semantic distance,
$0$ when the two are unrelated, and it does not assume the relation is
linear, which matters because only Thermometer and Johnson make it so.
Hamming distances are heavily tied, since many pairs of values sit at
the same distance, so the tie-corrected form is used throughout. Each
encoding is scored against its own metric (Section~\ref{sec:proximity}),
$d_{\text{cyc}}$ for the ring codes and $d_{\text{lin}}$ for the rest;
scored against the other, Thermometer and Johnson would both fall from
$1.000$ to $0.747$.

Global monotonicity separates the encodings that cliff severity does not.
Thermometer and Johnson reach $\rho_s = 1.000$ against $d_{\text{lin}}$
and $d_{\text{cyc}}$ respectively; the four hybrids evaluated here retain
between $0.980$ and $0.998$; Binary scores $0.323$ and Gray only $0.374$.
Gray's reflected construction also makes it cyclic, but scored against
$d_{\text{cyc}}$ it does barely better, at $0.392$. Gray therefore removes
the cliff entirely while carrying little information about semantic
separation, the distinction anticipated in Section~\ref{sec:topology}.

Table~\ref{tab:continuum} traces the whole hybrid family on this measure.
The exchange between width and structure is strongly non-linear:
reducing the linear variant from Thermometer's $255$ bits to $34$ at
$B = 8$ lowers $\rho_s$ only from $1.000$ to $0.990$, whereas the last
factor of four in width, from $34$ bits to Gray's $8$, costs almost all
of the remaining structure.

\begin{table}[t]
\centering
\caption{The hybrid family across block sizes at $N = 256$. $\rho_s$ is Spearman's correlation between Hamming and semantic distance, circular for the cyclic variant. $B = 1$ coincides bit for bit with Thermometer and Johnson. From $B = N/2$ both variants coincide bit for bit with Gray, and their $\rho_s$ values differ only because the cyclic variant is scored against circular distance. The cyclic variant reaches Gray's Hamming distances one step earlier: at $B = 64$ its coarse field is a two-bit Johnson ring, which is the two-bit Gray code with its bits swapped.}
\label{tab:continuum}
\resizebox{\columnwidth}{!}{
\begin{tabular}{r rr rr l}
\toprule
& \multicolumn{2}{c}{\textbf{Linear}} & \multicolumn{2}{c}{\textbf{Cyclic}} & \\
$B$ & \textbf{width} & $\rho_s$ & \textbf{width} & $\rho_s$ & \textbf{equivalent encoding} \\
\midrule
1 & 255 & 1.000 & 128 & 1.000 & Thermometer / Johnson \\
2 & 128 & 1.000 & 65 & 0.999 &  \\
4 & 65 & 0.998 & 34 & 0.996 &  \\
8 & 34 & 0.990 & 19 & 0.980 &  \\
16 & 19 & 0.956 & 12 & 0.911 &  \\
32 & 12 & 0.835 & 9 & 0.696 &  \\
64 & 9 & 0.579 & 8 & 0.392 &  \\
128 & 8 & 0.374 & 8 & 0.392& Gray \\
256 & 8 & 0.374 & 8 & 0.392 & Gray \\
\bottomrule
\end{tabular}
}
\end{table}

\subsubsection{Neighbourhood quality}

\begin{table}[t]
\centering
\caption{Neighbourhood fragmentation at four radii: the mean number of
disjoint runs of consecutive values in the Hamming ball around each
anchor, averaged over all $256$ anchors. The radius is a fraction of the
code width, with a floor of one bit; for the $8$-bit codes the three
smaller fractions all fall below one bit, so their columns coincide. The
text quotes the $10\%$ column.}
\label{tab:neighbourhood}
\begin{tabular}{lrrrr}
\toprule
& \multicolumn{4}{c}{\textbf{Radius, fraction of width}} \\
\cmidrule(l){2-5}
\textbf{Encoding} & $2\%$ & $5\%$ & $10\%$ & $20\%$ \\
\midrule
Binary & 7.5 & 7.5 & 7.5 & 23.9 \\
Gray & 7.0 & 7.0 & 7.0 & 21.1 \\
\addlinespace
Thermometer & 1.0 & 1.0 & 1.0 & 1.0 \\
Hybrid lin. $B{=}4$ & 3.0 & 3.9 & 3.7 & 3.4 \\
Hybrid lin. $B{=}8$ & 3.9 & 6.8 & 7.6 & 6.7 \\
\addlinespace
Johnson & 1.0 & 1.0 & 1.0 & 1.0 \\
Hybrid cyc. $B{=}4$ & 3.0 & 4.0 & 4.0 & 4.0 \\
Hybrid cyc. $B{=}8$ & 4.0 & 4.0 & 7.0 & 8.0 \\
\bottomrule
\end{tabular}
\end{table}

Neighbourhood quality asks what a code's local region contains. Around
each anchor we take the \emph{Hamming ball} of radius $\alpha W$, with a
floor of one bit, meaning every code word within that many bits of the
anchor, and count how many disjoint runs of consecutive values it holds;
the result is averaged over all anchors. The radius is chosen to match
the network's behaviour: by the tolerance law of Section~\ref{sec:exp5},
a ball of radius $\alpha W$ is what the network accepts at a gain of
$1/(1 - 2\alpha)$ times its minimum. The text quotes $\alpha = 10\%$,
which corresponds to a gain of $1.25$, at the low end of the range swept
on hardware, where the network has only just begun to tolerate mismatch.

A Hamming ball around a Gray code word contains about nine values
scattered across seven disjoint runs and reaching $144$ semantic units
from the anchor, behaviour essentially indistinguishable from Binary.
The same ball under Thermometer is a single contiguous interval.
Table~\ref{tab:neighbourhood} repeats the measure at $2\%$, $5\%$ and
$20\%$, and the ordering of the encodings is the same at every radius:
Thermometer and Johnson give a single run, the hybrids at most $B$ runs,
and Binary and Gray seven or more, rising above twenty at $20\%$.

\subsubsection{Summary}

Cliff severity separates Binary from everything else, but no further:
all seven remaining encodings are equally clean by that measure. Global
monotonicity and neighbourhood quality then separate those seven, and
they do so in the same direction: Gray's Hamming balls are as fragmented
as Binary's despite its perfect adjacency. Removing the Hamming cliff is
therefore necessary for the clustering behaviour sought here, but it is
not sufficient.

\subsection{Two Trained Values}
\label{sec:exp2}

\begin{table*}[t]
\centering
\caption{The fourteen second values of Section~\ref{sec:exp2} and their Hamming distance from the anchor $240$ under each encoding, with their semantic separation $|v_2 - 240|$ for comparison. The groups are the close neighbours above the anchor, those below, progressively distant values, and the top of the range. Thermometer reproduces the separation exactly and Johnson its circular equivalent; the hybrids scale it by $1/B$; Binary and Gray bear no consistent relation to it. The code words of every value are listed in Table~\ref{tab:sweepcodes}.}
\label{tab:sweep}
\begin{tabular}{rr rrrrrrrr}
\toprule
& & \multicolumn{8}{c}{\textbf{Hamming distance from the anchor $240$}} \\
\cmidrule(l){3-10}
\textbf{Second} & \textbf{Semantic} & & & \multicolumn{3}{c}{\textbf{Linear family}}
  & \multicolumn{3}{c}{\textbf{Cyclic family}} \\
\cmidrule(lr){5-7}\cmidrule(l){8-10}
\textbf{value} $v_2$ & \textbf{separation} & \textbf{Binary} & \textbf{Gray} & \textbf{Thermometer} & {$B{=}4$} & {$B{=}8$}
  & \textbf{Johnson} & {$B{=}4$} & {$B{=}8$} \\
  \midrule
\multicolumn{2}{l}{\textbf{anchor $240$}}
  & $1^{4}0^{4}$ & $10^{3}10^{3}$ & $1^{240}0^{15}$ & $1^{60}0^{5}$
  & $1^{30}0^{4}$ & $0^{112}1^{16}$ & $0^{28}1^{4}0^{2}$ & $0^{14}1^{2}0^{3}$ \\
\midrule
241 & 1 & 1 & 1 & 1 & 1 & 1 & 1 & 1 & 1 \\
242 & 2 & 1 & 2 & 2 & 2 & 2 & 2 & 2 & 2 \\
244 & 4 & 1 & 2 & 4 & 2 & 2 & 4 & 2 & 2 \\
248 & 8 & 1 & 2 & 8 & 2 & 2 & 8 & 2 & 2 \\
\addlinespace
239 & 1 & 5 & 1 & 1 & 1 & 1 & 1 & 1 & 1 \\
238 & 2 & 4 & 2 & 2 & 2 & 2 & 2 & 2 & 2 \\
236 & 4 & 3 & 2 & 4 & 2 & 2 & 4 & 2 & 2 \\
232 & 8 & 2 & 2 & 8 & 2 & 2 & 8 & 2 & 2 \\
\addlinespace
224 & 16 & 1 & 2 & 16 & 4 & 2 & 16 & 4 & 2 \\
192 & 48 & 2 & 2 & 48 & 12 & 6 & 48 & 12 & 6 \\
128 & 112 & 3 & 2 & 112 & 28 & 14 & 112 & 28 & 14 \\
64 & 176 & 3 & 4 & 176 & 44 & 22 & 80 & 20 & 10 \\
0 & 240 & 4 & 2 & 240 & 60 & 30 & 16 & 4 & 2 \\
\addlinespace
255 & 15 & 4 & 1 & 15 & 3 & 1 & 15 & 3 & 1 \\
\bottomrule
\end{tabular}
\end{table*}

\begin{table}[t]
\centering
\caption{Two trained values: the anchor $240$ together with each of the fourteen second values of Section~\ref{sec:exp2} and Table~\ref{tab:sweep}. If the network accepted exactly the code words lying between the two trained ones, the number of false positives would be $H - 1$, where $H$ is the Hamming distance between them. ``$f_p = H - 1$'' counts how many of the fourteen pairs meet this exactly; ``excess'' is the largest number of false positives above $H - 1$ for any pair; ``total $f_p$'' and ``mean acc.'' are summed and averaged over the fourteen pairs.}
\label{tab:exp2}
\resizebox{\columnwidth}{!}{
\begin{tabular}{lrrrrr}
\toprule
\textbf{Encoding} & \textbf{Width} & $f_p=H-1$ & \textbf{excess} & \textbf{total} $f_p$ & \textbf{mean acc.} \\
\midrule
Binary & 8 & 5/14 & +26 & 93 & 0.9741 \\
Gray & 8 & 3/14 & +6 & 29 & 0.9919 \\
Thermometer & 255 & 10/14 & +2 & 630 & 0.8242 \\
Hybrid lin. $B{=}4$ & 65 & 10/14 & +1 & 155 & 0.9568 \\
Hybrid lin. $B{=}8$ & 34 & 8/14 & +1 & 81 & 0.9774 \\
Johnson & 128 & 14/14 & +0 & 303 & 0.9155 \\
Hybrid cyc. $B{=}4$ & 34 & 10/14 & +1 & 75 & 0.9791 \\
Hybrid cyc. $B{=}8$ & 19 & 8/14 & +1 & 41 & 0.9886 \\
\bottomrule
\end{tabular}
}
\end{table}

The network was trained on the anchor together with a second value $v_2$, and tested exhaustively over all $256$ values. Fourteen second values were used, listed in Table~\ref{tab:sweep}, chosen from the binary form of the anchor, $240_{10} = 11110000_2$. The first four, $241$, $242$, $244$ and $248$, set one of its four zero bits each, so they lie one bit from the anchor in Binary but $1$, $2$, $4$ and $8$ units from it in value. The next four, $239$, $238$, $236$ and $232$, lie at the same semantic separations below the anchor, where Binary places them up to five bits away; the two groups therefore compare equal semantic distances under very different binary Hamming distances. Five further values, $224$, $192$, $128$, $64$ and $0$, extend the separation across the rest of the range, and $255$ tests the top of the range.

Table~\ref{tab:sweep} also shows why the sweep is informative: the same pair of values can be one bit apart under one encoding and many bits apart under another. Under Thermometer encoding the Hamming distance is the semantic separation exactly, and under the hybrids it is that separation scaled by about $1/B$, with a ripple of up to $\log_2 B$ bits from the fine field that dominates only at small separations. Johnson encoding gives its circular equivalent, so that the value $0$, which lies $240$ steps from the anchor, is only $16$ bits away. Gray encoding, by contrast, stays between one and four bits across the whole range, which is the $\rho_s = 0.374$ of Section~\ref{sec:exp1} in concrete form.

Table~\ref{tab:exp2} summarises the outcome and Table~\ref{tab:exp2full} gives the complete metrics. The false positives are not arbitrary. If the network accepted precisely the code words lying between the two trained ones, their number would be the Hamming distance between the two trained code words minus one, $f_p = H - 1$. Johnson meets this exactly for all fourteen pairs, and Thermometer and the hybrids for eight to ten, exceeding it by one or two in the rest. For Thermometer trained on $240$ and $0$, the $239$ accepted values are exactly the integers between them, consistent with the distance of $240$ recorded in Table~\ref{tab:sweep}. They are neighbouring activations rather than errors: the network has generalised over the interval its two trained values span.

Binary does not obey the relation. At $v_2 = 239$, a Hamming distance of $5$, it produces $30$ false positives, an excess of $26$; and since $239$ and $240$ are adjacent, no value lies between them in the range, so none of the $30$ is a neighbouring activation.

This inverts the apparent ranking by accuracy. Thermometer records the lowest mean accuracy over the fourteen pairs ($0.824$) precisely because it generalises over the largest contiguous intervals, while Binary's higher figure ($0.974$) conceals scattered errors. Accuracy is therefore a misleading summary here, and the relation between false positives and Hamming distance is the more informative quantity.

\subsubsection{A systematic effect of bit count}
\label{sec:bitcount}

Under Binary, the required scaling factor varies slightly between second values at equal Hamming distance, and the variation is not noise. It depends on the number of $1$-bits in the second code word and not on their position. At $H = 1$ the second value differs from the anchor $240_{10} = 11110000_2$ in a single bit. The four values that do so by setting one of its zero bits, $241$, $242$, $244$ and $248$, each have five set bits, and all require exactly $2.96051$, although the flipped bit occupies a different position in each. The remaining one-bit neighbour, $224$, clears a set bit instead and so has three, and it requires $2.95312$ (the search records the factor to five decimals; the tables report it rounded up to two). At $H = 4$ the factor rises monotonically with bit count, from $5.13955$ for $0$ (no set bits) through $5.18480$ for $238$ (six) to $5.19773$ for $255$ (eight). Within any Hamming-distance group the spread is at most $1.1\%$.

This follows from the dual-rail training protocol. Bits equal to $1$ and bits equal to $0$ are presented at different times relative to the teacher spike (Section~\ref{sec:architecture}), so their synapses are potentiated at different points on the STDP curve and need not acquire identical weights. A code word with more set bits therefore draws proportionally more of its drive from one rail and requires marginally more gain to reach threshold. The effect is an order of magnitude smaller than the differences between encodings reported here and does not affect any comparison, but it accounts for the residual variation within a Hamming-distance group, and it reappears in Section~\ref{sec:exp5} where the tolerance threshold falls exactly on a boundary between Hamming distances.

\subsection{Robustness as the Trained Set Grows}
\label{sec:exp3}

\begin{table*}[t]
\centering
\caption{A contiguous cluster around 240, grown from one to
eight trained values. Top: homeostatic factor. Middle: false positives over
all 256 values. Bottom: the factor divided by $f(1)/n$, the value coherent
summation would require -- $1.00$ means the summed weights reinforce
perfectly, and the excess is the ``don't care'' cancellation. Absolute
factors are not comparable between encodings, because the output threshold
is scaled with code width (Section~\ref{sec:architecture}): 8-bit codes
sit near 5 and wide codes near 2 for that reason alone. Compare trends
within a row, or the normalised values in the bottom block.}
\label{tab:exp3}
\begin{tabular}{lrrrrrr}
\toprule
\textbf{Encoding} & $n{=}1$ & $n{=}2$ & $n{=}3$ & $n{=}4$ & $n{=}6$ & $n{=}8$ \\
\midrule
\multicolumn{7}{l}{\textbf{\emph{Homeostatic factor}}} \\
Binary & 5.18 & 6.94 & 6.95 & 3.48 & 2.32 & 1.74 \\
Gray & 5.17 & 2.96 & 2.30 & 1.73 & 1.39 & 1.04 \\
Thermometer & 2.47 & 1.24 & 0.84 & 0.63 & 0.43 & 0.32 \\
Hybrid lin. $B{=}4$ & 2.15 & 1.09 & 0.74 & 0.56 & 0.38 & 0.29 \\
Hybrid lin. $B{=}8$ & 2.11 & 1.09 & 0.75 & 0.57 & 0.39 & 0.29 \\
Johnson & 2.24 & 1.14 & 0.76 & 0.58 & 0.39 & 0.30 \\
Hybrid cyc. $B{=}4$ & 2.11 & 1.09 & 0.75 & 0.57 & 0.39 & 0.29 \\
Hybrid cyc. $B{=}8$ & 2.19 & 1.16 & 0.82 & 0.62 & 0.44 & 0.33 \\
\midrule
\multicolumn{7}{l}{\textbf{\emph{False positives}}} \\
Binary & 0 & 30 & 46 & 28 & 26 & 24 \\
Gray & 0 & 0 & 0 & 0 & 0 & 0 \\
Thermometer & 0 & 0 & 4 & 2 & 6 & 0 \\
Hybrid lin. $B{=}4$ & 0 & 0 & 0 & 0 & 2 & 8 \\
Hybrid lin. $B{=}8$ & 0 & 0 & 0 & 0 & 0 & 0 \\
Johnson & 0 & 0 & 0 & 0 & 0 & 0 \\
Hybrid cyc. $B{=}4$ & 0 & 0 & 0 & 0 & 0 & 0 \\
Hybrid cyc. $B{=}8$ & 0 & 0 & 0 & 0 & 0 & 0 \\
\midrule
\multicolumn{7}{l}{\textbf{\emph{Factor relative to coherent summation}}} \\
Binary & 1.00 & 2.68 & 4.03 & 2.69 & 2.69 & 2.69 \\
Gray & 1.00 & 1.15 & 1.33 & 1.34 & 1.61 & 1.61 \\
Thermometer & 1.00 & 1.00 & 1.02 & 1.02 & 1.04 & 1.04 \\
Hybrid lin. $B{=}4$ & 1.00 & 1.01 & 1.03 & 1.04 & 1.06 & 1.08 \\
Hybrid lin. $B{=}8$ & 1.00 & 1.03 & 1.07 & 1.08 & 1.11 & 1.10 \\
Johnson & 1.00 & 1.02 & 1.02 & 1.04 & 1.04 & 1.07 \\
Hybrid cyc. $B{=}4$ & 1.00 & 1.03 & 1.07 & 1.08 & 1.11 & 1.10 \\
Hybrid cyc. $B{=}8$ & 1.00 & 1.06 & 1.12 & 1.13 & 1.21 & 1.21 \\
\bottomrule
\end{tabular}
\end{table*}

\begin{table}[t]
\centering
\caption{Semantic distance of the false positives from the
trained cluster, pooled over all trained-set sizes. ``within 4'' is the
share falling within four semantic units -- the network generalising over a
contiguous zone, as opposed to firing on unrelated values. Encodings with
no false positives are marked ``--''.}
\label{tab:exp3geom}
\resizebox{\columnwidth}{!}{
\begin{tabular}{lrrrr}
\toprule
\textbf{Encoding} & \textbf{total} $f_p$ & \textbf{median dist.} & \textbf{max dist.} & \textbf{within 4} \\
\midrule
Binary & 154 & 8.9 & 142 & 25\% \\
Gray & -- & -- & -- & -- \\
Thermometer & 12 & 1.7 & 3 & 100\% \\
Hybrid lin. $B{=}4$ & 10 & 2.2 & 4 & 100\% \\
Hybrid lin. $B{=}8$ & -- & -- & -- & -- \\
Johnson & -- & -- & -- & -- \\
Hybrid cyc. $B{=}4$ & -- & -- & -- & -- \\
Hybrid cyc. $B{=}8$ & -- & -- & -- & -- \\
\bottomrule
\end{tabular}
}
\end{table}

A contiguous cluster of values around $240$ was trained, growing from one to eight members, reproducing and extending the progression of \citet{Davies2025SupervisedSpiNNaker}.

Tables~\ref{tab:exp3} and~\ref{tab:exp3geom} report the results, and Table~\ref{tab:exp3full} the complete metrics. Binary degrades as that work predicts, and no other encoding degrades in the same way: the few untrained values the other encodings accept lie immediately beside the trained cluster. From two trained values onward Binary produces between $24$ and $46$ false positives, while Gray, Johnson and three of the four hybrids accept no untrained value at any trained-set size.

The homeostatic factor exposes the mechanism. Because the weight vectors are summed, an encoding whose trained code words reinforce coherently should require a factor of $f(1)/n$ for $n$ trained values, where $f(1)$ is the factor required by a single trained value. Thermometer, Johnson and the hybrids stay close to this, exceeding it by at most $4$ to $21\%$ depending on the encoding, whereas Binary requires up to $4.03$ times that value, at $n = 3$. The excess is the ``don't care'' cancellation of Section~\ref{sec:testing}, measured directly: where trained code words disagree, their summed weights cancel, and the gain that compensates for the loss admits the false positives.

Gray, though also free of the cliff, does not stay at the coherent value: its factor rises to $1.61$ times $f(1)/n$ at $n = 8$. Its weak global structure is already costing gain, even where it has not yet cost accuracy; Section~\ref{sec:exp6} shows the same trend at a wider range.

Where the accepted values fall matters as much as how many there are. Every untrained value accepted by Thermometer and by the linear hybrid at $B = 4$, the only encodings other than Binary to accept any, lies within four semantic units of the trained cluster, at a median distance of under three: these are neighbouring activations. Binary's lie at a median of $8.9$ and reach $142$ units away, with only a quarter falling near the cluster, and are false positives in the ordinary sense. Where the structured encodings accept untrained values at all, they generalise over the neighbourhood of what they were taught; Binary responds to unrelated values.

Gray's perfect accuracy here deserves comment, given its poor global structure. The trained set is a tight contiguous cluster, precisely the regime in which local smoothness suffices; Gray's weakness appears in Section~\ref{sec:exp2}, where the trained values are far apart and its false positives are scattered. Local smoothness is enough for local clustering, but not when the network must discriminate across the range.

\subsubsection{The excess is predictable}
\label{sec:coherence}

The factors of Table~\ref{tab:exp3} are not only measurable but predictable from the code words alone. Write each code word as a vector $t_i \in \{-1,+1\}^{W}$, taking $+1$ where the bit is one and $-1$ where it is zero. Training on a single value and applying the reciprocal wiring of Eq.~\eqref{eq:reciprocal} gives the output neuron a weight vector proportional to $t_i$, and summing over $n$ independently trained values gives $\sum_i t_i$. The drive produced by a trained value $j$ is therefore the inner product
\begin{equation}
\bigl\langle t_j, \textstyle\sum_i t_i \bigr\rangle
= W + \sum_{i \neq j} \bigl( W - 2H_{ij} \bigr),
\label{eq:innerproduct}
\end{equation}
where $H_{ij}$ is the Hamming distance between the two code words. Each pair contributes its full width less twice the distance between them: bits on which the two agree reinforce, and bits on which they disagree cancel.

The homeostatic factor must make every trained value fire, so it is set by the weakest of them: it is inversely proportional to the smallest such drive over the trained set, and its ratio to the coherent value $f(1)/n$ is $nW$ divided by that minimum. This requires no measurement, only the pairwise distances of the trained code words. Evaluated for the clusters of Table~\ref{tab:exp3} it gives $4.00$ for Binary at $n = 3$ against the measured $4.03$, $1.60$ for Gray at $n = 8$ against $1.61$, and $1.19$ for the cyclic hybrid at $B = 8$ against $1.21$; across all forty conditions the largest discrepancy is $0.03$. The prediction assumes that every bit learns a weight of the same magnitude, and the residual is consistent with the small difference between the two rails described in Section~\ref{sec:bitcount}.

\begin{table}[t]
\centering
\caption{Accepted sets against the predicted Hamming ball of radius
$h = \lfloor W(1-1/m)/2 \rfloor$, over all four targets and seven
multipliers plus the minimum. ``one shell more'' counts cells in which the
hardware accepted the next Hamming shell as well; ``partial'' counts cells
at which the prediction falls exactly on a shell boundary.}
\label{tab:exp5law}
\begin{tabular}{lrrrr}
\toprule
\textbf{Encoding} & $W$ & \textbf{exact} & \textbf{one shell more} & \textbf{partial} \\
\midrule
Binary & 8 & 31/32 & 0 & 1 \\
Gray & 8 & 31/32 & 0 & 1 \\
Thermometer & 255 & 20/32 & 12 & 0 \\
Hybrid lin. $B{=}4$ & 65 & 28/32 & 4 & 0 \\
Hybrid lin. $B{=}8$ & 34 & 32/32 & 0 & 0 \\
Johnson & 128 & 32/32 & 0 & 0 \\
Hybrid cyc. $B{=}4$ & 34 & 32/32 & 0 & 0 \\
Hybrid cyc. $B{=}8$ & 19 & 32/32 & 0 & 0 \\
\bottomrule
\end{tabular}
\end{table}

\begin{table*}[t]
\centering
\caption{Homeostatic tolerance, target 240. Each cell gives the number of values accepted and, in parentheses, the number of disjoint runs of consecutive values they form; $(1)$ is a single semantic interval. Counts are limited by the end of the range, which lies only $15$ values above the target. The multiplier scales each encoding's own minimum factor.}
\label{tab:exp5sweep}
\begin{tabular}{lrrrrrrr}
\toprule
\textbf{Encoding} & $\times1.1$ & $\times1.25$ & $\times1.5$ & $\times2.0$ & $\times3.0$ & $\times5.0$ & $\times8.0$ \\
\midrule
Binary & 1\,(1) & 1\,(1) & 9\,(7) & 37\,(21) & 37\,(21) & 93\,(35) & 93\,(35) \\
Gray & 1\,(1) & 1\,(1) & 9\,(7) & 37\,(22) & 37\,(22) & 93\,(36) & 93\,(36) \\
Thermometer & 25\,(1) & 42\,(1) & 58\,(1) & 80\,(1) & 101\,(1) & 118\,(1) & 127\,(1) \\
Hybrid lin. $B{=}4$ & 20\,(4) & 36\,(3) & 52\,(3) & 76\,(3) & 96\,(2) & 116\,(3) & 124\,(3) \\
Hybrid lin. $B{=}8$ & 6\,(4) & 27\,(5) & 44\,(4) & 68\,(5) & 92\,(4) & 108\,(4) & 116\,(5) \\
Johnson & 11\,(1) & 25\,(1) & 43\,(1) & 65\,(1) & 85\,(1) & 103\,(1) & 113\,(1) \\
Hybrid cyc. $B{=}4$ & 5\,(3) & 20\,(4) & 36\,(4) & 60\,(4) & 84\,(4) & 100\,(4) & 108\,(4) \\
Hybrid cyc. $B{=}8$ & 1\,(1) & 6\,(4) & 32\,(8) & 48\,(8) & 80\,(8) & 96\,(8) & 112\,(8) \\
\bottomrule
\end{tabular}
\end{table*}

\begin{table}[t]
\centering
\caption{Values accepted only by going round the end of the range, at
$\times2.0$ tolerance, by target. Only a cyclic encoding should accept
any; the interior target $128$ is a control at which none should. Binary and
Gray are marked n/a: their accepted sets are scattered, so a value that
happens to lie nearer by wrapping reflects a stray hit, not topology.}
\label{tab:exp5wrap}
\begin{tabular}{lrrrr}
\toprule
\textbf{Encoding} & \textbf{0} & \textbf{128} & \textbf{240} & \textbf{255} \\
\midrule
Binary & n/a & n/a & n/a & n/a \\
Gray & n/a & n/a & n/a & n/a \\
Thermometer & 0 & 0 & 0 & 0 \\
Hybrid lin. $B{=}4$ & 0 & 0 & 0 & 0 \\
Hybrid lin. $B{=}8$ & 0 & 0 & 0 & 0 \\
Johnson & 32 & 0 & 17 & 32 \\
Hybrid cyc. $B{=}4$ & 28 & 0 & 16 & 28 \\
Hybrid cyc. $B{=}8$ & 20 & 0 & 12 & 20 \\
\bottomrule
\end{tabular}
\end{table}

\subsection{Homeostatic Tolerance and Topology}
\label{sec:exp5}

Sections~\ref{sec:exp2} and~\ref{sec:exp3} operate at the minimum homeostatic factor. With a single trained value the network there is an exact-match detector: every other code word differs from the trained one in at least one bit and so receives less drive. To observe what the network generalises to, a single value was trained and the factor raised above that minimum by seven multipliers from $1.1$ to $8$, at four targets: the range ends $0$ and $255$, where the linear and cyclic topologies differ, and the interior values $128$ and $240$, where they should not. At each level all $256$ values were presented, giving $256$ conditions over the eight encodings, four targets and eight factor levels.

\subsubsection{Tolerance is a radius in code space}

A bit that disagrees with the trained code word swaps an excitatory contribution $+w$ for an inhibitory $-w$, by the reciprocal wiring of Eq.~\eqref{eq:reciprocal}, and so costs $2w$ of drive. At a gain $m$ times the minimum, the output neuron therefore still fires for any code word within
\begin{equation}
h(m) = \left\lfloor \frac{W}{2}\left(1 - \frac{1}{m}\right) \right\rfloor
\label{eq:tolerance}
\end{equation}
mismatched bits of the target, where $W$ is the code width. The network should accept exactly the code words at distance $h(m)$ or less.

The derivation uses only the code width and the reciprocal wiring of a single output neuron, so it applies to any neuron receiving a dual-rail input in this way: in a layer of neurons trained on different values, each would accept its own ball of radius $h(m)$ about its own trained code word. Two conditions bound it. Two conditions bound it. The uniform weight $w$ assumes a single trained value; with several, the weights differ from bit to bit and the drive becomes the inner product of Eq.~\eqref{eq:innerproduct}, of which Eq.~\eqref{eq:tolerance} is the special case. What the network computes is an inner product between its input and the accumulated weight vector, and Hamming distance governs its behaviour because, when that vector is a single code word with weights of uniform magnitude, the two determine each other. The derivation also assumes that a neuron's firing depends only on its own drive, so it would not survive lateral inhibition or any competitive mechanism between output neurons. Neither case is tested here.

Table~\ref{tab:exp5law} compares this prediction with the hardware. The accepted set has exactly the predicted size in $238$ of the $256$ measured conditions. Where the accepted values were few enough to be recorded individually, agreement in size always meant agreement in content: all $140$ such sets are identical to the prediction, value for value. Johnson and three of the four hybrids match in every condition.

The remaining $18$ conditions depart from the prediction by at most one bit of tolerance, and always towards accepting more rather than less. In $16$ of them the network also accepted the next \emph{Hamming shell}, that is every code word at exactly one further bit of distance. Twelve occur in Thermometer and four in the linear hybrid at $B = 4$, and the cause is the resolution of the measurement rather than of the network. The minimum factor is located to within $0.01$ and rounded upwards, so the gain actually applied can exceed the nominal multiple by up to $0.01/f(1)$, about $0.4$--$0.5\%$ here. By Eq.~\eqref{eq:tolerance} this raises the effective radius by up to $W/(2m)$ times that fraction: about half a bit for Thermometer at $\times 1.1$, a quarter of a bit at $\times 2$, and less for narrower codes. Wherever the predicted radius falls just short of an integer, that is enough to admit the next shell: the linear hybrid at $B = 4$ departs only at $\times 1.1$, where its predicted radius is $2.95$. Because the factor is rounded upwards, the error is always on the tolerant side.

The last two conditions occur where Eq.~\eqref{eq:tolerance} evaluates to an exact integer, placing the threshold on a shell boundary: Binary at target $128$ and Gray at target $255$, both at $\times 2$ tolerance, where $W(1-1/m)/2 = 2$. There the network accepts every code word within one bit of the target and seven of the twenty-eight at distance two. The seven are not arbitrary: they are precisely those code words containing the same number of set bits as the target, while the twenty-one rejected ones each contain two more. This is the asymmetry reported in Section~\ref{sec:bitcount}. Because the two rails are presented at different times relative to the teacher spike, their synapses acquire slightly different weights, so a code word with more set bits requires marginally more gain to reach threshold. The effect is of order one percent and has no consequence except where the threshold falls exactly on a shell boundary, as it does here.

Finally, the measurement is reproducible. The target $240$ trained alone in this experiment is the $n = 1$ case of Section~\ref{sec:exp3}, and the two independent hardware runs return identical minimum factors for all
eight encodings.

\subsubsection{The encoding determines what the radius means}
\label{sec:encodingradius}

Since the gain controls a radius in code space, the semantic meaning of tolerance is fixed entirely by the encoding (Table~\ref{tab:exp5sweep}).

Thermometer and Johnson accept a single contiguous interval at every tolerance and at every target, and it grows smoothly: from the target $240$, the furthest accepted value lies $12$ semantic units away at $\times 1.1$ and $111$ away at $\times 8$ for Thermometer, and $5$ then $56$ for Johnson. For these encodings, raising the homeostatic factor widens the detected neighbourhood around the trained value, and does nothing else.

Binary and Gray behave differently in kind. Their $8$-bit code space admits only a few distinct radii, so tolerance is quantised into shells of $1$, $9$, $37$ and $93$ code words: the network accepts nothing beyond the target up to $\times 1.25$, and then nine values at once. Under Binary those nine are scattered across seven disjoint runs and reach $128$ semantic units from the target, and at $\times 8$ the $93$ accepted values form $35$ runs; under Gray the nine reach $225$ units away. Raising the factor for these encodings admits more values, but not nearer ones.

The hybrids lie between the two. Their accepted sets grow smoothly with tolerance, as the unary codes do, but are broken into disjoint runs -- at most four at $B = 4$ and eight at $B = 8$. The count rises to that bound and stays there, unlike Binary's, which keeps growing, and it follows from the fine field: within a block the Gray code carries no global ordering, so the set of accepted code words crosses block boundaries unevenly.

Combining the tolerance law of Eq.~\eqref{eq:tolerance} with the block structure of Section~\ref{sec:encodings} gives a scaling relation. The semantic radius detected by a hybrid encoding is the product of its block size and its Hamming tolerance, $r = B\,h(m)$, which holds in all sixty conditions where the accepted set extends beyond the trained value. Substituting the code widths of Eq.~\eqref{eq:widths} and neglecting the $\log_2 B$ term gives $r \approx \frac{N}{2}(1 - 1/m)$ for the linear family and $r \approx \frac{N}{4}(1 - 1/m)$ for the cyclic one, in which the block size cancels. The detected neighbourhood therefore depends only on the value range, the gain and the topology, and not on the width of the code used to represent it: at $\times 8$ tolerance the furthest value accepted by Thermometer and by the linear hybrids at $B = 4$ and $B = 8$ lies $111$, $112$ and $112$ units from the target, while the three codes occupy $255$, $65$ and $34$ bits respectively. Narrowing the code therefore leaves the size of the detected neighbourhood unchanged, though, as noted above, it does cost its contiguity.

\subsubsection{Topology}

At the range ends the linear and cyclic families separate cleanly (Table~\ref{tab:exp5wrap}). Johnson and the cyclic hybrids accept values reached only by passing the end of the range: at target $255$ with $\times 2$ tolerance, Johnson accepts $223$ to $255$ and $0$ to $31$, a symmetric interval of $32$ values on either side of the target. Thermometer and the linear hybrids never do so, at any target or tolerance, and at the interior target $128$ none of the structured encodings does, as expected when the range ends lie beyond reach. The topology of each code, established analytically in Section~\ref{sec:topology}, is therefore reproduced in the behaviour of the trained network.

\subsection{Extending the Range to 1024 Values}
\label{sec:exp6}

\begin{table}[t]
\centering
\caption{A contiguous cluster around $960$ at $N = 1024$. Thermometer
(1023 bits) and Johnson (512 bits) exceed the default input capacity of Section~\ref{sec:packetlimit} at this range; the comparison is with the only established codes that fit within it. Top: false positives over all 1024 values. Bottom: the
homeostatic factor relative to coherent summation, $f(1)/n$.}
\label{tab:exp6}
\begin{tabular}{lrrrrr}
\toprule
\textbf{Encoding} & \textbf{Width} & $n{=}1$ & $n{=}2$ & $n{=}4$ & $n{=}8$ \\
\midrule
\multicolumn{6}{l}{\textbf{\emph{False positives}}} \\
Binary & 10 & 0 & 125 & 124 & 120 \\
Gray & 10 & 0 & 0 & 0 & 0 \\
Hybrid lin. $B{=}8$ & 130 & 0 & 0 & 12 & 0 \\
Hybrid lin. $B{=}16$ & 67 & 0 & 0 & 0 & 24 \\
Hybrid cyc. $B{=}8$ & 67 & 0 & 0 & 0 & 16 \\
Hybrid cyc. $B{=}16$ & 36 & 0 & 0 & 0 & 0 \\
\midrule
\multicolumn{6}{l}{\textbf{\emph{Factor relative to coherent summation}}} \\
Binary & 10 & 1.00 & 3.36 & 3.36 & 3.38 \\
Gray & 10 & 1.00 & 1.11 & 1.26 & 1.45 \\
Hybrid lin. $B{=}8$ & 130 & 1.00 & 1.02 & 1.04 & 1.04 \\
Hybrid lin. $B{=}16$ & 67 & 1.00 & 1.01 & 1.04 & 1.08 \\
Hybrid cyc. $B{=}8$ & 67 & 1.00 & 1.01 & 1.04 & 1.08 \\
Hybrid cyc. $B{=}16$ & 36 & 1.00 & 1.02 & 1.06 & 1.10 \\
\bottomrule
\end{tabular}
\end{table}

At $N = 1024$, Thermometer would need $1023$ bits and Johnson $512$. Under the default configuration of Section~\ref{sec:packetlimit} neither fits on SpiNNaker, and both would be silently truncated; Johnson would just fit within SpiNNaker2's limit of $512$ input events. The buffer could be enlarged to accommodate them, but each decision would then occupy $1023$ or $512$ buffer entries and deliver as many packets. The hybrids fit within the default: $130$ and $67$ bits for the linear variant at $B = 8$ and $16$, and $67$ and $36$ bits for the cyclic. The comparison at this range is therefore with Binary and Gray, the only established codes that also fit the default, repeating the protocol of  ection~\ref{sec:exp3} around the value $960$ ($1111000000_2$, the $10$-bit analogue of $240$). Table~\ref{tab:exp6} summarises the outcome and Table~\ref{tab:exp6full} gives the complete metrics. Every encoding trained correctly, and no trained value was missed.

Binary degrades more severely than at $8$ bits. With two trained values, $959$ and $960$, it accepts $125$ untrained values against $2$ trained ones, a precision of $0.016$, and its factor runs at $3.4$ times the coherent value from $n = 2$ onwards. The accepted values are the block $896$ to $1022$: every value whose three highest bits are $111$, except $1023$. This is the cancellation of Section~\ref{sec:testing} at its most visible. The two trained values lie either side of a seven-bit cliff, so the seven low-order synapses on which they disagree cancel, and the network is left detecting only the three bits they share. Its acceptances are contiguous, but their extent is set by the binary structure of the code rather than by the trained values: a block of $127$ values from two.

The hybrids remain coherent, their factors within $10\%$ of $f(1)/n$ at every trained-set size, and accept no untrained value in $13$ of their $16$ conditions. Where they do, $12$ to $24$ values at a time, the accepted values arrive in short runs aligned with the block structure: for the linear hybrid at $B = 16$ with eight trained values, runs of four at $928$--$931$, $944$--$947$, $952$--$955$, $964$--$967$, $972$--$975$ and $988$--$991$. This is the fine-field fragmentation of Section~\ref{sec:exp5} at a larger block size. All of them lie within $28$ units of the cluster in a range of $1024$, but only $38\%$ fall within four units, compared with all of Thermometer's at $N = 256$: the hybrids generalise to the neighbourhood of the cluster, but less tightly than the unary codes they replace.

Gray again accepts no untrained value, while its factor rises steadily, from $1.11$ to $1.45$ times the coherent value as the cluster grows. This reproduces the trend of Section~\ref{sec:exp3} at a wider range, and indicates that its weak global structure is consuming homeostatic margin before it produces errors.

The hybrids thus deliver at $1024$ values what the unary codes deliver at $256$: coherent scaling, and acceptances confined to the neighbourhood of the trained values. They do so within a default configuration that the unary codes exceed, and at roughly an eighth to a fifteenth of the width of the unary code with the same topology.

\section{Discussion}
\label{sec:discussion}

The results support four main conclusions about spatial representation in the SNN architecture studied here.

First, homeostatic scaling acts in code space rather than semantic space. For a single trained pattern, increasing the gain enlarges the set of code words whose Hamming distance from the target remains within the firing threshold. Equation~\eqref{eq:tolerance} predicts this Hamming radius from code width and gain alone, and the hardware measurements reproduce the predicted accepted set exactly in $238$ of $256$ conditions; of the remaining $18$, sixteen admit one further Hamming shell where the predicted radius falls just short of an integer and the resolution of the homeostatic search tips it over, and two fall exactly on a shell boundary, where the small asymmetry between rail weights determines which code words fire. What this tolerance means semantically is therefore decided entirely by how the encoding embeds the underlying variable.

Second, local continuity and global distance preservation are distinct properties. Gray encoding guarantees a one-bit transition between adjacent values and so removes the Hamming cliff, but its global Hamming distance is only weakly related to semantic separation: $\rho_s = 0.374$, barely above Binary's $0.323$. Thermometer and Johnson encode linear and circular distance exactly, and the hybrids retain most of this structure at much smaller widths. This explains why Gray performs well when the trained values form a compact cluster, yet needs progressively more homeostatic margin as the trained set grows, reaching $1.61$ times the coherent value, whereas Thermometer, Johnson and the hybrids stay within $21\%$.

Third, the interaction between several learned patterns can be predicted from their code geometry. In the bipolar representation of Eq.~\eqref{eq:innerproduct}, agreement between two code words reinforces the accumulated synaptic drive and disagreement cancels it, so the additional homeostatic factor required by a trained set follows from its pairwise Hamming distances, with a maximum discrepancy of $0.03$ from the measured values (Section~\ref{sec:coherence}). The same mechanism explains the large gain that Binary encoding requires, and why its acceptances follow the bits that values share rather than their proximity: scattered across the range at $256$ values, and at $1024$ values the aligned block of values sharing the three highest bits of the trained values.

Fourth, semantic structure has a hardware cost. A spatial code of width $W$ delivers $W$ input events to the receiving neuron for every decision, and the number of simultaneous input events a neuron can accept bounds the usable width: about $255$ under the default configuration of first-generation SpiNNaker, where wider codes are silently truncated, and $512$ on SpiNNaker2 (Section~\ref{sec:packetlimit}). Thermometer and Johnson offer exact distance preservation at an event count that grows linearly with the number of represented values, while the block size of the hybrids controls how much global structure and neighbourhood contiguity is kept for a given width. The choice is therefore not simply between ``accurate'' and ``compact'' codes, but between combinations of semantic topology, tolerance resolution and per-decision event load. In practice, this suggests a simple rule: choose a cyclic or linear code to match the topology of the variable; use Thermometer or Johnson when the range fits within the input limit; and otherwise use the smallest block size whose width, from Eq.~\eqref{eq:widths}, fits within it, since smaller blocks keep neighbourhoods more contiguous.

\subsection{Limitations}
\label{sec:limitations}

The evaluation uses synthetic values over a single range at a time, one network architecture, and first-generation SpiNNaker for every result except the input-capacity measurements. The experiments ran slower than real time, so the processing cost of wide codes, which would bound the rate of decisions at real time, was not measured. On SpiNNaker2 we established the limit of $512$ input events per neuron but not its origin: whether it can be configured, and whether it limits events per timestep or connections per neuron, remain open. The results are also drawn from a single output neuron; the tolerance law is derived per neuron and should extend to a layer of independently trained neurons, but this is not demonstrated here.

The weight vectors for multiple trained values are obtained by training each value separately and summing, not by presenting the values sequentially or interleaved to a single plastic network. This is a deliberate idealisation: it makes the learned vector an exact linear function of the trained set, which is what allows the coherence ratios of Section~\ref{sec:exp3} to be predicted from pairwise Hamming distances alone (Section~\ref{sec:coherence}). Sequential STDP would not be linear in the same way. The nearest-pair rule pairs each presentation only with its nearest post-synaptic spike, so the update a value induces depends on what has already been learned, and the weight ceiling of $2$ would clip bits on which several trained values agree: at a learned weight of about $0.375$ per value, synapses shared by six or more values would saturate, and the coherent scaling of $f(1)/n$ would cease to hold. Whether on-chip sequential learning converges to the summed vector for small trained sets, and how it departs from it for large ones, remains to be tested.

No downstream dataset or multi-feature classification problem is evaluated, and the introduction explains why: the exhaustive protocol that makes the relations above exact is not available on a natural dataset. The consequence is that nothing here shows how the encodings behave when several features are presented together. One difficulty is already visible from the tolerance law: if features are concatenated into one code word, the Hamming distance is the sum of the per-feature distances and the tolerance radius applies to that sum, so a large deviation in one feature can be offset by exact agreement in the others. A multi-feature classifier built on these encodings would need either per-feature detectors with their own radii or a weighting across features, and the choice between them is an open design question.

The hybrids do not fully reproduce their parents' behaviour. Their accepted sets are fragmented into up to $B$ runs by the Gray fine field, and their false positives, though confined near the trained cluster, are less tightly concentrated than Thermometer's. The block size therefore trades not only correlation but neighbourhood contiguity, and smaller blocks should be preferred where contiguity matters.

The tolerance law is exact only to the resolution of the homeostatic search: rounding the factor upwards can admit one further Hamming shell wherever the predicted radius falls just short of an integer, which occurred mostly in the widest code, and a finer search would be needed to exploit Thermometer's full precision. Finally, the hybrid construction resembles the segmented thermometer-plus-binary architectures of data-converter design, and the contribution here is the Gray fine field with reflected block ordering and its characterisation in a neuromorphic setting, rather than the two-level idea itself.

\section{Conclusions}
\label{sec:conclusions}

Spatial encoding removes the latency of rate coding but makes classification depend on the geometry of the code. For a dual-rail network with a single output neuron on first-generation SpiNNaker, homeostatic scaling sets a Hamming radius around each trained code word, and the encoding alone decides what that radius means: a contiguous interval of values for Thermometer and Johnson, wrapping at the ends of the range for the cyclic codes, and a scattered set of values for Binary and Gray.

The encodings commonly proposed against the Hamming cliff are therefore not equivalent. Gray removes the cliff but discards global distance structure, and pays for it in homeostatic margin as trained sets grow; Thermometer and Johnson preserve that structure at a width that grows with the value range. The hybrid construction introduced here contains all three as limiting cases of a single block-size parameter and trades width against preservation of semantic distance at a favourable rate: the cyclic hybrid at $B = 8$ accepts no untrained value as the trained cluster grows, in $19$ bits where Johnson needs $128$.

Width matters because it is the number of input events paid on every decision, and both platforms bound it: on SpiNNaker by a configurable buffer of the simulation software, which by default silently truncates codes wider than $255$ bits during training, and on SpiNNaker2 by a limit of $512$ input events per neuron. The hybrid construction reduces width by roughly an order of magnitude while leaving the radius of the detected neighbourhood unchanged.

Future work will apply these encodings to network-traffic classification, where continuous, ordinal and cyclic features occur together, and will compute the mapping on the hardware itself rather than as a pre-processing step, so that feature values can be encoded as they arrive.

\section*{Acknowledgement}

For the purpose of open access, the authors have applied a Creative Commons Attribution (CC BY) licence to any Author Accepted Manuscript version arising from this submission.

This work has been supported by the Department of Computing and Mathematics at the Manchester Metropolitan University and authorised through ethical review 93014.

Prof. Di Nuovo acknowledges the support of the UK Engineering and Physical Sciences Research Council (grant number EP/X018733/1 for the project ALDENS), and Innovate UK (grant number 10089807 for the Horizon Europe project PRIMI Grant agreement n. 101120727).

Prof. Han acknowledges the support of the UK Engineering and Physical Sciences Research Council (grants number EP/X013707/1 and UKRI3606).

The authors extend their gratitude to Prof. Steve Furber and all members, current and past, of the APT group at the University of Manchester for their invaluable support.

\section*{Data and code availability}

The data and the code \citep{Davies2026EncodedDataset} used to generate this article are available on GitHub at the URL: \url{https://github.com/sergiodavies/EncodedSpatialLearningCodeAndDataset}.

\section*{Conflicts of interest}
The authors declare that they have no known competing financial interests or personal relationships that could have appeared to influence the work reported in this paper.

\section*{Declaration of Generative AI and AI-assisted technologies in the writing process}
During the preparation of this work the authors used Anthropic Claude in order to improve readability and language and as support for the generation of the code for the experiments. After using this tool, the authors reviewed and edited the content as needed and take full responsibility for the content of the publication.

\printcredits

\bibliographystyle{cas-model2-names}

\bibliography{references}



\bio{a1}
Sergio Davies is a senior lecturer in AI in the Department of Computing and Mathematics at Manchester Metropolitan University. He received his Laurea (equivalent to MSc Eng) in Telecommunication Engineering from the University ``Federico II'' in Napoli, Italy in 2006. Following his Ph.D. in Computer Science at the University of Manchester in 2012, Sergio continued his research on spiking neural networks as a postdoc in the Human Brain Project until 2016. Transitioning to industry, he took leadership roles in strategic advising as a consultant, and then managed various research projects covering computer hardware design, embedded systems architecture, software development and system networking. In 2019, Sergio returned to academia, initially at Sheffield Hallam University before joining Manchester Metropolitan University. His current research focuses on practical application of spiking neural networks in real-world scenarios.
\endbio


\bio{a2}
Muhammad Aitsam is a Research Fellow in Artificial Intelligence and Robotics at Sheffield Hallam University, UK. He received his PhD in Computing from the same institution, where his doctoral research focused on neuromorphic computing and event-based vision for intelligent robotic systems. His research interests include neuromorphic computing, event-based vision, spiking neural networks, human-robot interaction, and interactive and autonomous robotics. Dr Aitsam has contributed to several UK and European research projects, including the EU-funded PERSEO project, the EPSRC-funded ALDENS project, and the Innovate UK-funded PRIMI project. He also contributes to the research community through peer review and programme committee activities for international conferences and journals in artificial intelligence, computer vision, neuromorphic computing, and robotics.
\endbio

\bio{a3}
Morgan-Tyler Stone is a PhD candidate in the Department of Computing and Mathematics at Manchester Metropolitan University, UK. Their research investigates neuromorphic intrusion detection, using spiking neural networks on the SpiNNaker platform to build intrusion detection systems that operate in real time at low energy cost. They received a BSc in Computer Science for Games from Sheffield Hallam University. Their wider interests include neuromorphic and edge computing, cybersecurity, and energy-efficient real-time intelligent systems.
\endbio

\bio{a4}
Professor Alessandro Di Nuovo is Director of the Centre of Excellence in Artificial Intelligence and Robotics and Professor of Machine Intelligence at Sheffield Hallam University, where he has been based since 2015. He is also a Full Professor at the University of Catania, Italy, an affiliation he has held since 2024.

He obtained both his Laurea (MSc Eng) and PhD in Informatics Engineering from the University of Catania in 2005 and 2009, respectively. His research focuses on computational intelligence and its applications to cognitive modelling, human-robot interaction, computer-aided assessment of intellectual disabilities, and embedded intelligent systems.

Professor Di Nuovo has authored more than 150 scientific publications and has an extensive track record of leading interdisciplinary research and innovation in both fundamental and applied AI and robotics. His work has attracted funding from major national and international agencies, including the Engineering and Physical Sciences Research Council (EPSRC) and the European Union, as well as a range of industrial partners. His research aims to advance trustworthy, efficient, and human-centred intelligent systems with real-world impact across society and industry.
\endbio

\bio{a5}
Prof. Liangxiu Han is Professor of Computer Science at Manchester Metropolitan University, UK. Her research focuses on scalable, data-driven artificial intelligence and machine learning, with applications across healthcare, precision agriculture, smart cities, cybersecurity and energy. She has extensive experience in developing AI methods for large-scale multimodal data, including imaging, sensor, geospatial, numerical and textual data. She has led and contributed to national and international research programmes funded by UKRI, NIHR, the European Union, GCRF/Newton, industry and charities, and is currently Principal Investigator or Co-Investigator on a number of research projects across these areas.
\endbio

\onecolumn
\appendix
\setcounter{table}{0}
\renewcommand{\thetable}{A.\arabic{table}}
\captionsetup[longtable]{font=footnotesize, width=.78\textwidth}
\setlength{\LTpre}{\medskipamount}

\section{Complete Classification Metrics}
\label{app:metrics}
{\normalsize All values in the range were presented to each trained network; the five metrics are those of \citet{Davies2025SupervisedSpiNNaker}.\par}

\subsection{Code words of the second values}

\begin{footnotesize}
\begin{longtable}{r ll lll lll}
\caption{Code words of the anchor $240$ and of the fourteen second values of Table~\ref{tab:sweep} under each encoding. Long codes are written in run-length form, $1^{a}0^{b}$ denoting $a$ ones followed by $b$ zeros; the Hamming distances of Table~\ref{tab:sweep} can be read off by comparing each row with the anchor.}\label{tab:sweepcodes}\\
\toprule
& & & \multicolumn{3}{c}{Linear family} & \multicolumn{3}{c}{Cyclic family} \\
\cmidrule(lr){4-6}\cmidrule(l){7-9}
Value & Binary & Gray & Thermometer & $B{=}4$ & $B{=}8$ & Johnson & $B{=}4$ & $B{=}8$ \\
\midrule
\endfirsthead
\multicolumn{9}{l}{\tablename\ \thetable{} -- continued} \\
\toprule
Value & Binary & Gray & Thermometer & $B{=}4$ & $B{=}8$ & Johnson & $B{=}4$ & $B{=}8$ \\
\midrule
\endhead
\midrule
\multicolumn{9}{r}{\emph{continued}} \\
\endfoot
\bottomrule
\endlastfoot
240 & 11110000 & 10001000 & $1^{240}0^{15}$ & $1^{60}0^{5}$ & $1^{30}0^{4}$ & $0^{112}1^{16}$ & $0^{28}1^{4}0^{2}$ & $0^{14}1^{2}0^{3}$ \\
\midrule
241 & 11110001 & 10001001 & $1^{241}0^{14}$ & $1^{60}0^{4}1$ & $1^{30}0^{3}1$ & $0^{113}1^{15}$ & $0^{28}1^{4}01$ & $0^{14}1^{2}0^{2}1$ \\
242 & 11110010 & 10001011 & $1^{242}0^{13}$ & $1^{60}0^{3}1^{2}$ & $1^{30}0^{2}1^{2}$ & $0^{114}1^{14}$ & $0^{28}1^{6}$ & $0^{14}1^{2}01^{2}$ \\
244 & 11110100 & 10001110 & $1^{244}0^{11}$ & $1^{61}0^{2}10$ & $1^{30}01^{2}0$ & $0^{116}1^{12}$ & $0^{29}1^{4}0$ & $0^{14}1^{4}0$ \\
248 & 11111000 & 10000100 & $1^{248}0^{7}$ & $1^{62}0^{3}$ & $1^{32}0^{2}$ & $0^{120}1^{8}$ & $0^{30}1^{2}0^{2}$ & $0^{15}1^{2}0^{2}$ \\
\addlinespace
239 & 11101111 & 10011000 & $1^{239}0^{16}$ & $1^{59}0^{6}$ & $1^{29}0^{5}$ & $0^{111}1^{17}$ & $0^{27}1^{5}0^{2}$ & $0^{13}1^{3}0^{3}$ \\
238 & 11101110 & 10011001 & $1^{238}0^{17}$ & $1^{59}0^{5}1$ & $1^{29}0^{4}1$ & $0^{110}1^{18}$ & $0^{27}1^{5}01$ & $0^{13}1^{3}0^{2}1$ \\
236 & 11101100 & 10011010 & $1^{236}0^{19}$ & $1^{59}0^{4}10$ & $1^{29}0^{3}10$ & $0^{108}1^{20}$ & $0^{27}1^{6}0$ & $0^{13}1^{3}010$ \\
232 & 11101000 & 10011100 & $1^{232}0^{23}$ & $1^{58}0^{7}$ & $1^{29}0^{2}10^{2}$ & $0^{104}1^{24}$ & $0^{26}1^{6}0^{2}$ & $0^{13}1^{4}0^{2}$ \\
\addlinespace
224 & 11100000 & 10010000 & $1^{224}0^{31}$ & $1^{56}0^{9}$ & $1^{28}0^{6}$ & $0^{96}1^{32}$ & $0^{24}1^{8}0^{2}$ & $0^{12}1^{4}0^{3}$ \\
192 & 11000000 & 10100000 & $1^{192}0^{63}$ & $1^{48}0^{17}$ & $1^{24}0^{10}$ & $0^{64}1^{64}$ & $0^{16}1^{16}0^{2}$ & $0^{8}1^{8}0^{3}$ \\
128 & 10000000 & 11000000 & $1^{128}0^{127}$ & $1^{32}0^{33}$ & $1^{16}0^{18}$ & $1^{128}$ & $1^{32}0^{2}$ & $1^{16}0^{3}$ \\
64 & 01000000 & 01100000 & $1^{64}0^{191}$ & $1^{16}0^{49}$ & $1^{8}0^{26}$ & $1^{64}0^{64}$ & $1^{16}0^{18}$ & $1^{8}0^{11}$ \\
0 & 00000000 & 00000000 & $0^{255}$ & $0^{65}$ & $0^{34}$ & $0^{128}$ & $0^{34}$ & $0^{19}$ \\
\addlinespace
255 & 11111111 & 10000000 & $1^{255}$ & $1^{63}0^{2}$ & $1^{31}0^{3}$ & $0^{127}1$ & $0^{31}10^{2}$ & $0^{15}10^{3}$ \\
\end{longtable}
\end{footnotesize}

\subsection{Two trained values}

\begin{footnotesize}
\begin{longtable}{rrr rrrr rrrrr}
\caption{Two trained values: complete classification metrics. Anchor $240$ trained together with $v_2$; $H$ is the Hamming distance between the two code words and $S$ the homeostatic scaling factor. Counts are over all $256$ values.}\label{tab:exp2full}\\
\toprule
$v_2$ & $H$ & $S$ & TP & TN & FP & FN & Acc. & Prec. & Neg.\,pred. & Sens. & Spec. \\
\midrule
\endfirsthead
\multicolumn{12}{l}%
{\tablename\ \thetable{} -- continued} \\
\toprule
$v_2$ & $H$ & $S$ & TP & TN & FP & FN & Acc. & Prec. & Neg.\,pred. & Sens. & Spec. \\
\midrule
\endhead
\midrule
\multicolumn{12}{r}{\emph{continued}} \\
\endfoot
\bottomrule
\endlastfoot
\multicolumn{12}{l}{\textbf{Binary} (8 bits)} \\
241 & 1 & 2.97 & 2 & 254 & 0 & 0 & 1.0000 & 1.0000 & 1.0000 & 1.0000 & 1.0000 \\
242 & 1 & 2.97 & 2 & 254 & 0 & 0 & 1.0000 & 1.0000 & 1.0000 & 1.0000 & 1.0000 \\
244 & 1 & 2.97 & 2 & 254 & 0 & 0 & 1.0000 & 1.0000 & 1.0000 & 1.0000 & 1.0000 \\
248 & 1 & 2.97 & 2 & 254 & 0 & 0 & 1.0000 & 1.0000 & 1.0000 & 1.0000 & 1.0000 \\
224 & 1 & 2.96 & 2 & 254 & 0 & 0 & 1.0000 & 1.0000 & 1.0000 & 1.0000 & 1.0000 \\
232 & 2 & 3.45 & 2 & 252 & 2 & 0 & 0.9922 & 0.5000 & 1.0000 & 1.0000 & 0.9921 \\
192 & 2 & 3.45 & 2 & 252 & 2 & 0 & 0.9922 & 0.5000 & 1.0000 & 1.0000 & 0.9921 \\
236 & 3 & 4.14 & 2 & 249 & 5 & 0 & 0.9805 & 0.2857 & 1.0000 & 1.0000 & 0.9803 \\
128 & 3 & 4.14 & 2 & 248 & 6 & 0 & 0.9766 & 0.2500 & 1.0000 & 1.0000 & 0.9764 \\
64 & 3 & 4.14 & 2 & 248 & 6 & 0 & 0.9766 & 0.2500 & 1.0000 & 1.0000 & 0.9764 \\
238 & 4 & 5.19 & 2 & 240 & 14 & 0 & 0.9453 & 0.1250 & 1.0000 & 1.0000 & 0.9449 \\
0 & 4 & 5.18 & 2 & 240 & 14 & 0 & 0.9453 & 0.1250 & 1.0000 & 1.0000 & 0.9449 \\
255 & 4 & 5.20 & 2 & 240 & 14 & 0 & 0.9453 & 0.1250 & 1.0000 & 1.0000 & 0.9449 \\
239 & 5 & 6.94 & 2 & 224 & 30 & 0 & 0.8828 & 0.0625 & 1.0000 & 1.0000 & 0.8819 \\
\addlinespace
\multicolumn{12}{l}{\textbf{Gray} (8 bits)} \\
241 & 1 & 2.96 & 2 & 254 & 0 & 0 & 1.0000 & 1.0000 & 1.0000 & 1.0000 & 1.0000 \\
239 & 1 & 2.96 & 2 & 254 & 0 & 0 & 1.0000 & 1.0000 & 1.0000 & 1.0000 & 1.0000 \\
255 & 1 & 2.96 & 2 & 254 & 0 & 0 & 1.0000 & 1.0000 & 1.0000 & 1.0000 & 1.0000 \\
242 & 2 & 3.45 & 2 & 252 & 2 & 0 & 0.9922 & 0.5000 & 1.0000 & 1.0000 & 0.9921 \\
244 & 2 & 3.45 & 2 & 252 & 2 & 0 & 0.9922 & 0.5000 & 1.0000 & 1.0000 & 0.9921 \\
248 & 2 & 3.45 & 2 & 252 & 2 & 0 & 0.9922 & 0.5000 & 1.0000 & 1.0000 & 0.9921 \\
238 & 2 & 3.45 & 2 & 252 & 2 & 0 & 0.9922 & 0.5000 & 1.0000 & 1.0000 & 0.9921 \\
236 & 2 & 3.45 & 2 & 252 & 2 & 0 & 0.9922 & 0.5000 & 1.0000 & 1.0000 & 0.9921 \\
232 & 2 & 3.45 & 2 & 252 & 2 & 0 & 0.9922 & 0.5000 & 1.0000 & 1.0000 & 0.9921 \\
224 & 2 & 3.45 & 2 & 252 & 2 & 0 & 0.9922 & 0.5000 & 1.0000 & 1.0000 & 0.9921 \\
192 & 2 & 3.45 & 2 & 252 & 2 & 0 & 0.9922 & 0.5000 & 1.0000 & 1.0000 & 0.9921 \\
128 & 2 & 3.45 & 2 & 252 & 2 & 0 & 0.9922 & 0.5000 & 1.0000 & 1.0000 & 0.9921 \\
0 & 2 & 3.45 & 2 & 252 & 2 & 0 & 0.9922 & 0.5000 & 1.0000 & 1.0000 & 0.9921 \\
64 & 4 & 5.16 & 2 & 245 & 9 & 0 & 0.9648 & 0.1818 & 1.0000 & 1.0000 & 0.9646 \\
\addlinespace
\multicolumn{12}{l}{\textbf{Thermometer} (255 bits)} \\
241 & 1 & 1.24 & 2 & 254 & 0 & 0 & 1.0000 & 1.0000 & 1.0000 & 1.0000 & 1.0000 \\
239 & 1 & 1.24 & 2 & 254 & 0 & 0 & 1.0000 & 1.0000 & 1.0000 & 1.0000 & 1.0000 \\
242 & 2 & 1.25 & 2 & 251 & 3 & 0 & 0.9883 & 0.4000 & 1.0000 & 1.0000 & 0.9882 \\
238 & 2 & 1.25 & 2 & 251 & 3 & 0 & 0.9883 & 0.4000 & 1.0000 & 1.0000 & 0.9882 \\
244 & 4 & 1.25 & 2 & 251 & 3 & 0 & 0.9883 & 0.4000 & 1.0000 & 1.0000 & 0.9882 \\
236 & 4 & 1.25 & 2 & 251 & 3 & 0 & 0.9883 & 0.4000 & 1.0000 & 1.0000 & 0.9882 \\
248 & 8 & 1.28 & 2 & 247 & 7 & 0 & 0.9727 & 0.2222 & 1.0000 & 1.0000 & 0.9724 \\
232 & 8 & 1.28 & 2 & 247 & 7 & 0 & 0.9727 & 0.2222 & 1.0000 & 1.0000 & 0.9724 \\
255 & 15 & 1.32 & 2 & 239 & 15 & 0 & 0.9414 & 0.1176 & 1.0000 & 1.0000 & 0.9409 \\
224 & 16 & 1.33 & 2 & 237 & 17 & 0 & 0.9336 & 0.1053 & 1.0000 & 1.0000 & 0.9331 \\
192 & 48 & 1.52 & 2 & 207 & 47 & 0 & 0.8164 & 0.0408 & 1.0000 & 1.0000 & 0.8150 \\
128 & 112 & 2.19 & 2 & 143 & 111 & 0 & 0.5664 & 0.0177 & 1.0000 & 1.0000 & 0.5630 \\
64 & 176 & 3.96 & 2 & 79 & 175 & 0 & 0.3164 & 0.0113 & 1.0000 & 1.0000 & 0.3110 \\
0 & 240 & 21.01 & 2 & 15 & 239 & 0 & 0.0664 & 0.0083 & 1.0000 & 1.0000 & 0.0591 \\
\addlinespace
\multicolumn{12}{l}{\textbf{Hybrid lin. $B{=}4$} (65 bits)} \\
241 & 1 & 1.09 & 2 & 254 & 0 & 0 & 1.0000 & 1.0000 & 1.0000 & 1.0000 & 1.0000 \\
239 & 1 & 1.09 & 2 & 254 & 0 & 0 & 1.0000 & 1.0000 & 1.0000 & 1.0000 & 1.0000 \\
242 & 2 & 1.11 & 2 & 252 & 2 & 0 & 0.9922 & 0.5000 & 1.0000 & 1.0000 & 0.9921 \\
244 & 2 & 1.11 & 2 & 252 & 2 & 0 & 0.9922 & 0.5000 & 1.0000 & 1.0000 & 0.9921 \\
248 & 2 & 1.11 & 2 & 253 & 1 & 0 & 0.9961 & 0.6667 & 1.0000 & 1.0000 & 0.9961 \\
238 & 2 & 1.11 & 2 & 252 & 2 & 0 & 0.9922 & 0.5000 & 1.0000 & 1.0000 & 0.9921 \\
236 & 2 & 1.11 & 2 & 252 & 2 & 0 & 0.9922 & 0.5000 & 1.0000 & 1.0000 & 0.9921 \\
232 & 2 & 1.11 & 2 & 253 & 1 & 0 & 0.9961 & 0.6667 & 1.0000 & 1.0000 & 0.9961 \\
255 & 3 & 1.13 & 2 & 252 & 2 & 0 & 0.9922 & 0.5000 & 1.0000 & 1.0000 & 0.9921 \\
224 & 4 & 1.15 & 2 & 251 & 3 & 0 & 0.9883 & 0.4000 & 1.0000 & 1.0000 & 0.9882 \\
192 & 12 & 1.32 & 2 & 243 & 11 & 0 & 0.9570 & 0.1538 & 1.0000 & 1.0000 & 0.9567 \\
128 & 28 & 1.89 & 2 & 227 & 27 & 0 & 0.8945 & 0.0690 & 1.0000 & 1.0000 & 0.8937 \\
64 & 44 & 3.33 & 2 & 211 & 43 & 0 & 0.8320 & 0.0444 & 1.0000 & 1.0000 & 0.8307 \\
0 & 60 & 13.90 & 2 & 195 & 59 & 0 & 0.7695 & 0.0328 & 1.0000 & 1.0000 & 0.7677 \\
\addlinespace
\multicolumn{12}{l}{\textbf{Hybrid lin. $B{=}8$} (34 bits)} \\
241 & 1 & 1.09 & 2 & 254 & 0 & 0 & 1.0000 & 1.0000 & 1.0000 & 1.0000 & 1.0000 \\
239 & 1 & 1.09 & 2 & 254 & 0 & 0 & 1.0000 & 1.0000 & 1.0000 & 1.0000 & 1.0000 \\
255 & 1 & 1.09 & 2 & 254 & 0 & 0 & 1.0000 & 1.0000 & 1.0000 & 1.0000 & 1.0000 \\
242 & 2 & 1.12 & 2 & 252 & 2 & 0 & 0.9922 & 0.5000 & 1.0000 & 1.0000 & 0.9921 \\
244 & 2 & 1.12 & 2 & 252 & 2 & 0 & 0.9922 & 0.5000 & 1.0000 & 1.0000 & 0.9921 \\
248 & 2 & 1.12 & 2 & 252 & 2 & 0 & 0.9922 & 0.5000 & 1.0000 & 1.0000 & 0.9921 \\
238 & 2 & 1.12 & 2 & 252 & 2 & 0 & 0.9922 & 0.5000 & 1.0000 & 1.0000 & 0.9921 \\
236 & 2 & 1.12 & 2 & 252 & 2 & 0 & 0.9922 & 0.5000 & 1.0000 & 1.0000 & 0.9921 \\
232 & 2 & 1.12 & 2 & 252 & 2 & 0 & 0.9922 & 0.5000 & 1.0000 & 1.0000 & 0.9921 \\
224 & 2 & 1.12 & 2 & 253 & 1 & 0 & 0.9961 & 0.6667 & 1.0000 & 1.0000 & 0.9961 \\
192 & 6 & 1.28 & 2 & 249 & 5 & 0 & 0.9805 & 0.2857 & 1.0000 & 1.0000 & 0.9803 \\
128 & 14 & 1.79 & 2 & 241 & 13 & 0 & 0.9492 & 0.1333 & 1.0000 & 1.0000 & 0.9488 \\
64 & 22 & 2.97 & 2 & 233 & 21 & 0 & 0.9180 & 0.0870 & 1.0000 & 1.0000 & 0.9173 \\
0 & 30 & 8.91 & 2 & 225 & 29 & 0 & 0.8867 & 0.0645 & 1.0000 & 1.0000 & 0.8858 \\
\addlinespace
\multicolumn{12}{l}{\textbf{Johnson} (128 bits)} \\
241 & 1 & 1.14 & 2 & 254 & 0 & 0 & 1.0000 & 1.0000 & 1.0000 & 1.0000 & 1.0000 \\
239 & 1 & 1.14 & 2 & 254 & 0 & 0 & 1.0000 & 1.0000 & 1.0000 & 1.0000 & 1.0000 \\
242 & 2 & 1.15 & 2 & 253 & 1 & 0 & 0.9961 & 0.6667 & 1.0000 & 1.0000 & 0.9961 \\
238 & 2 & 1.15 & 2 & 253 & 1 & 0 & 0.9961 & 0.6667 & 1.0000 & 1.0000 & 0.9961 \\
244 & 4 & 1.16 & 2 & 251 & 3 & 0 & 0.9883 & 0.4000 & 1.0000 & 1.0000 & 0.9882 \\
236 & 4 & 1.16 & 2 & 251 & 3 & 0 & 0.9883 & 0.4000 & 1.0000 & 1.0000 & 0.9882 \\
248 & 8 & 1.20 & 2 & 247 & 7 & 0 & 0.9727 & 0.2222 & 1.0000 & 1.0000 & 0.9724 \\
232 & 8 & 1.20 & 2 & 247 & 7 & 0 & 0.9727 & 0.2222 & 1.0000 & 1.0000 & 0.9724 \\
255 & 15 & 1.27 & 2 & 240 & 14 & 0 & 0.9453 & 0.1250 & 1.0000 & 1.0000 & 0.9449 \\
224 & 16 & 1.29 & 2 & 239 & 15 & 0 & 0.9414 & 0.1176 & 1.0000 & 1.0000 & 0.9409 \\
0 & 16 & 1.29 & 2 & 239 & 15 & 0 & 0.9414 & 0.1176 & 1.0000 & 1.0000 & 0.9409 \\
192 & 48 & 1.80 & 2 & 207 & 47 & 0 & 0.8164 & 0.0408 & 1.0000 & 1.0000 & 0.8150 \\
64 & 80 & 3.00 & 2 & 175 & 79 & 0 & 0.6914 & 0.0247 & 1.0000 & 1.0000 & 0.6890 \\
128 & 112 & 8.97 & 2 & 143 & 111 & 0 & 0.5664 & 0.0177 & 1.0000 & 1.0000 & 0.5630 \\
\addlinespace
\multicolumn{12}{l}{\textbf{Hybrid cyc. $B{=}4$} (34 bits)} \\
241 & 1 & 1.09 & 2 & 254 & 0 & 0 & 1.0000 & 1.0000 & 1.0000 & 1.0000 & 1.0000 \\
239 & 1 & 1.09 & 2 & 254 & 0 & 0 & 1.0000 & 1.0000 & 1.0000 & 1.0000 & 1.0000 \\
242 & 2 & 1.12 & 2 & 252 & 2 & 0 & 0.9922 & 0.5000 & 1.0000 & 1.0000 & 0.9921 \\
244 & 2 & 1.12 & 2 & 252 & 2 & 0 & 0.9922 & 0.5000 & 1.0000 & 1.0000 & 0.9921 \\
248 & 2 & 1.12 & 2 & 253 & 1 & 0 & 0.9961 & 0.6667 & 1.0000 & 1.0000 & 0.9961 \\
238 & 2 & 1.12 & 2 & 252 & 2 & 0 & 0.9922 & 0.5000 & 1.0000 & 1.0000 & 0.9921 \\
236 & 2 & 1.12 & 2 & 252 & 2 & 0 & 0.9922 & 0.5000 & 1.0000 & 1.0000 & 0.9921 \\
232 & 2 & 1.12 & 2 & 253 & 1 & 0 & 0.9961 & 0.6667 & 1.0000 & 1.0000 & 0.9961 \\
255 & 3 & 1.15 & 2 & 252 & 2 & 0 & 0.9922 & 0.5000 & 1.0000 & 1.0000 & 0.9921 \\
224 & 4 & 1.20 & 2 & 251 & 3 & 0 & 0.9883 & 0.4000 & 1.0000 & 1.0000 & 0.9882 \\
0 & 4 & 1.19 & 2 & 251 & 3 & 0 & 0.9883 & 0.4000 & 1.0000 & 1.0000 & 0.9882 \\
192 & 12 & 1.63 & 2 & 243 & 11 & 0 & 0.9570 & 0.1538 & 1.0000 & 1.0000 & 0.9567 \\
64 & 20 & 2.55 & 2 & 235 & 19 & 0 & 0.9258 & 0.0952 & 1.0000 & 1.0000 & 0.9252 \\
128 & 28 & 5.93 & 2 & 227 & 27 & 0 & 0.8945 & 0.0690 & 1.0000 & 1.0000 & 0.8937 \\
\addlinespace
\multicolumn{12}{l}{\textbf{Hybrid cyc. $B{=}8$} (19 bits)} \\
241 & 1 & 1.16 & 2 & 254 & 0 & 0 & 1.0000 & 1.0000 & 1.0000 & 1.0000 & 1.0000 \\
239 & 1 & 1.16 & 2 & 254 & 0 & 0 & 1.0000 & 1.0000 & 1.0000 & 1.0000 & 1.0000 \\
255 & 1 & 1.16 & 2 & 254 & 0 & 0 & 1.0000 & 1.0000 & 1.0000 & 1.0000 & 1.0000 \\
242 & 2 & 1.23 & 2 & 252 & 2 & 0 & 0.9922 & 0.5000 & 1.0000 & 1.0000 & 0.9921 \\
244 & 2 & 1.23 & 2 & 252 & 2 & 0 & 0.9922 & 0.5000 & 1.0000 & 1.0000 & 0.9921 \\
248 & 2 & 1.23 & 2 & 252 & 2 & 0 & 0.9922 & 0.5000 & 1.0000 & 1.0000 & 0.9921 \\
238 & 2 & 1.23 & 2 & 252 & 2 & 0 & 0.9922 & 0.5000 & 1.0000 & 1.0000 & 0.9921 \\
236 & 2 & 1.23 & 2 & 252 & 2 & 0 & 0.9922 & 0.5000 & 1.0000 & 1.0000 & 0.9921 \\
232 & 2 & 1.23 & 2 & 252 & 2 & 0 & 0.9922 & 0.5000 & 1.0000 & 1.0000 & 0.9921 \\
224 & 2 & 1.23 & 2 & 253 & 1 & 0 & 0.9961 & 0.6667 & 1.0000 & 1.0000 & 0.9961 \\
0 & 2 & 1.23 & 2 & 253 & 1 & 0 & 0.9961 & 0.6667 & 1.0000 & 1.0000 & 0.9961 \\
192 & 6 & 1.61 & 2 & 249 & 5 & 0 & 0.9805 & 0.2857 & 1.0000 & 1.0000 & 0.9803 \\
64 & 10 & 2.31 & 2 & 245 & 9 & 0 & 0.9648 & 0.1818 & 1.0000 & 1.0000 & 0.9646 \\
128 & 14 & 4.15 & 2 & 241 & 13 & 0 & 0.9492 & 0.1333 & 1.0000 & 1.0000 & 0.9488 \\
\addlinespace
\end{longtable}
\end{footnotesize}

\subsection{Growing trained sets}

\begin{footnotesize}
\begin{longtable}{rr rrrr rrrrr r}
\caption{Growing trained sets: complete classification metrics. A contiguous cluster of $n$ values around $240$ is trained; $S$ is the homeostatic scaling factor. ``med.\,$d$'' is the median semantic distance of the false positives from the trained cluster.}\label{tab:exp3full}\\
\toprule
$n$ & $S$ & TP & TN & FP & FN & Acc. & Prec. & Neg.\,pred. & Sens. & Spec. & med.\,$d$ \\
\midrule
\endfirsthead
\multicolumn{12}{l}%
{\tablename\ \thetable{} -- continued} \\
\toprule
$n$ & $S$ & TP & TN & FP & FN & Acc. & Prec. & Neg.\,pred. & Sens. & Spec. & med.\,$d$ \\
\midrule
\endhead
\midrule
\multicolumn{12}{r}{\emph{continued}} \\
\endfoot
\bottomrule
\endlastfoot
\multicolumn{12}{l}{\textbf{Binary} (8 bits)} \\
1 & 5.18 & 1 & 255 & 0 & 0 & 1.0000 & 1.0000 & 1.0000 & 1.0000 & 1.0000 & -- \\
2 & 6.94 & 2 & 224 & 30 & 0 & 0.8828 & 0.0625 & 1.0000 & 1.0000 & 0.8819 & 8.0 \\
3 & 6.95 & 3 & 207 & 46 & 0 & 0.8203 & 0.0612 & 1.0000 & 1.0000 & 0.8182 & 12.5 \\
4 & 3.48 & 4 & 224 & 28 & 0 & 0.8906 & 0.1250 & 1.0000 & 1.0000 & 0.8889 & 7.5 \\
6 & 2.32 & 6 & 224 & 26 & 0 & 0.8984 & 0.1875 & 1.0000 & 1.0000 & 0.8960 & 7.0 \\
8 & 1.74 & 8 & 224 & 24 & 0 & 0.9062 & 0.2500 & 1.0000 & 1.0000 & 0.9032 & 6.5 \\
\addlinespace
\multicolumn{12}{l}{\textbf{Gray} (8 bits)} \\
1 & 5.17 & 1 & 255 & 0 & 0 & 1.0000 & 1.0000 & 1.0000 & 1.0000 & 1.0000 & -- \\
2 & 2.96 & 2 & 254 & 0 & 0 & 1.0000 & 1.0000 & 1.0000 & 1.0000 & 1.0000 & -- \\
3 & 2.30 & 3 & 253 & 0 & 0 & 1.0000 & 1.0000 & 1.0000 & 1.0000 & 1.0000 & -- \\
4 & 1.73 & 4 & 252 & 0 & 0 & 1.0000 & 1.0000 & 1.0000 & 1.0000 & 1.0000 & -- \\
6 & 1.39 & 6 & 250 & 0 & 0 & 1.0000 & 1.0000 & 1.0000 & 1.0000 & 1.0000 & -- \\
8 & 1.04 & 8 & 248 & 0 & 0 & 1.0000 & 1.0000 & 1.0000 & 1.0000 & 1.0000 & -- \\
\addlinespace
\multicolumn{12}{l}{\textbf{Thermometer} (255 bits)} \\
1 & 2.47 & 1 & 255 & 0 & 0 & 1.0000 & 1.0000 & 1.0000 & 1.0000 & 1.0000 & -- \\
2 & 1.24 & 2 & 254 & 0 & 0 & 1.0000 & 1.0000 & 1.0000 & 1.0000 & 1.0000 & -- \\
3 & 0.84 & 3 & 249 & 4 & 0 & 0.9844 & 0.4286 & 1.0000 & 1.0000 & 0.9842 & 1.5 \\
4 & 0.63 & 4 & 250 & 2 & 0 & 0.9922 & 0.6667 & 1.0000 & 1.0000 & 0.9921 & 1.0 \\
6 & 0.43 & 6 & 244 & 6 & 0 & 0.9766 & 0.5000 & 1.0000 & 1.0000 & 0.9760 & 2.0 \\
8 & 0.32 & 8 & 248 & 0 & 0 & 1.0000 & 1.0000 & 1.0000 & 1.0000 & 1.0000 & -- \\
\addlinespace
\multicolumn{12}{l}{\textbf{Hybrid lin. $B{=}4$} (65 bits)} \\
1 & 2.15 & 1 & 255 & 0 & 0 & 1.0000 & 1.0000 & 1.0000 & 1.0000 & 1.0000 & -- \\
2 & 1.09 & 2 & 254 & 0 & 0 & 1.0000 & 1.0000 & 1.0000 & 1.0000 & 1.0000 & -- \\
3 & 0.74 & 3 & 253 & 0 & 0 & 1.0000 & 1.0000 & 1.0000 & 1.0000 & 1.0000 & -- \\
4 & 0.56 & 4 & 252 & 0 & 0 & 1.0000 & 1.0000 & 1.0000 & 1.0000 & 1.0000 & -- \\
6 & 0.38 & 6 & 248 & 2 & 0 & 0.9922 & 0.7500 & 1.0000 & 1.0000 & 0.9920 & 1.0 \\
8 & 0.29 & 8 & 240 & 8 & 0 & 0.9688 & 0.5000 & 1.0000 & 1.0000 & 0.9677 & 2.5 \\
\addlinespace
\multicolumn{12}{l}{\textbf{Hybrid lin. $B{=}8$} (34 bits)} \\
1 & 2.11 & 1 & 255 & 0 & 0 & 1.0000 & 1.0000 & 1.0000 & 1.0000 & 1.0000 & -- \\
2 & 1.09 & 2 & 254 & 0 & 0 & 1.0000 & 1.0000 & 1.0000 & 1.0000 & 1.0000 & -- \\
3 & 0.75 & 3 & 253 & 0 & 0 & 1.0000 & 1.0000 & 1.0000 & 1.0000 & 1.0000 & -- \\
4 & 0.57 & 4 & 252 & 0 & 0 & 1.0000 & 1.0000 & 1.0000 & 1.0000 & 1.0000 & -- \\
6 & 0.39 & 6 & 250 & 0 & 0 & 1.0000 & 1.0000 & 1.0000 & 1.0000 & 1.0000 & -- \\
8 & 0.29 & 8 & 248 & 0 & 0 & 1.0000 & 1.0000 & 1.0000 & 1.0000 & 1.0000 & -- \\
\addlinespace
\multicolumn{12}{l}{\textbf{Johnson} (128 bits)} \\
1 & 2.24 & 1 & 255 & 0 & 0 & 1.0000 & 1.0000 & 1.0000 & 1.0000 & 1.0000 & -- \\
2 & 1.14 & 2 & 254 & 0 & 0 & 1.0000 & 1.0000 & 1.0000 & 1.0000 & 1.0000 & -- \\
3 & 0.76 & 3 & 253 & 0 & 0 & 1.0000 & 1.0000 & 1.0000 & 1.0000 & 1.0000 & -- \\
4 & 0.58 & 4 & 252 & 0 & 0 & 1.0000 & 1.0000 & 1.0000 & 1.0000 & 1.0000 & -- \\
6 & 0.39 & 6 & 250 & 0 & 0 & 1.0000 & 1.0000 & 1.0000 & 1.0000 & 1.0000 & -- \\
8 & 0.30 & 8 & 248 & 0 & 0 & 1.0000 & 1.0000 & 1.0000 & 1.0000 & 1.0000 & -- \\
\addlinespace
\multicolumn{12}{l}{\textbf{Hybrid cyc. $B{=}4$} (34 bits)} \\
1 & 2.11 & 1 & 255 & 0 & 0 & 1.0000 & 1.0000 & 1.0000 & 1.0000 & 1.0000 & -- \\
2 & 1.09 & 2 & 254 & 0 & 0 & 1.0000 & 1.0000 & 1.0000 & 1.0000 & 1.0000 & -- \\
3 & 0.75 & 3 & 253 & 0 & 0 & 1.0000 & 1.0000 & 1.0000 & 1.0000 & 1.0000 & -- \\
4 & 0.57 & 4 & 252 & 0 & 0 & 1.0000 & 1.0000 & 1.0000 & 1.0000 & 1.0000 & -- \\
6 & 0.39 & 6 & 250 & 0 & 0 & 1.0000 & 1.0000 & 1.0000 & 1.0000 & 1.0000 & -- \\
8 & 0.29 & 8 & 248 & 0 & 0 & 1.0000 & 1.0000 & 1.0000 & 1.0000 & 1.0000 & -- \\
\addlinespace
\multicolumn{12}{l}{\textbf{Hybrid cyc. $B{=}8$} (19 bits)} \\
1 & 2.19 & 1 & 255 & 0 & 0 & 1.0000 & 1.0000 & 1.0000 & 1.0000 & 1.0000 & -- \\
2 & 1.16 & 2 & 254 & 0 & 0 & 1.0000 & 1.0000 & 1.0000 & 1.0000 & 1.0000 & -- \\
3 & 0.82 & 3 & 253 & 0 & 0 & 1.0000 & 1.0000 & 1.0000 & 1.0000 & 1.0000 & -- \\
4 & 0.62 & 4 & 252 & 0 & 0 & 1.0000 & 1.0000 & 1.0000 & 1.0000 & 1.0000 & -- \\
6 & 0.44 & 6 & 250 & 0 & 0 & 1.0000 & 1.0000 & 1.0000 & 1.0000 & 1.0000 & -- \\
8 & 0.33 & 8 & 248 & 0 & 0 & 1.0000 & 1.0000 & 1.0000 & 1.0000 & 1.0000 & -- \\
\addlinespace
\end{longtable}
\end{footnotesize}

\subsection{Extended range, $N = 1024$}

\begin{footnotesize}
\begin{longtable}{rr rrrr rrrrr}
\caption{Extended range: complete classification metrics at $N = 1024$. Thermometer and Johnson exceed the default input capacity at this range and are not evaluated.}\label{tab:exp6full}\\
\toprule
$n$ & $S$ & TP & TN & FP & FN & Acc. & Prec. & Neg.\,pred. & Sens. & Spec. \\
\midrule
\endfirsthead
\multicolumn{11}{l}%
{\tablename\ \thetable{} -- continued} \\
\toprule
$n$ & $S$ & TP & TN & FP & FN & Acc. & Prec. & Neg.\,pred. & Sens. & Spec. \\
\midrule
\endhead
\midrule
\multicolumn{11}{r}{\emph{continued}} \\
\endfoot
\bottomrule
\endlastfoot
\multicolumn{11}{l}{\textbf{Binary} (10 bits)} \\
1 & 4.14 & 1 & 1023 & 0 & 0 & 1.0000 & 1.0000 & 1.0000 & 1.0000 & 1.0000 \\
2 & 6.95 & 2 & 897 & 125 & 0 & 0.8779 & 0.0157 & 1.0000 & 1.0000 & 0.8777 \\
4 & 3.48 & 4 & 896 & 124 & 0 & 0.8789 & 0.0312 & 1.0000 & 1.0000 & 0.8784 \\
8 & 1.75 & 8 & 896 & 120 & 0 & 0.8828 & 0.0625 & 1.0000 & 1.0000 & 0.8819 \\
\addlinespace
\multicolumn{11}{l}{\textbf{Gray} (10 bits)} \\
1 & 4.14 & 1 & 1023 & 0 & 0 & 1.0000 & 1.0000 & 1.0000 & 1.0000 & 1.0000 \\
2 & 2.30 & 2 & 1022 & 0 & 0 & 1.0000 & 1.0000 & 1.0000 & 1.0000 & 1.0000 \\
4 & 1.30 & 4 & 1020 & 0 & 0 & 1.0000 & 1.0000 & 1.0000 & 1.0000 & 1.0000 \\
8 & 0.75 & 8 & 1016 & 0 & 0 & 1.0000 & 1.0000 & 1.0000 & 1.0000 & 1.0000 \\
\addlinespace
\multicolumn{11}{l}{\textbf{HybridLin-B8} (130 bits)} \\
1 & 2.24 & 1 & 1023 & 0 & 0 & 1.0000 & 1.0000 & 1.0000 & 1.0000 & 1.0000 \\
2 & 1.14 & 2 & 1022 & 0 & 0 & 1.0000 & 1.0000 & 1.0000 & 1.0000 & 1.0000 \\
4 & 0.58 & 4 & 1008 & 12 & 0 & 0.9883 & 0.2500 & 1.0000 & 1.0000 & 0.9882 \\
8 & 0.29 & 8 & 1016 & 0 & 0 & 1.0000 & 1.0000 & 1.0000 & 1.0000 & 1.0000 \\
\addlinespace
\multicolumn{11}{l}{\textbf{HybridCyc-B8} (67 bits)} \\
1 & 2.15 & 1 & 1023 & 0 & 0 & 1.0000 & 1.0000 & 1.0000 & 1.0000 & 1.0000 \\
2 & 1.09 & 2 & 1022 & 0 & 0 & 1.0000 & 1.0000 & 1.0000 & 1.0000 & 1.0000 \\
4 & 0.56 & 4 & 1020 & 0 & 0 & 1.0000 & 1.0000 & 1.0000 & 1.0000 & 1.0000 \\
8 & 0.29 & 8 & 1000 & 16 & 0 & 0.9844 & 0.3333 & 1.0000 & 1.0000 & 0.9843 \\
\addlinespace
\multicolumn{11}{l}{\textbf{HybridLin-B16} (67 bits)} \\
1 & 2.15 & 1 & 1023 & 0 & 0 & 1.0000 & 1.0000 & 1.0000 & 1.0000 & 1.0000 \\
2 & 1.09 & 2 & 1022 & 0 & 0 & 1.0000 & 1.0000 & 1.0000 & 1.0000 & 1.0000 \\
4 & 0.56 & 4 & 1020 & 0 & 0 & 1.0000 & 1.0000 & 1.0000 & 1.0000 & 1.0000 \\
8 & 0.29 & 8 & 992 & 24 & 0 & 0.9766 & 0.2500 & 1.0000 & 1.0000 & 0.9764 \\
\addlinespace
\multicolumn{11}{l}{\textbf{HybridCyc-B16} (36 bits)} \\
1 & 2.11 & 1 & 1023 & 0 & 0 & 1.0000 & 1.0000 & 1.0000 & 1.0000 & 1.0000 \\
2 & 1.08 & 2 & 1022 & 0 & 0 & 1.0000 & 1.0000 & 1.0000 & 1.0000 & 1.0000 \\
4 & 0.56 & 4 & 1020 & 0 & 0 & 1.0000 & 1.0000 & 1.0000 & 1.0000 & 1.0000 \\
8 & 0.29 & 8 & 1016 & 0 & 0 & 1.0000 & 1.0000 & 1.0000 & 1.0000 & 1.0000 \\
\addlinespace
\end{longtable}
\end{footnotesize}

\end{document}